\documentclass[preprint,12pt,authoryear]{elsarticle}
\usepackage{amsmath}
\usepackage{amssymb}
\usepackage{bm}
\usepackage{booktabs}
\usepackage{multirow}
\usepackage{tabularx}
\usepackage{array}
\usepackage{makecell}
\usepackage[table]{xcolor}  
\usepackage{graphicx}
\usepackage{float}
\usepackage{subcaption}
\usepackage{algorithm}
\usepackage{algorithmicx}
\usepackage{algpseudocode}
\usepackage{tikz}
\usepackage{pgfplots}
\pgfplotsset{compat=1.18}
\usepackage[utf8]{inputenc}
\usepackage{textcomp}
\usepackage{textgreek}
\usepackage{verbatim}
\usepackage{pdflscape}
\usepackage{natbib}
\setcitestyle{round}
\usepackage{titlesec}

\titleformat{\paragraph}
  {\normalfont\normalsize}
  {}
  {0pt}
  {}

\titlespacing*{\paragraph}
  {0pt}
  {1.5ex plus 0.5ex minus 0.2ex}
  {0.5ex}
\usepackage[justification=justified, labelfont=bf]{caption}
\usepackage{hyperref}

\begin{document}
 
\begin{frontmatter}
 
\title{C-LiTE: A Lightweight Foundation Model for Cellular feature learning in Histopathology}
 
\author[1]{Esha Sadia Nasir\corref{cor1}}
\ead{esha.nasir@warwick.ac.uk}   
\cortext[cor1]{Corresponding author}
\author[1]{Jiaqi Lv}
\author[1]{Gozde N. Gunesli}
\author[1]{Shan E Ahmed Raza}

\affiliation[1]{organization={VISION Lab, Department of Computer Science, University of Warwick},
                city={Coventry},
                country={UK}}
% \affiliation[2]{organization={Department of Computer Science, University of Warwick},
%                 city={Coventry},
%                 country={UK}}

\begin{abstract}
Characterising individual cells is fundamental to computational pathology, as cellular morphology informs tasks such as tumour grading, mitotic assessment and immune profiling. However, most pathology foundation models are trained on tissue patches and capture cellular features only implicitly, while existing cell-level models are either restricted to specific cell populations or rely on computationally expensive architectures with hundreds of millions of parameters. Applying vision transformers (ViTs) directly to small cellular crops also presents an architectural challenge: upsampling increases computational cost without adding morphological information and may introduce interpolation artefacts, whereas conventional patch embeddings applied at native resolution yield insufficient spatial tokens for effective self-attention.

To address these limitations, we introduce C-LiTE, a lightweight foundation model for cellular representation learning in H\&E-stained histopathology, with only 22 million parameters. C-LiTE employs a native-resolution tokenisation strategy tailored to cellular inputs, processing $64\times64$ px cell-centred crops using a $4\times4$ patch embedding to produce a $16\times16$ token grid without image upsampling. To transfer visual knowledge from large-scale pre-training, we adapt the patch embedding weights of a pre-trained ViT-5-S using a Moore–Penrose pseudoinverse-based transformation, enabling weight initialisation at the finer patch resolution. We further assemble a pre-training dataset of 60 million cell-centred crops from The Cancer Genome Atlas (TCGA), balanced across 19 cancer types and 16 tissue types. To our knowledge, this represents the largest publicly available dataset for self-supervised pre-training on H\&E images at the cellular level.

We evaluate C-LiTE on seven public datasets spanning mitotic figure recognition, nucleus and cell-type classification, and immune cell phenotyping. Under frozen-feature evaluation, C-LiTE achieves a mean macro-F1 of 0.670, approaching CytoFormer (0.682) with approximately 30× fewer parameters, while outperforming tissue-level foundation models, including Virchow2 and UNI2. These results highlight the potential of compact, cell-centric architectures for learning transferable cellular representations.

\end{abstract}


\end{frontmatter}

% \newcolumntype{P}[1]{>{\RaggedRight\footnotesize}p{#1}}

\section{Introduction}

% background 
Haematoxylin and eosin (H\&E) staining forms the basis of histopathological diagnosis, providing information about tissue architecture and cellular composition. A single H\&E-stained WSI contains hundreds of thousands of nuclei belonging to different cell types. Analysis of these WSIs involves identifying the cell types (inferred from nuclei) present, quantifying their relative abundance, and characterising their spatial organisation \citep{yao2026cytoformer}. These cellular features, such as immune infiltration, provide important information for diagnostic, prognostic, and treatment decisions \citep{virchow1863cellular, zhang2025systematic} in clinical practice. However, extracting such information accurately at scale remains challenging. In current computational workflows, this often requires manual annotation of cells or nuclei by pathologists, followed by the development of supervised models specific to individual datasets \citep{horst_cellVit_2023,lv2026kongnet,nasir2023nuclei}. Such models typically require fine-tuning or retraining when applied to new datasets, limiting their generalisability and increasing the reliance on task-specific and dataset-specific annotations.  
% self supervised pathology foundation model

Recent advances in self-supervised foundation models offer a promising alternative, enabling the learning of transferable features from large-scale histopathology data that can be adapted to a wide range of downstream tasks. Models such as \textit{UNI} \citep{chen_towards_2024}, \textit{Virchow2} \citep{zimmermann_virchow2_2024}, \textit{CONCH} \citep{lu_towards_2023}, \textit{CTransPath} \citep{wang_transformer-based_2022}, \textit{Phikon} \citep{filiot_phikon-v2_2024}, \textit{Prov-GigaPath} \citep{xu_whole-slide_2024}, and \textit{TITAN} \citep{ding_multimodal_2025} have established a powerful paradigm for learning generalisable features from WSIs. These models are trained on either image patches containing hundreds of cells or whole-slide images (WSIs) containing thousands of cells, allowing them to capture a combination of cellular morphology, stromal composition, tissue architecture, and spatial context.
% existing SSL pathology models limitation
However, these models are primarily developed for whole-slide or patch-level feature extraction and cannot adequately capture the fine-grained morphological features required to distinguish individual cell types. Moreover, the substantial computational and memory requirements of these models present a significant challenge for cell-level analysis. Applying these models independently to thousands of nuclei within a single WSI would be computationally expensive and difficult to scale, limiting their practical utility for high-throughput cellular characterisation.

\begin{figure}[H]
    \centering
    \includegraphics[width=0.93\textwidth]{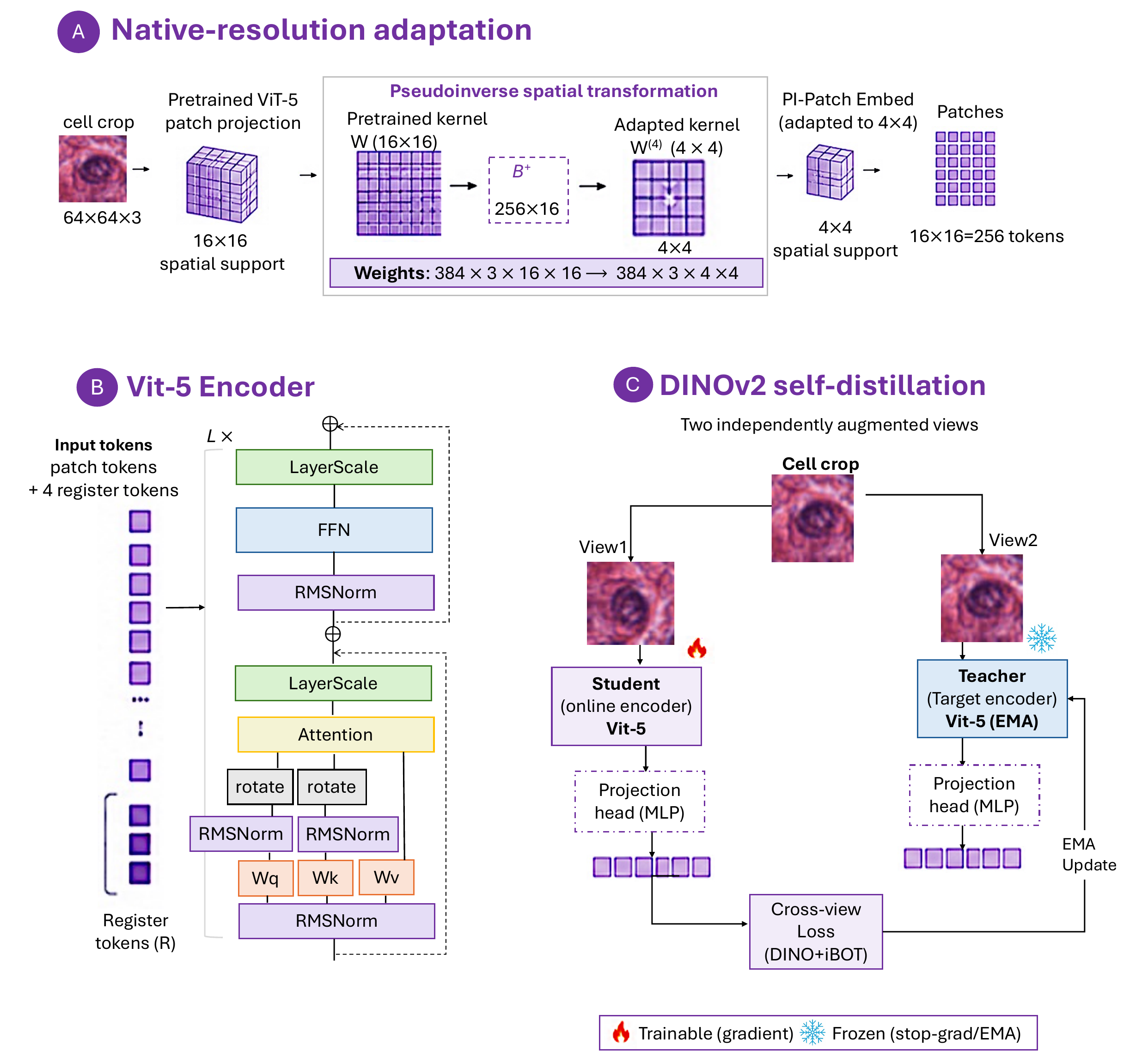}
    \captionsetup{width=\textwidth}
    \caption{\textbf{Proposed C-LiTE architecture for cell-centric feature learning.} Block (A) illustrates pseudoinverse initialisation (PI-PatchEmbed), which adapts the spatial resolution of the pre-trained patch projection from $16 \times 16$ to $4 \times 4$. Block (B) shows the adapted ViT-5 architecture used as the backbone for processing cell-centric image crops. Block (C) illustrates the DINOv2-style teacher–student self-distillation framework used for pre-training. The teacher model is an exponential moving average (EMA) of the student model.}
    \label{fig:method}
\end{figure}

% cell based foundation model and its limitation
Recognising these limitations, recent work has shifted towards cell-centric feature learning in histopathology \citep{chadoutaud2026lemon,yao2026cytoformer}. By focusing explicitly on individual cells, these approaches aim to learn features that better capture intrinsic cellular morphology and state. However, this shift towards cell-level modelling does not fully resolve the computational challenge. For example, CytoFormer \citep{yao2026cytoformer}, a recent cell-centric foundation model, comprises approximately 681 million parameters, retaining substantial computational and memory requirements comparable to those of existing pathology foundation models. This presents practical challenges for inference and limits the efficiency of cell-centric foundation models in large-scale cellular analysis. These limitations highlight the need for an efficient and lightweight foundation model for H\&E-stained cells. Such a model should learn rich and generalisable cellular representations while reducing computational requirements. This need is further reinforced by growing concerns about the computational and environmental costs of modern AI. The Green AI framework advocates treating computational efficiency as an objective alongside predictive performance, highlighting the financial and environmental costs of deep learning and the contribution of inference to overall AI deployment costs \citep{schwartz2020green,strubell2019energy,luccioni2024power}. To address this need, we introduce \textbf{C-LiTE}, a compact self-supervised model for generalisable cellular feature learning at single-cell resolution. C-LiTE is pre-trained on approximately 60M cell-centric image crops spanning 19 cancer types, 16 tissue types, and multiple staining protocols. This diverse training corpus captures a broad spectrum of cellular morphologies. We adapt the ViT-5 backbone within a teacher–student self-supervised framework using DINO and iBOT objectives. The resulting model learns transferable cellular representations that generalise across datasets and downstream tasks while preserving continuous morphological variation without being constrained by predefined cell categories. C-LiTE supports a diverse range of applications, including label-free cellular phenotyping, cell classification, mitosis classification, and morphology-based feature extraction. Across extensive evaluations, it achieves competitive performance against substantially larger cell-centric foundation models and tile-level pathology foundation models evaluated on cell crops. These results demonstrate the ability of C-LiTE to deliver high-quality cellular representations at substantially lower computational cost. The overall architecture of the proposed C-LiTE framework is presented in \autoref{fig:method}.
%contirbutions of proposed model
The main contributions of our work are as follows:
\begin{itemize}
    \item We develop \textbf{C-LiTE}, comprising:
    \begin{itemize}
        \item a compact \textbf{22M-parameter} model for cellular feature learning, adapting the ViT-5 backbone to process $64 \times 64$ px cell crops at native resolution within a self-supervised teacher--student framework.
        \item a large and diverse cell-centric pre-training corpus comprising over \textbf{60M cell-centric image crops}, spanning \textbf{19 cancer types} and \textbf{16 tissue types}, and covering diverse tissue contexts.
        \item the \textbf{Pseudoinverse-Initialised Patch Embedding} formulation \textbf{(PI-PatchEmbed)}, which transfers pre-trained ViT-5-S weights from $16 \times 16$ to $4 \times 4$ patch projections using a Moore--Penrose pseudoinverse-based spatial transfer operator.
    \end{itemize}
    \item We demonstrate that C-LiTE learns transferable cellular representations that generalise across datasets and downstream tasks, achieving performance comparable to CytoFormer~\citep{yao2026cytoformer} despite having approximately \textbf{30$\times$ fewer parameters}.
    \item We demonstrate the versatility of C-LiTE across diverse tasks, including immune cell phenotyping, cell-type classification, pan-cancer nucleus classification, hierarchical cell classification, and mitosis classification.
    \item We provide open access to the C-LiTE models, pre-training dataset, source code, and training configurations to facilitate reproducibility and encourage the research community to build upon our work.
\end{itemize}

\section{Related Work}
\subsection{Pathology Foundation Models}
The emergence of pathology foundation models has reshaped computational pathology, with self-supervised pre-training becoming a widely adopted approach for learning transferable features from large-scale histopathology data \citep{hua_pathoduet_2024,zhong2026suda}. Early transformer-based methods, such as CTransPath \citep{wang_transformer-based_2022}, demonstrated that features learned from a large collection of unlabelled tissue patches could transfer effectively across a range of downstream tasks. HIPT (Hierarchical Image Pyramid Transformer) \citep{chen_scaling_2022} subsequently extended this paradigm through hierarchical pre-training at multiple magnification levels, enabling the modelling of information across cellular and tissue scales.

Subsequent work has increasingly focused on scaling training data and model capacity, together with advances in pre-training strategies. For example, \textit{UNI} \citep{chen_towards_2024} adapted the DINOv2 framework and pre-trained a ViT-Large model on more than 100M image patches extracted from over 100K diagnostic slides spanning 20 tissue types. Building on this approach, \textit{UNI2} employs a ViT-Giant architecture and was trained on approximately 200M image tiles sampled from more than 350K diverse H\&E and IHC-stained slides.

\textit{Virchow} and \textit{Virchow2} \citep{zimmermann_virchow2_2024} further increased both model and dataset scale, employing a ViT-Huge architecture and training on approximately 3.1M WSIs. Their training framework uses a similar self-supervised teacher--student paradigm with global and local crops, together with extended-context translation (ECT) augmentation to preserve cellular morphology. In addition, the standard KoLeo regulariser was replaced with a kernel density estimation (KDE)-based regulariser to improve the learned feature distribution. \textit{Phikon} further increased the scale and diversity of training data, using approximately 460M pathology tiles spanning more than 30 cancer sites. It also adopts a DINOv2-based self-supervised framework \citep{filiot_phikon-v2_2024}. Alongside these image-based approaches, alternative pre-training paradigms have emerged that extend beyond image-only representation learning. \textit{CONCH} (CONtrastive learning from Captions for Histopathology) \citep{lu_towards_2023} introduced contrastive vision--language learning using diverse histopathology images and biomedical text, including more than 1.17M image--caption pairs within a task-agnostic pre-training framework.
\textit{Prov-GigaPath} \citep{xu_whole-slide_2024} shifted the focus towards slide-level representation learning for WSIs understanding. It was pre-trained on 1.3B $256 \times 256$ px image tiles extracted from 171K WSIs spanning 31 tissue types. A key component of its training strategy was the incorporation of an adapted LongNet architecture to enable efficient modelling of whole-slide sequences. Extending the development of pathology foundation models towards multimodal learning, \textit{TITAN} \citep{ding_multimodal_2025} was trained on 335K WSIs using self-supervised learning and vision--language alignment. Collectively, these studies have established tissue patches and WSIs as the predominant training units for pathology foundation models, while demonstrating the value of increasing data scale, model capacity, and pre-training diversity for learning generalisable pathology features \citep{rajendran_foundation_2027}.

\subsection{Supervised Cell Classification Models}

At the cellular level, computational pathology has traditionally relied on supervised methods for cell detection, segmentation, and classification. Approaches such as HoVer-Net \citep{graham_hover-net_2019}, StarDist \citep{schmidt_cell_2018}, and CellViT \citep{horst_cellVit_2023} have achieved strong performance through training on densely annotated datasets with predefined cell taxonomies. However, these methods are typically designed for specific downstream tasks, with their learned representations closely tied to the annotations and cell categories available during training. Developing such models often requires extensive manual annotation by experts, which limits scalability and can restrict generalisability across datasets. These limitations have motivated a shift towards learning transferable cellular features directly from large-scale, unlabelled histopathology data.

\subsection{Pathology Cell Foundation Models}
Building on this motivation, recent foundation models have explored self-supervised learning to capture generalisable features across diverse cellular morphologies and tissue contexts. \textit{VOLTA} \citep{nakhli2024volta} introduced environment-aware contrastive learning to jointly model individual cells and their surrounding micro-environment. It was trained on approximately 700K cells from 4 cancer types and demonstrated that meaningful cellular phenotypes can be learnt without manual annotation. \textit{DinoBloom} \citep{linguraru_DinoBloom_2024} further extended self-supervised pre-training to approximately 380K white blood cells (WBCs) from blood and bone marrow tissue across 13 publicly available datasets, producing features that transfer across a range of haematological tasks. More recently, \textit{LEMON} \citep{chadoutaud2026lemon} adopted a MoCo v3-based self-supervised framework and was trained on approximately 10M individual nuclei from 10 tissue types. Using $40 \times 40$ px nuclear crops, LEMON demonstrated that nuclear morphology contains rich biological information that can be captured without explicit cell-type labels. Similarly, CytoFormer \citep{yao2026cytoformer} further advanced cell-centric foundation modelling in histopathology by leveraging paired H\&E and spatial transcriptomics data to derive molecularly informed cell-type supervision across 16 tissue types. Together, these studies demonstrate the feasibility of learning transferable cellular representations at scale, but differ substantially in their cellular coverage, supervision, and model design. 
\paragraph{\textit{Limitations of Existing Methods}}
Existing supervised approaches remain focused on specific cellular datasets, while self-supervised cell foundation models require substantially larger architectures and, in some cases, additional molecular supervision. These approaches also face challenges in preserving fine-grained cellular information. For example, resizing cell crops from $56 \times 56$ $\mu$m to $224 \times 224$ $\mu$m changes the native spatial scale of the input and may lead to the loss of fine morphological detail. In contrast, \textit{LEMON} operates on smaller $40 \times 40$ crops with an $8 \times 8$ patch size, producing only 25 patch tokens. This is substantially fewer than the 256 patch tokens produced by a standard DINOv2 configuration using a $224 \times 224$ px input with a $14 \times 14$ patch size. The limited number of patch tokens may therefore restrict the spatial information available to the transformer for modelling fine-grained cellular morphology. 

These limitations raise many open questions. 
\begin{itemize}
    \item How effectively can pre-trained models transfer to cell crops smaller than their standard input resolution?
    \item What patch size is most appropriate for self-supervised learning at cellular resolution? 
    \item More broadly, can a compact model trained directly on a large and diverse collection of histopathological cells learn transferable feature representations of cellular morphology without relying on increasingly large architectures or explicit molecular supervision?
\end{itemize} 

Building on these questions, C-LiTE extends cell-centric foundation modelling to a substantially larger and more diverse histopathology corpus while maintaining a compact architecture.
% Unlike approaches developed for specialised cellular domains or supported by molecular supervision, C-LiTE is trained directly on more than 60M cell-centric crops spanning 19 cancer types, 16 tissue types and multiple tissue contexts. This combination of cellular diversity and computational efficiency provides a complementary approach to existing tissue- and cell-centric foundation models.

\section{Pre-training Data}
\label{sec:pretraining_data}

A key contribution of C-LiTE is the construction of, to the best of our knowledge, one of the largest and most diverse cell-centric histopathology pre-training corpora reported to date. The corpus comprises approximately 60M cell-centred image crops derived from TCGA WSIs across 19 cancer types. This heterogeneous collection provides a large-scale pre-training resource designed to capture diverse cellular morphologies and support the learning of robust and transferable cellular representations. The resulting model is subsequently evaluated across a diverse set of downstream tasks, including immune cell phenotyping, nucleus classification, cell-type classification, and mitosis classification.
\begin{figure}[p]
    \centering
    \includegraphics[width=0.92\textwidth]{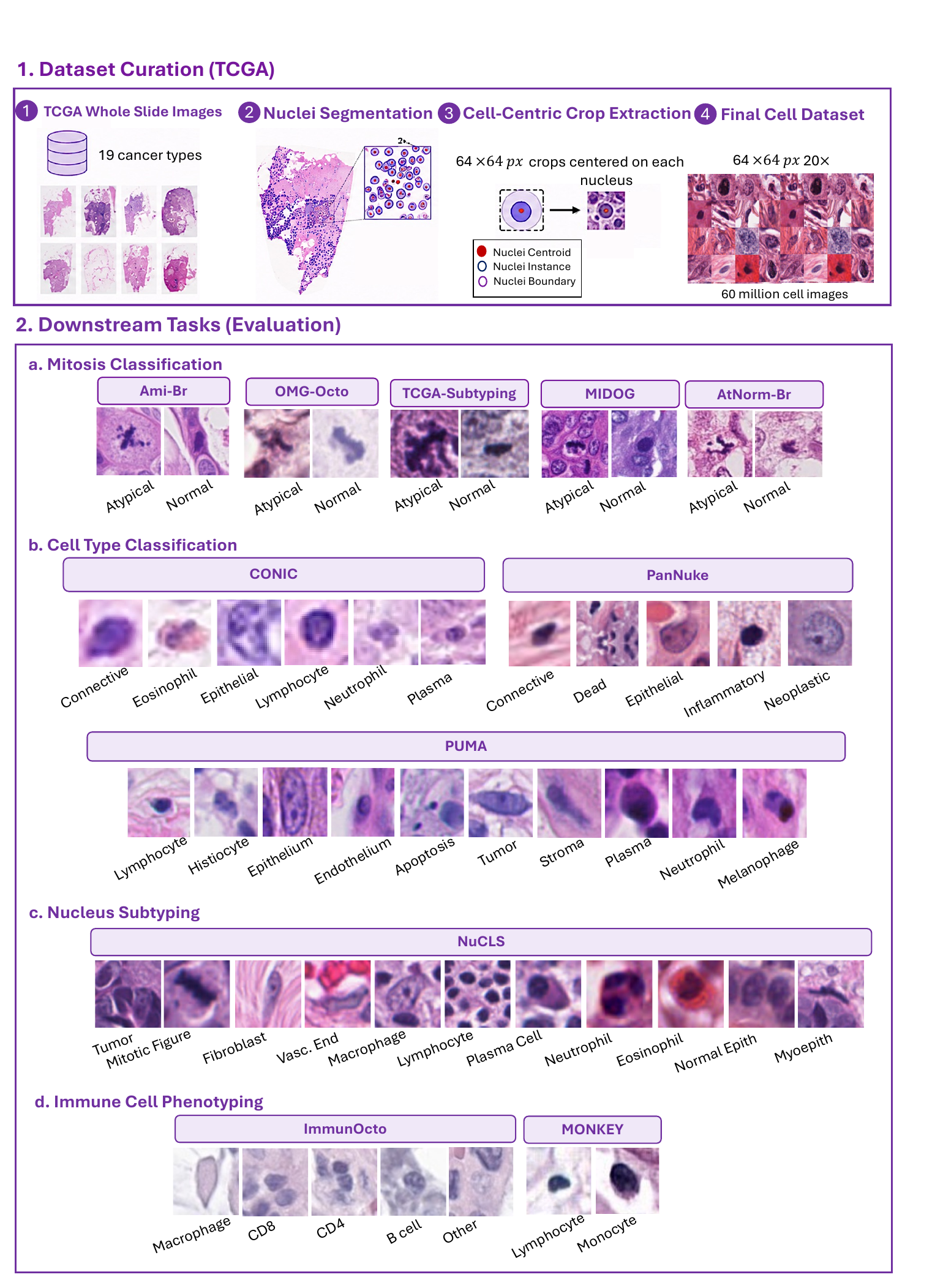}
    \captionsetup{width=\textwidth}
    \caption{\textbf{A pan-cancer H\&E cell dataset constructed by extracting cell crops from TCGA using HoVer-Net.} Data curation and downstream tasks: (1) WSIs from 19 cancer types in TCGA; (2) nucleus segmentation using HoVer-Net; (3) extraction of $64 \times 64$ px cell-centric crops for each segmented nucleus; and (4) construction of a final cell database containing 60 million cells. \textbf{Downstream tasks show the datasets used to benchmark the performance of C-LiTE against other state-of-the-art methods.}}
    \label{fig:downstream}
\end{figure}
\subsection{TCGA Cell Dataset}

The primary source of pre-training data in this study is The Cancer Genome Atlas (TCGA). The dataset covers 19 cancer types, including breast carcinoma (BRCA), lung adenocarcinoma (LUAD), lung squamous cell carcinoma (LUSC), ovarian serous cystadenocarcinoma (OV), colon adenocarcinoma (COAD), rectal adenocarcinoma (READ), cervical squamous cell carcinoma and endocervical adenocarcinoma (CESC), kidney renal clear cell carcinoma (KIRC), kidney renal papillary cell carcinoma (KIRP), pancreatic adenocarcinoma (PAAD), liver hepatocellular carcinoma (LIHC), prostate adenocarcinoma (PRAD), uterine corpus endometrial carcinoma (UCEC), uterine carcinosarcoma (UCS), thyroid carcinoma (THCA), stomach adenocarcinoma (STAD), head and neck squamous cell carcinoma (HNSC), bladder urothelial carcinoma (BLCA), and adrenocortical carcinoma (ACC). These 19 cancer types represent 16 distinct tissue types. Specifically, lung (LUAD and LUSC), kidney (KIRC and KIRP), and uterus (UCEC and UCS) each contribute two cancer-type cohorts originating from the same tissue, whereas the remaining 13 cancer types each correspond to a distinct tissue type. Throughout the paper, we use cancer type to refer to the 19 cohort-level categories and tissue type to refer to the 16 distinct tissues represented in the pre-training corpus.

\subsubsection{Data Generation}
Each WSI was processed using HoVer-Net \citep{graham_hover-net_2019} to obtain instance-level nuclear segmentations at a resolution of $0.50$ microns per pixel (mpp), corresponding to approximately $20\times$ magnification. Inference was performed on a high-performance computing (HPC) cluster equipped with \textit{NVIDIA A100 GPUs}. For each detected nucleus, a fixed-size $64 \times 64$ px cell-centric crop was extracted, centred on the predicted nuclear centroid. Because these detections were used directly to define crop locations without manual verification, the resulting corpus may contain detection noise arising from missed nuclei, false-positive detections, or inaccurate centroid estimates. Given the scale of the corpus ($\sim$60M crops), we did not explicitly quantify or correct for this noise, instead relying on the robustness of the self-supervised pre-training objective to occasional imperfect or miscentred crops. We have presented the data generation process in \autoref{fig:downstream}.

\subsubsection{Balanced Cancer-Type Sampling}
We observed that the number of extracted cells varies considerably across TCGA cancer types. Directly training on the full corpus would therefore result in substantial differences in the contribution of each cancer type to pre-training.
To reduce this imbalance, we fix the total pre-training budget at $N_{\mathrm{total}} = 60{,}000{,}000$ cells. Given the lightweight ViT-5-S backbone, this fixed budget also provides a controlled scale for pre-training while enabling balanced representations from each cancer type.
 We allocate the pre-training budget equally across each cancer type, such that the number of sampled cells per cancer type is

\begin{equation}
N_{\mathrm{sample}} = \frac{N_{\mathrm{total}}}{|C|}.
\label{eq:cell_balance}
\end{equation}

\begin{figure}[H]
    \centering
    \includegraphics[width=1\textwidth]{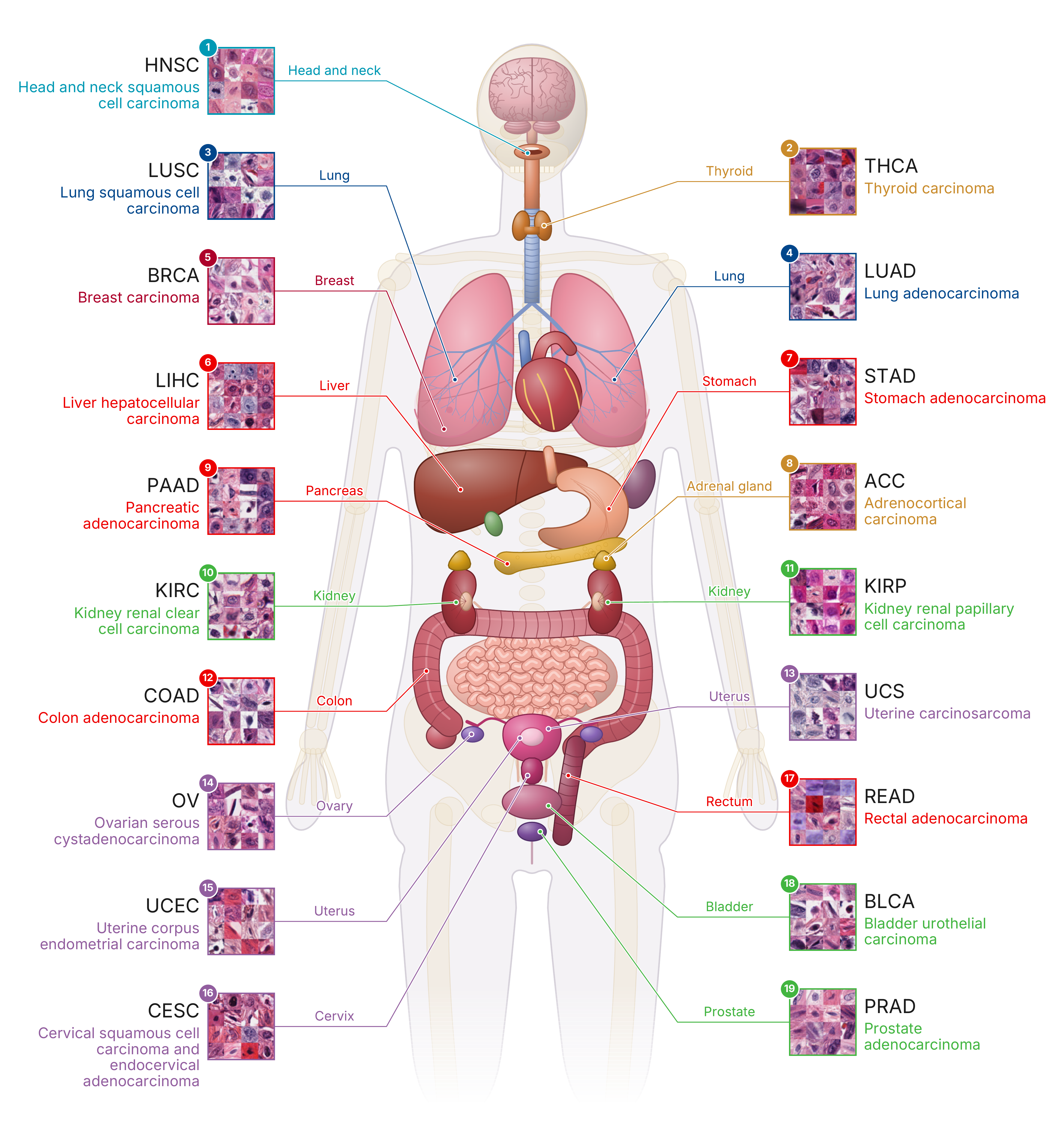}
    \captionsetup{width=\textwidth}
    \caption{Examples of $64 \times 64$ px H\&E cell crops curated from TCGA tissue slides from 16 tissue types and 19 cancer types for C-LiTE pre-training.}
    \label{fig:cell_crops}
\end{figure}

With $N_{\mathrm{total}} = 60{,}000{,}000$ and $|C| = 19$, $N_{\mathrm{sample}}$ in \autoref{eq:cell_balance} corresponds to $3{,}157{,}895$ cells per cancer type. For each cancer type, these cells are randomly selected without replacement from the available pool of $N_c$ cells, ensuring that each cancer type contributes equally to the pre-training corpus. Specifically, $N_c$ denotes the total number of available cells for cancer type $c$, and $C$ denotes the set of cancer types considered. Example cell crops from the 19 cancer types, grouped by organ, are shown in~\autoref{fig:cell_crops}.

\section{Methods}
\label{sec:method}
The overall framework for C-LiTE is shown in \autoref{fig:method}, and the downstream tasks are shown in \autoref{fig:downstream}. Here, pre-training data are first generated using HoVer-Net \citep{graham_hover-net_2019} and then subsampled to ensure balanced representation across cancer types.

\paragraph{\textit{Network architecture:}}
% It also preserves the original $16 \times 16$ spatial token topology.
The approach has three key components. First, we adapt Pseudoinverse-Initialised Patch Embedding (PI-PatchEmbed) to transfer the pre-trained $16 \times 16$ patch projection of ViT-5-S to a $4 \times 4$ projection for cell-scale images. This allows $64 \times 64$ px cell crops to use pre-trained ViT-5 weights directly, without resizing.  Second, we develop two pathology-specific distilled models based on UNI2 and Virchow2. We apply knowledge distillation to a subset of our data, using 1M images from each cancer type and 19M images in total. These distilled models provide pathology-specific weights for initialising the subsequent self-supervised learning. Third, we pre-train the adapted encoder using a DINOv2-style self-supervised learning framework. We modify the standard DINOv2 training recipe by using only global crops. We do not use local crops because each input already corresponds to a complete cell-centred image at the target spatial scale.

 \subsection{Pseudoinverse-Initialised Patch Embedding (PI-PatchEmbed)}
\label{sec:cell_encoder}

C-LiTE uses ViT-5-S \citep{wang2026Vit} as its ViT backbone. We selected ViT-5-S over standard ViT-S because its architectural refinements, including RMSNorm, QK-Norm, LayerScale, and 2-dimensional rotary positional embeddings (2D RoPE), improve performance at small parameter counts compared with a standard ViT-S of comparable size \citep{wang2026Vit}. These properties are well suited to the limited spatial extent of the cell crops. We retain these components and modify only the input patch projection. The original ViT-5-S uses a $16 \times 16$ patch projection for $224 \times 224$ px inputs, yielding a $14 \times 14$ spatial token grid. For $64 \times 64$ px cell crops, we replace this with a $4 \times 4$ projection. 

% Thus, the adaptation preserves the spatial token topology of the original architecture.

The adaptation of patch size requires transferring the spatial structure encoded by the pre-trained patch projection to a smaller spatial support. We apply PI-PatchEmbed using a pseudoinverse spatial transformation to initialise the $4 \times 4$ projection from the pre-trained $16 \times 16$ weights.

\subsubsection{Spatial Transfer Operator}
Given a flattened spatial filter of the pre-trained patch projection, denoted as $w_{16}$ in \autoref{eq:w16}:
\begin{equation}
w_{16} \in \mathbb{R}^{16^2}
\label{eq:w16}
\end{equation}
We define the linear operator $B$, given in \autoref{eq:B_def}, which corresponds to bicubic anti-aliased resizing from the $16 \times 16$ spatial domain to the $4 \times 4$ target domain:
\begin{equation}
B \in \mathbb{R}^{4^2 \times 16^2} = \mathbb{R}^{16 \times 256}
\label{eq:B_def}
\end{equation}

The matrix $B$ from \autoref{eq:B_def} is constructed by applying the interpolation operation to the $16 \times 16$ spatial vector space. For an input vector $x \in \mathbb{R}^{256}$, the corresponding resized representation is given in \autoref{eq:Bx}:
\begin{equation}
y = Bx, 
\qquad
x \in \mathbb{R}^{256}
\qquad
y \in \mathbb{R}^{16}
\label{eq:Bx}
\end{equation}
Here, $B$ is rectangular and has no conventional inverse. Instead of resizing the pre-trained kernel directly with $B$, we estimate the $4 \times 4$ kernel $w_4$ whose response is best preserved when mapped back through the resize operator by solving the least-squares problem in \autoref{eq:lsq}:
\begin{equation}
w_4^{*}=\operatorname*{arg\,min}_{w_4\in\mathbb{R}^{16}}\left\lVert B^{\top}w_4-w_{16}\right\rVert_2^2
\label{eq:lsq}
\end{equation}

This criterion tests whether up-projecting the small kernel through \(B^{\top}\) reconstructs the original pre-trained kernel, thereby preserving the patch projection under resizing. The least-squares solution to \autoref{eq:lsq} is obtained using the Moore-Penrose pseudoinverse of \(B\), denoted by \(B^+\) \citep{courrieu2008fastcomputationmoorepenroseinverse}. The pseudoinverse and the resulting solution are given in \autoref{eq:pinv}:
\begin{equation}
B^+\in\mathbb{R}^{256\times16},
\qquad
w_4^{*}=\left(B^{+}\right)^{\top}w_{16}
\label{eq:pinv}
\end{equation}
The Moore-Penrose pseudoinverse is computed numerically using
singular value decomposition (SVD), with a relative singular-value
cutoff of $10^{-4}$.

This pseudoinverse resizing follows the PI-resize formulation introduced in \citep{beyer2023flexiVitmodelpatchsizes} and is used here to transfer each pre-trained spatial filter to the target $4 \times 4$ patch size.

\subsubsection{Patch-Kernel Adaptation}
The pre-trained ViT-5-S patch projection (\(W\)) is defined in \autoref{eq:W_general}, where \(C_{\mathrm{out}}\) denotes the encoder embedding dimension (384 for ViT-5-S) and \(C_{\mathrm{in}}\) denotes the number of input channels. Thus, \(W\) can be written as \autoref{eq:W_concrete}:

\begin{equation}
W \in
\mathbb{R}^{C_{\mathrm{out}} \times C_{\mathrm{in}} \times 16 \times 16}
\label{eq:W_general}
\end{equation}

\begin{equation}
W \in
\mathbb{R}^{384 \times 3 \times 16 \times 16}
\label{eq:W_concrete}
\end{equation}

% We flatten the spatial dimensions of the projection to obtain $W_f$

To facilitate the subsequent spatial transfer operation, we flatten the two spatial dimensions of \(W\), while retaining its input and output channel dimensions. This gives the flattened projection matrix \(W_f\), as shown in \autoref{eq:Wf}:
\begin{equation}
W_f
\in
\mathbb{R}^{(C_{\mathrm{out}}C_{\mathrm{in}})\times256}
\label{eq:Wf}
\end{equation}

and the pseudoinverse spatial transformation from \autoref{eq:pinv} is then applied independently to each flattened filter, as expressed in \autoref{eq:Wf4}. The transformed weights are reshaped to $W^{(4)}$, as given in \autoref{eq:W4}:
\begin{equation}
W_f^{(4)}=W_fB^+
\label{eq:Wf4}
\end{equation}

\begin{equation}
W^{(4)}
\in
\mathbb{R}^{C_{\mathrm{out}}\times C_{\mathrm{in}}\times4\times4}
\label{eq:W4}
\end{equation}

Consequently, as shown by the transition from \autoref{eq:W_concrete} to \autoref{eq:W4}, the original $384 \times 3 \times 16 \times 16$ projection is converted into a $384 \times 3 \times 4 \times 4$ projection while preserving the embedding dimension. 

\subsection{Pathology Foundation model Distillation to ViT-5-S}

\begin{figure}[H]
    \centering
    \includegraphics[width=1\textwidth]{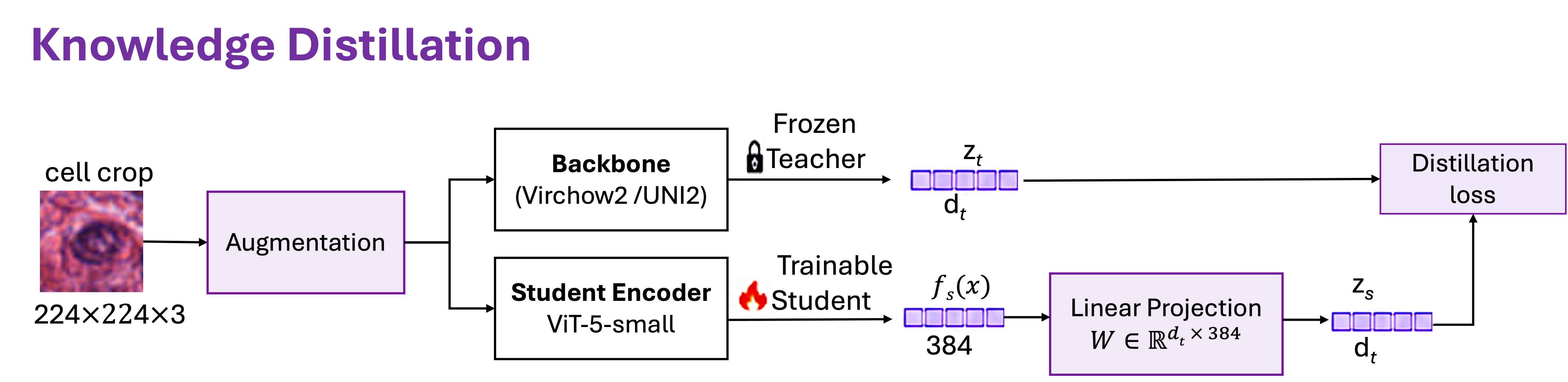}
    \captionsetup{width=\textwidth}
    \caption{Distillation process for transferring weights from a frozen teacher backbone (Virchow2/UNI2) to a trainable ViT-5-S student encoder. Cell crops are processed in parallel by the teacher and student, and the student encoder's embeddings are linearly projected into the teacher's embedding space and aligned using a distillation loss.}
\label{fig:distill}
\end{figure}

For initialising the model with pathology-specific pre-trained weights, we applied knowledge distillation from Virchow2 \citep{zimmermann_virchow2_2024} and UNI2 \citep{chen_towards_2024} on a subset of our training data and developed a ViT-5-S model. The subset was built by randomly sampling 1M cell crops from each TCGA tissue type, giving a cohort of 19M cell crops. Knowledge distillation was then applied by keeping the pre-trained teacher (Virchow2 or UNI2) frozen and making the student (ViT-5-S with a $16 \times 16$ patch projection) trainable. At this stage, no labels are used, and the student is trained purely to reproduce the teacher's embedding of each image. For this step, each $64 \times 64$ px crop is resized to $224 \times 224$ px and augmented with random horizontal and vertical flips and random rotations of up to $90^{\circ}$. The teacher and student receive an identical augmented view of each image. Since the student embedding ($d_s = 384$) and the teacher embedding ($d_t = 1536$ for UNI2 and $d_t = 1280$ for Virchow2) differ in dimensionality, a linear projection $W \in \mathbb{R}^{d_t \times d_s}$ is attached to the student output to match the teacher's embedding dimension. It is trained jointly with the student and discarded after distillation. This projection is required for loss estimation. For loss calculation, we use mean squared error (MSE) and cosine distance jointly. Thus, this training produces two distilled ViT-5-S models, one using UNI2 and the other using Virchow2 as the teacher. \autoref{fig:distill} demonstrates the distillation process used to transfer weights from the pre-trained UNI2 and Virchow2 models to the lightweight ViT-5-S architecture. The distillation training configurations are provided in Table \ref{tab:distill_config}.

\subsection{ViT-5 adapted DINOv2 self-supervised Training}
\label{sec:dino_training}
For self-supervised training on our dataset, we adapted the DINOv2 framework with the following key modifications. We replaced the original ViT backbone with ViT-5 and adapted the encoder to operate on cell-level crops. To investigate the effect of different initialisation strategies, we initialised the self-supervised training from three distinct sets of pre-trained weights: (i) ViT-5 weights pre-trained on ImageNet, (ii) ViT-5 weights distilled from UNI2, and (iii) ViT-5 weights distilled from Virchow2. These three initialisation strategies resulted in three C-LiTE variants:
\begin{itemize}
    \item C-LiTE: initialised from ImageNet pre-trained weights.
    \item C-LiTE$_U^*$: initialised from UNI2-distilled weights.
    \item C-LiTE$_V^*$: initialised from Virchow2-distilled weights.
\end{itemize}

\subsubsection{Cell-Centric ViT-5 Encoder}
\label{sec:encoder}

For an RGB cell crop ($x \in \mathbb{R}^{64 \times 64 \times 3}$), the adapted $4 \times 4$ patch projection produces 256 spatial tokens. A learnable class token $z_{\mathrm{cls}}$ ($z_{\mathrm{cls}} \in \mathbb{R}^{384}$) is prepended to the patch token sequence and serves as the global image representation used for downstream evaluation. Here, for the $i$-th patch $x_i \in \mathbb{R}^{4 \times 4 \times 3}$, the patch embedding is defined in \autoref{eq:patch_embed} and the resulting spatial token sequence $Z$ is given in \autoref{eq:Z_seq}. Following ViT-5, four learnable register tokens are appended to the spatial tokens as in \autoref{eq:registers}, giving the encoder's final input $Z_0$ defined in \autoref{eq:Z0}.

\begin{equation}
z_i=
W_e\operatorname{vec}(x_i)+b_e,
\qquad
z_i\in\mathbb{R}^{384}
\label{eq:patch_embed}
\end{equation}

\begin{equation}
Z=[z_1,z_2,\ldots,z_{256}]
\label{eq:Z_seq}
\end{equation}

\begin{equation}
R=[r_1,\ldots,r_4],
\qquad
r_k\in\mathbb{R}^{384}
\label{eq:registers}
\end{equation}
\begin{equation}
Z_0=
[z_{\mathrm{cls}},z_1,\ldots,z_{256},r_1,\ldots,r_4]
\label{eq:Z0}
\end{equation}

\paragraph{\textit{Cell-Centric Augmentation}}
\label{sec:cell_augmentation}

For each $64 \times 64$ px cell crop, two global views are generated independently. Both views retain the original field of view and undergo stochastic transformations comprising random rotation between $0^\circ$ and $360^\circ$ ($p=0.3$), horizontal and vertical flipping ($p=0.5$ independently), and $90^\circ$ rotation ($p=0.5$). Colour jittering ($p=0.8$), random greyscale conversion ($p=0.1$), and Gaussian blurring ($p=0.01$) are additionally applied.

The framework combines the DINO global self-distillation objective with the iBOT masked patch prediction objective. Two instances of the adapted ViT-5 encoder are maintained: a student network $f_{\theta_s}$ and a teacher network $f_{\theta_t}$. Both networks have identical architectures and are initialised with the same pre-trained weights. The student is updated through gradient-based optimisation, while the teacher parameters are updated using an exponential moving average (EMA) of the student parameters.

\subsubsection{DINO Image-level Objective}

For the global objective, we use the cross-entropy (CE) loss from DINOv2~\citep{oquab_dinov2_2023}. Given two differently augmented global views of the same cell, denoted by $v_1$ and $v_2$, the class token from each view is used as its image-level representation. The student class token is passed through the student DINO projection head, an MLP that maps the representation to $K$ prototype scores. These scores are converted into a probability distribution using a temperature-scaled softmax, yielding the student distribution $p_s(v_i)$ for view $v_i$. The teacher class token is processed analogously by the teacher projection head, after which Sinkhorn-Knopp (SK) normalisation is applied to the temperature-scaled prototype scores to obtain the teacher distribution $p_t(v_i)$. Here, $p_s^{(k)}(v_i)$ and $p_t^{(k)}(v_i)$ denote the probabilities assigned to prototype $k$ by the student and teacher, respectively, where $k \in \{1,\ldots,K\}$.

The DINO objective matches the teacher distribution from one view to the student distribution from the other view. For the two global views, the loss is defined as
\begin{equation}
\mathcal{L}_{\mathrm{DINO}}
= -\frac{1}{2}\sum_{i\neq j}\sum_{k=1}^{K}
p_t^{(k)}(v_i)\log p_s^{(k)}(v_j),
\label{eq}
\end{equation}
where $i,j \in \{1,2\}$ index the two global views, and $k$ indexes the $K$ prototypes. The student parameters are updated through gradient-based optimisation, while the teacher parameters are updated using an exponential moving average (EMA) of the student parameters.

\subsubsection{iBOT Masked Patch Prediction}

Following \citep{oquab_dinov2_2023} in parallel with the DINO objective, we use the iBOT objective. iBOT provides patch-level self-supervision by masking a subset of student patch tokens and predicting the corresponding teacher patch representations. The teacher receives the unmasked global view and provides targets for the masked locations. For view $v_i$, let $\mathcal{M}^{(i)}$ denote the set of masked patch indices. The iBOT loss is computed as the cross-entropy between the teacher and student predictions at these locations, as defined in \autoref{eq:ibot_loss}:
\begin{equation}
\mathcal{L}_{\mathrm{iBOT}}
= -\frac{1}{2}
\sum_{i=1}^{2}
\frac{1}{|\mathcal{M}^{(i)}|}
\sum_{k\in\mathcal{M}^{(i)}}
p_t^{(i,k)}
\log p_s^{(i,k)},
\label{eq:ibot_loss}
\end{equation}
where $p_s^{(i,k)}$ and $p_t^{(i,k)}$ denote the student and teacher probability distributions for patch $k$ in view $v_i$, respectively. Here, $i$ indexes the two global views and $k$ indexes the masked patches in $\mathcal{M}^{(i)}$. The student parameters are updated by gradient-based optimisation, while the teacher parameters are updated using an exponential moving average of the student parameters.
\subsubsection{Joint Pre-training Objective}

The final self-supervised objective combines the global DINO loss and the patch-level iBOT loss, as given in \autoref{eq:joint_loss}:
\begin{equation}
\mathcal{L}=
\lambda_{\mathrm{DINO}}
\mathcal{L}_{\mathrm{DINO}}
+
\lambda_{\mathrm{iBOT}}
\mathcal{L}_{\mathrm{iBOT}}
\label{eq:joint_loss}
\end{equation}
where $\lambda_{\mathrm{DINO}}$ and $\lambda_{\mathrm{iBOT}}$ control the relative contribution of the two objectives. The combined objective provides complementary supervision at global and local levels.

\subsubsection{KoLeo regularisation}

We apply the KoLeo regulariser to encourage a more uniform distribution of the learnt features within a batch. The regulariser is based on the Kozachenko-Leonenko differential entropy estimator. Given a batch of $n$ feature vectors $(x_1,\ldots,x_n)$, the KoLeo loss is defined in \autoref{eq:koleo_loss} and \autoref{eq:dist}:
\begin{equation}
\mathcal{L}_{\mathrm{KoLeo}}
= -\frac{1}{n}
\sum_{i=1}^{n}
\log(d_{n,i}),
\label{eq:koleo_loss}
\end{equation}
where
\begin{equation}
d_{n,i}=
\min_{j\neq i}
\left|
x_i-x_j
\right|_2
\label{eq:dist}
\end{equation}
is the distance between $x_i$ and its nearest neighbour in the batch. Before computing the regulariser, all feature vectors are $\ell_2$-normalised.

\section{Experiments}
\label{sec:experiments}

To evaluate C-LiTE, we compare it with state-of-the-art foundation models. All models were evaluated on the benchmark datasets described below under a common linear-probing protocol and using similar evaluation metrics.

\subsection{Experimental Settings}
\label{sec:experimental_setup}

All models were evaluated in a frozen backbone setting, with the pre-trained encoder kept fixed and a lightweight classifier trained on the resulting feature embeddings. For consistency across models, we used the class token embedding from the final transformer layer as the input representation for downstream evaluation.

\subsubsection{Baselines}
\label{sec:comparison_models}

The performance was compared against nine baseline models spanning three pre-training categories, summarised in~\autoref{tab:comparison_models}: \textit{natural-image models}, pre-trained on natural images without any pathology-specific data; \textit{patch-level pathology foundation models}, pre-trained on tissue patches containing multiple cells; and \textit{cell-level foundation models}, pre-trained directly on individual cells. Natural-image models (DINOv2) provide a lower bound: pre-trained without any pathology exposure, they establish how much transfer is achievable from generic visual features alone.

Patch-level pathology foundation models (Phikon, Virchow2, UNI2) are pre-trained on tissue patches containing hundreds of cells and therefore encode cellular morphology jointly with tissue architecture and stromal context; this additional context can aid tasks where surrounding tissue is diagnostic, but it also couples the representation to the micro-environment, as discussed in ~\autoref{sec:discussion}. Cell-level foundation models (DinoBloom, LEMON, CytoFormer) are pre-trained directly on cell crops and so isolate cellular morphology from tissue context. Among these, CytoFormer additionally incorporates molecular supervision from spatial transcriptomics, which is expected to sharpen decision boundaries for molecularly defined subtypes (e.g.\ CD4/CD8/CD20 populations) at the cost of a substantially larger architecture. This spans a parameter range of 22M--681M, allowing direct comparison from small models i.e., 22M parameters to computationally large parametric models i.e., 681M parameters. This comparison isolates the effect of pre-training domain from that of model scale. All baseline weights were obtained from their respective public releases and were trained using the linear classification head described in ~\autoref{sec:linear_probing}.

\begin{table}[H]
\centering
\captionsetup{width=\textwidth}
\caption{Baseline comparison models grouped by pre-training category. Params reflects the frozen backbone evaluated in ~\autoref{sec:results}. Input Res. denotes each model's native input resolution and patch size at inference.}
\label{tab:comparison_models}
\setlength{\tabcolsep}{4pt}
\resizebox{\textwidth}{!}{%
\begin{tabular}{llccl}
\toprule
\textbf{Model} & \textbf{Category} & \textbf{Params} & \textbf{Input Res. (Patch)} & \textbf{Pre-training Data} \\
\midrule
DINOv2 (ViT-S/14)~\citep{oquab_dinov2_2023} & Natural-image & 22M  & $224\times224$ (14) & LVD-142M \\
DINOv2 (ViT-B/14)~\citep{oquab_dinov2_2023} & Natural-image & 86M  & $224\times224$ (14) & LVD-142M \\
DINOv2 (ViT-L/14)~\citep{oquab_dinov2_2023} & Natural-image & 300M & $224\times224$ (14) & LVD-142M \\
\midrule
Phikon~\citep{filiot_phikon-v2_2024}     & Patch-level pathology & 85.8M & $224\times224$ (16) & TCGA WSI patches \\
Virchow2~\citep{zimmermann_virchow2_2024} & Patch-level pathology & 632M  & $224\times224$ (14) & Multi-institution 3.1M WSIs \\
UNI2~\citep{chen_towards_2024}             & Patch-level pathology & 681M  & $224\times224$ (14) & $>$100M patches, 100k+ slides \\
\midrule
DinoBloom~\citep{linguraru_DinoBloom_2024} & Cell-level & 22M  & $224\times224$ (14) & Blood/bone-marrow cells \\
LEMON~\citep{chadoutaud2026lemon}          & Cell-level & 22M  & $40\times40$ (8)    & Nucleus crops \\
CytoFormer~\citep{yao2026cytoformer}       & Cell-level & 681M & $224\times224$ (14) & H\&E + spatial transcriptomics, 16 organs \\
\midrule
\textbf{C-LiTE (Proposed)} & Cell-level & 22M & $64\times64$ (4) & TCGA cell crops, 19 cancer types \\
\bottomrule
\end{tabular}%
}
\end{table}
\subsubsection{Benchmark Datasets}
\label{sec:benchmark_datasets}

To evaluate the quality and generalisability of the learned representations, we benchmark C-LiTE across 7 publicly available datasets covering a diverse range of downstream tasks. These benchmarks include immune cell phenotyping, fine-grained cell classification, pan-cancer nucleus classification, mitosis classification, and hierarchical nucleus subtyping. Collectively, they span diverse organs, staining protocols, imaging resolutions, annotation strategies, and cellular taxonomies, providing a comprehensive assessment of representation transfer across biological and imaging domains. \autoref{fig:downstream} shows the process of dataset curation and downstream tasks performed via C-LiTE. 

\paragraph{\textit{Mitosis Classification:}}

The mitosis classification task involves classifying mitotic figures as normal or atypical. For this, we have employed the dataset from the MIDOG 2025 challenge \citep{aubreville2026mitosisdetectionwildmultitumor}. These datasets include the official MIDOG25 training data, along with Ami-Br \citep{palm_histologic_2025}, OMG-Octo \citep{shen2025omg}, AtNorm-Br \citep{banerjee2025benchmarkingdeeplearningvision} and TCGA Subtyping \citep{jahanifar2025mitosis}. Collectively, these datasets encompass multiple tissue types, institutions, and imaging domains, providing a challenging benchmark for mitosis recognition and domain generalisation \citep{10.1007/978-3-032-25180-0_6}. The corresponding dataset characteristics are summarised in~\autoref{tab:midog_data}.

\paragraph{\textit{NuCLS:}}

A multi-class classification dataset of single-cell images from breast cancer H\&E slides. Its key feature is a hierarchical annotation scheme, with labels reviewed and confirmed by pathologists. This structure allows us to evaluate the representative power of our embeddings on both fine-grained (e.g., distinguishing immune cell types) and coarse-grained (e.g., tumour vs. stromal) classification tasks. We refer to them as NuCLS (super) for broad categories, NuCLS (main) for main cell type categories and NuCLS (raw) for fine-grained type and subtype classifications \citep{amgad2022nucls}. The dataset statistics are summarised in~\autoref{tab:nucls_data}. In evaluation, we have reported the results only for NuCLS super category. However, the model can be used for other categories as well.

\paragraph{\textit{PanNuke:}}
The PanNuke dataset \citep{gamper2020pannuke} is a widely used benchmark for nuclei detection and classification in H\&E-stained images.
It contains 187,259 annotated nuclei from 19 tissue types, extracted
from over 20,000 WSIs at 40$\times$ objective magnification, and covers five cell types: neoplastic cells,
non-neoplastic epithelial cells, inflammatory cells, connective cells, and
dead cells. The dataset statistics are summarised in~\autoref{tab:pannuke_data}.

\paragraph{\textit{CoNIC:}}

The CoNIC dataset \citep{graham2021lizard} was released for the
2022 CoNIC Challenge and consists of H\&E-stained images for nuclei
detection and classification collected from 16 medical centres. The
images are of size $256 \times 256$ px at 20$\times$ objective magnification
(approximately 0.5 mpp). The dataset contains approximately 570,000
annotated nuclei across six cell types: epithelial cells, plasma cells,
lymphocytes, neutrophils, eosinophils, and connective cells. The dataset characteristics are summarised in~\autoref{tab:CoNIC_data}.

\paragraph{\textit{Machine-learning for Optimal detection of iNflammatory cells in the KidnEY (MONKEY):}}

The MONKEY dataset was released for the MONKEY challenge \cite{Van2026monkey} and comprises periodic acid-Schiff (PAS)-stained kidney slides for monocyte and lymphocyte detection and classification. Each WSI was scanned at a resolution of 0.24 mpp. Expert pathologists annotated the centroids of two cell types, lymphocytes and monocytes, across 231 large regions of interest from 153 WSIs collected at six pathology departments. Additional dataset details are summarised in~\autoref{tab:monkey_data}.

\paragraph{\textit{Immunocto:}}

Immunocto is the largest dataset of H\&E-stained cell phenotypes derived from mIF cell annotations \citep{simard_Immunocto_2025}. It contains single-cell crops sized $64 \times 64$ px at $40 \times$ magnification, along with nuclear masks and cell-phenotyping labels. The dataset contains 6.85M cell images, including more than 2.28M immune cells classified into four major populations: CD4$^{+}$ T cells, CD8$^{+}$ T cells, CD20$^{+}$ B cells, and CD68$^{+}$/CD163$^{+}$ macrophages. Its large scale and molecularly derived annotations make Immunocto a challenging benchmark for immune cell phenotyping. Dataset details are summarised in~\autoref{tab:Immunocto_data}.

\paragraph{\textit{Panoptic segmentation of nUclei and tissue in advanced MelanomA (PUMA):}}

The PUMA dataset was released as part of the 2025 challenge \citep{schuiveling_associations_2027} and contains H\&E-stained images of advanced melanoma with nucleus-level annotations. The full dataset comprises 155 primary melanoma regions of interest (ROIs) and 155 metastatic melanoma ROIs spanning eight organs. Each ROI is of size
$1024 \times 1024$ px and was scanned at 40$\times$ magnification (0.22 mpp).~\autoref{tab:puma_data} provides statistical details of the dataset.

We note that PanNuke, NuCLS and the TCGA subset of the mitosis benchmark include images derived from TCGA slides, which may overlap with the pre-training corpus. As pre-training is fully self-supervised, no downstream labels are exposed to the encoder; nevertheless, to assess generalisation to unseen data, we also evaluate on CoNIC, MONKEY, Immunocto and PUMA, which originate from sources independent of TCGA. Additional details of these benchmark datasets is summarised in \autoref{tab:benchmark_summary}.
\paragraph{\textit{Downstream Crop Extraction}}

For all benchmark datasets, $64 \times 64$ px cell-centred crops were extracted from the provided images at each dataset's native resolution, centred on the annotated nucleus centroid. For C-LiTE, these crops were used directly as input without resizing. For baseline models, crops were resized to each model's native input resolution (\autoref{tab:comparison_models}): $224 \times 224$ px for DINOv2, Phikon, UNI2, Virchow2, DinoBloom and CytoFormer, and $40 \times 40$ px for LEMON. Consequently, all models received the same field of view for each cell. We note that Immunocto, MONKEY and PUMA are acquired at approximately $0.22$–$0.25$ mpp, whereas the C-LiTE pre-training corpus was extracted at $0.5$ mpp; for these datasets, a $64 \times 64$ px crop therefore covers a smaller physical area (approximately $16 \times 16~\mu$m) than during pre-training (approximately $32 \times 32~\mu$m). The strong performance of C-LiTE on these benchmarks suggests that the learned representations are robust to this change in scale.

\subsubsection{Evaluation Metrics}
\label{sec:eval_metrics}

Downstream evaluation spans four task types with differing numbers of classes and degrees of class imbalance, ranging from binary classification (MIDOG, MONKEY) to 10-class classification (PUMA). To allow fair comparison across this range, we report a fixed set of metrics for every benchmark. Balanced accuracy and macro F1 are defined here; the remaining metrics include AUROC, mAP, macro precision, macro recall, and weighted F1.

Balanced accuracy averages per-class recall, thereby accounting for class imbalance, as shown in~\autoref{tab:pannuke_data} to~\autoref{tab:nucls_data}:
\begin{equation}
    \text{Balanced Acc.} = \frac{1}{K}\sum_{k=1}^{K} \frac{TP_k}{TP_k + FN_k},
    \label{eq:balanced_acc}
\end{equation}

where $K$ in \autoref{eq:balanced_acc} is the number of classes, and $TP_k$, $FN_k$ denote the true positives and false negatives for class $k$. Macro F1 averages the per-class harmonic mean of precision and recall, with equal weight regardless of class frequency:
\begin{equation}
    \text{Macro F1} = \frac{1}{K}\sum_{k=1}^{K} \frac{2 \cdot P_k \cdot R_k}{P_k + R_k},
    \label{eq:macro_f1}
\end{equation}
where $P_k$ and $R_k$ in \autoref{eq:macro_f1} denote precision and recall for class $k$. We adopt macro F1 as our primary summary statistic in cross-model comparisons. Weighted F1 is reported alongside macro F1 to additionally reflect performance under the natural (imbalanced) class distribution encountered at deployment.

All metrics are computed per random seed and reported as the mean (standard deviation) across five-fold cross-validation (CV) splits. In the case of PanNuke, we applied a standard three-fold split (~\autoref{sec:linear_probing}). For Immunocto, results are reported over a single large-scale train/test split defined in the original paper. Given its size (6.85M cells), calculating five-fold results was impractical for this dataset.

\subsubsection{Linear Probing Protocol}
\label{sec:linear_probing}

We used linear probing to evaluate the transferability of the frozen cell representations. Using each backbone, feature embeddings were extracted from the training and test sets, and these embeddings were used to train a multinomial logistic regression classifier. The logistic regression classifier uses $L_2$ regularisation. To account for differences in feature dimensionality and the number of classes across models and datasets, the inverse regularisation strength is defined as shown in~\autoref{eq:logistic_regularisation}:

\begin{equation}
    C = \frac{M \cdot K}{100}
    \label{eq:logistic_regularisation}
\end{equation}

where $M$ denotes the feature dimensionality and $K$ denotes the number of classes. The classifier is trained for a maximum of 1000 iterations.

\subsubsection{Compute}
\label{sec:compute}

We have reported separate compute and runtimes for the data generation and training phases.

\paragraph{\textit{Data Generation Runtimes}}

Mean per-slide GPU inference time was approximately 1.03 hours (measured on TCGA-BRCA). CPU post-processing averaged 16.21 hours per slide on a single core (estimated from 221 random TCGA-BRCA WSIs); in practice, this step was usually parallelised across multiple cores. Runtimes vary with slide size and tissue content.

\paragraph{\textit{Training Compute}}

Training was performed on 2 NVIDIA A100 80GB GPUs using mixed-precision training with FP16 parameters and gradient scaling. The model was trained with a per-GPU batch size of 256, corresponding to an effective batch size of 512 across the two GPUs. We used the AdamW optimiser with a base learning rate of $2.0 \times 10^{-4}$, a 20-epoch warm-up period, and a final weight decay of 0.2. Distributed training employed a parameter sharding strategy to improve GPU memory efficiency. Training configurations have been listed in~\autoref{tab:training_config}.

\subsection{Results}
\label{sec:results}

\autoref{tab:midog_perclass_prf1_main}--\autoref{tab:puma_perclass_prf1_main} summarise macro-F1 across the seven benchmarks, from which three consistent patterns emerge. First, despite its compact size, C-LiTE was competitive with the largest pathology foundation models throughout. It achieved the best macro-F1 on MIDOG (0.790$\pm$0.006), was effectively tied with CytoFormer on MONKEY (0.785 vs.\ 0.786), and outperformed every other 22M-parameter model on all seven benchmarks. Second, the effect of distillation depended on the task. C-LiTE$_U^*$, distilled from UNI2, achieved the best macro-F1 of all models on PanNuke (0.743$\pm$0.005) and NuCLS (0.613$\pm$0.002), whereas on MIDOG, CoNIC and MONKEY, C-LiTE outperformed both C-LiTE$_U^*$ and its Virchow2-distilled counterpart, C-LiTE$_V^*$. Third, CytoFormer led on the four remaining benchmarks (MONKEY, CoNIC, Immunocto and PUMA), but its margin over the best C-LiTE variant was substantial only on CoNIC (0.769 vs.\ 0.718 for C-LiTE); on Immunocto and PUMA the gap was 0.005 (C-LiTE$_V^*$) and 0.018 (C-LiTE$_U^*$), respectively.

\begin{figure}[H]
    \centering
    \includegraphics[width=1\textwidth]{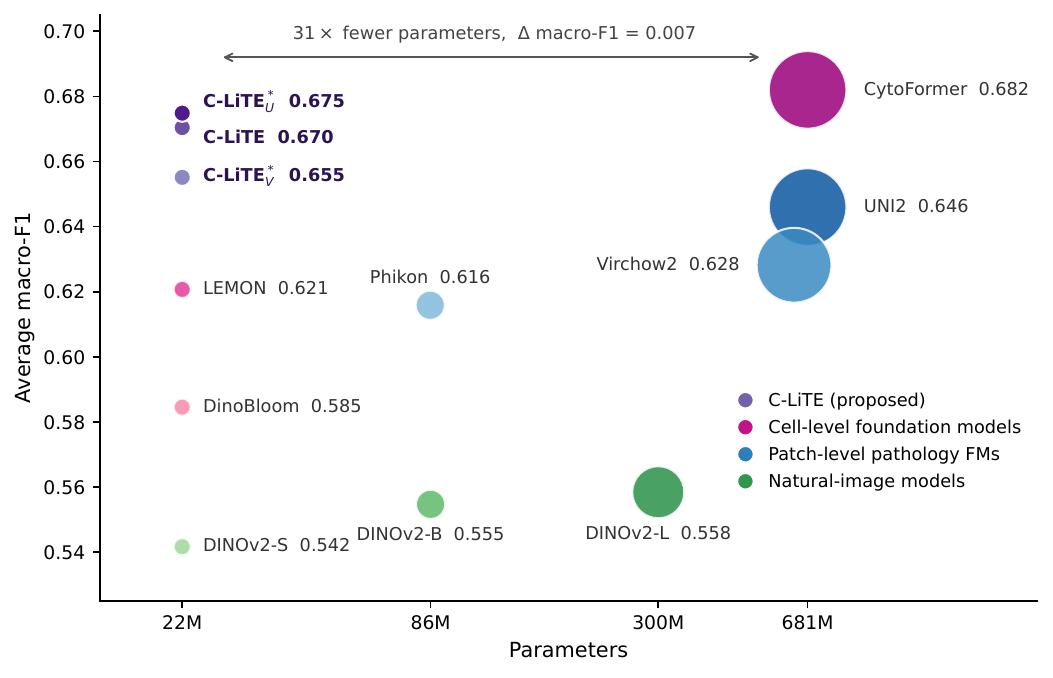}
    \captionsetup{width=\textwidth}
    \caption{Average macro-F1 across the seven benchmarks versus parameter count. Bubble area is proportional to parameter count. With 22M parameters, approximately 30-fold fewer than the largest cell and patch-level foundation models (681M), C-LiTE$_U^*$ (0.675) and C-LiTE (0.670) approach CytoFormer (0.682) in average macro-F1, and C-LiTE, C-LiTE$_V^*$ and C-LiTE$_U^*$ all exceed every other method.}
    \label{fig:param_vs_f1}
\end{figure}
These results are all the more notable given the difference in model size (\autoref{fig:param_vs_f1}). C-LiTE, C-LiTE$_V^*$ and C-LiTE$_U^*$ each have 22M parameters, approximately $\mathbf{30\times}$ fewer than UNI2 and CytoFormer (681M) and $28\times$ fewer than Virchow2 (632M). Averaged across benchmarks, C-LiTE$_U^*$ (0.675) and C-LiTE (0.670) came within 0.012 of CytoFormer (0.682), and all three variants, including C-LiTE$_V^*$ (0.655), exceeded UNI2 (0.646) and Virchow2 (0.628). Models of equal size fell well short (LEMON, 0.621; DinoBloom, 0.585; DINOv2 ViT-S/14, 0.542). Moreover, the distilled students generally surpassed their considerably larger teachers: C-LiTE$_U^*$ outperformed UNI2 on all seven benchmarks, and C-LiTE$_V^*$ outperformed Virchow2 on five. The following sections examine each task in turn.

\subsubsection{Normal vs.\ Atypical Mitosis Classification}

We begin with MIDOG, a clinically relevant binary task that requires distinguishing atypical from normal mitotic figures on the basis of subtle morphological differences, as shown in~\autoref{tab:midog_main}. This task most clearly favoured cell-level representations. C-LiTE achieved the highest macro-F1 (0.790$\pm$0.006) and ranked first on every metric, followed by C-LiTE$_V^*$ (0.786$\pm$0.007) and C-LiTE$_U^*$ (0.775$\pm$0.009). Even the 22M-parameter DinoBloom and LEMON (0.757 and 0.755) outperformed UNI2 (0.726) and Virchow2 (0.729), indicating that representations learned from whole tissue patches are less well suited to this fine morphological distinction. Distillation nonetheless transferred useful information, as C-LiTE$_V^*$ and C-LiTE$_U^*$ each improved substantially on their respective teachers. Per-class precision, recall and F1 are given in \autoref{tab:midog_perclass_prf1_main}.
\begin{table}[H]
\centering
\captionsetup{width=\textwidth}
\caption{Linear-probe classification on MIDOG. Best value per column in \textbf{bold}, second best \underline{underlined}.}
\label{tab:midog_main}
\hspace*{-0.05\textwidth}\resizebox{1.1\textwidth}{!}{%
\setlength{\tabcolsep}{1pt}%
\begin{tabular}{llcccccccc}
\toprule
\textbf{Model} & \textbf{Params} & \textbf{Acc.} & \textbf{Bal. Acc.} & \textbf{AUROC} & \textbf{mAP} & \textbf{Macro F1} & \textbf{Macro P} & \textbf{Macro R} & \textbf{Weighted F1} \\
\midrule
DINOv2 (ViT-S/14)               & 22M   & 0.834 (0.005) & 0.672 (0.011) & 0.811 (0.005) & 0.929 (0.002) & 0.700 (0.012) & 0.775 (0.009) & 0.672 (0.011) & 0.815 (0.006) \\
DINOv2 (ViT-B/14)               & 86M   & 0.838 (0.003) & 0.686 (0.005) & 0.816 (0.008) & 0.931 (0.004) & 0.714 (0.006) & 0.778 (0.007) & 0.686 (0.005) & 0.822 (0.004) \\
DINOv2 (ViT-L/14)~\citep{oquab_dinov2_2023}           & 300M  & 0.837 (0.003) & 0.693 (0.005) & 0.821 (0.001) & 0.930 (0.002) & 0.718 (0.005) & 0.770 (0.006) & 0.693 (0.005) & 0.823 (0.003) \\
\midrule
Phikon~\citep{filiot_phikon-v2_2024}                 & 85.8M & 0.838 (0.006) & 0.694 (0.010) & 0.818 (0.006) & 0.930 (0.003) & 0.720 (0.011) & 0.772 (0.011) & 0.694 (0.010) & 0.824 (0.007) \\
UNI2~\citep{chen_towards_2024}         & 681M & 0.834 (0.005)  & 0.707 (0.008) & 0.822 (0.008) & 0.934 (0.004) & 0.726 (0.009)  & 0.758 (0.009)  & 0.707 (0.008) & 0.825 (0.005) \\
Virchow2~\citep{zimmermann_virchow2_2024}                & 632M  & 0.838 (0.007) & 0.707 (0.008) & 0.825 (0.010) & 0.934 (0.004) & 0.729 (0.010) & 0.768 (0.015) & 0.707 (0.008) & 0.828 (0.007) \\
\midrule
DinoBloom~\citep{linguraru_DinoBloom_2024}             & 22M   & 0.856 (0.005) & 0.731 (0.007) & 0.869 (0.006) & 0.954 (0.002) & 0.757 (0.008)  & 0.803 (0.012) & 0.731 (0.007) & 0.846 (0.004) \\
LEMON~\citep{chadoutaud2026lemon}                 & 22M    & 0.857 (0.003) & 0.725 (0.005) & 0.857 (0.008) & 0.948 (0.004) & 0.755 (0.005) & 0.812 (0.009)  & 0.725 (0.005) & 0.846 (0.003) \\
CytoFormer~\citep{yao2026cytoformer}           & 681M & 0.851 (0.003) & 0.737 (0.006) & 0.865 (0.006) & 0.953 (0.003) & 0.757 (0.006)  & 0.787 (0.006)  & 0.737 (0.006) & 0.844 (0.003) \\
\midrule
\textbf{C-LiTE}  & 22M & \textbf{0.871 (0.004)} & \textbf{0.768 (0.006)} & \textbf{0.902 (0.004)} & \textbf{0.967 (0.002)} & \textbf{0.790 (0.006)} & \textbf{0.824 (0.007)} & \textbf{0.768 (0.006)} & \textbf{0.865 (0.004)} \\
\textbf{C-LiTE$_V^*$}  & 22M & \underline{0.870 (0.004)} & \underline{0.762 (0.008)} & \underline{0.892 (0.006)} & \underline{0.963 (0.003)} & \underline{0.786 (0.007)} & \textbf{0.824 (0.007)} & \underline{0.762 (0.008)} & \underline{0.863 (0.004)}  \\
\textbf{C-LiTE$_U^*$}  & 22M & 0.865 (0.006) & 0.748 (0.009) & 0.885 (0.007) & 0.961 (0.003) & 0.775 (0.009) & \underline{0.819 (0.011)} & 0.748 (0.009) & 0.857 (0.006) \\
\bottomrule
\end{tabular}%
}
\end{table}

\subsubsection{Hierarchical Nuclear Classification}

NuCLS presents a contrasting picture. Its hierarchical nuclear taxonomy makes it one of the most difficult benchmarks, with no model exceeding a macro-F1 of 0.62, as shown in~\autoref{tab:nucls_main}. In this setting, patch-level models were as effective as cell-level ones: UNI2 (0.602$\pm$0.005), Virchow2 (0.600$\pm$0.008) and CytoFormer (0.598$\pm$0.005) performed almost identically. Distillation proved decisive. C-LiTE$_U^*$ achieved the highest macro-F1 of all models (0.613$\pm$0.002), surpassing its teacher UNI2, whereas C-LiTE (0.590$\pm$0.005) trailed the large models slightly. The difference appears to lie in minority-class performance. C-LiTE ranked second only to C-LiTE$_U^*$ in accuracy, AUROC and weighted F1, but its balanced accuracy (0.573) was lower than that of UNI2 (0.593), suggesting that distillation from UNI2 primarily improved recognition of the rarer classes. This benefit did not extend to C-LiTE$_V^*$ (0.559), which fell below both C-LiTE and its teacher. Per-class results are reported in~\autoref{tab:nucls_perclass_prf1_main}.

\begin{table}[H]
\centering
\captionsetup{width=\textwidth}
\caption{Linear-probe classification on NuCLS (Super level). Best value per column in \textbf{bold}, second best \underline{underlined}.}
\label{tab:nucls_main}
\hspace*{-0.05\textwidth}\resizebox{1.1\textwidth}{!}{%
\setlength{\tabcolsep}{1pt}%
\begin{tabular}{llcccccccc}
\toprule
\textbf{Model} & \textbf{Params} & \textbf{Acc.} & \textbf{Bal. Acc.} & \textbf{AUROC} & \textbf{mAP} & \textbf{Macro F1} & \textbf{Macro P} & \textbf{Macro R} & \textbf{Weighted F1} \\
\midrule
DINOv2 (ViT-S/14)               & 22M   & 0.685 (0.003) & 0.488 (0.004) & 0.878 (0.001) & 0.535 (0.006) & 0.495 (0.006) & 0.556 (0.016) & 0.488 (0.004) & 0.656 (0.004) \\
DINOv2 (ViT-B/14)               & 86M   & 0.689 (0.003) & 0.503 (0.005) & 0.881 (0.002) & 0.541 (0.004) & 0.509 (0.006) & 0.544 (0.009) & 0.503 (0.005) & 0.663 (0.003) \\
DINOv2 (ViT-L/14)~\citep{oquab_dinov2_2023}              & 300M  & 0.691 (0.001) & 0.512 (0.006) & 0.883 (0.002) & 0.545 (0.007) & 0.519 (0.007) & 0.550 (0.009) & 0.512 (0.006) & 0.668 (0.002) \\
\midrule
Phikon~\citep{filiot_phikon-v2_2024}                & 85.8M & 0.708 (0.003) & 0.554 (0.012) & 0.897 (0.002) & 0.610 (0.011) & 0.569 (0.013) & 0.607 (0.013) & 0.554 (0.012) & 0.690 (0.004) \\
UNI2~\citep{chen_towards_2024}            & 681M & 0.721 (0.003) & \underline{0.593 (0.007)} & 0.907 (0.001) & 0.643 (0.003) & \underline{0.602 (0.005)} & 0.622 (0.005) & \underline{0.593 (0.007)} & 0.707 (0.004) \\
Virchow2~\citep{zimmermann_virchow2_2024}               & 632M  & 0.721 (0.002) & 0.582 (0.009) & 0.906 (0.001) & \underline{0.649 (0.009)} & 0.600 (0.008) & 0.640 (0.009) & 0.582 (0.009) & 0.706 (0.002) \\
\midrule
DinoBloom~\citep{linguraru_DinoBloom_2024}             & 22M   & 0.703 (0.003) & 0.516 (0.004) & 0.895 (0.001) & 0.573 (0.002) & 0.527 (0.005) & 0.577 (0.006) & 0.516 (0.004) & 0.679 (0.003) \\
LEMON~\citep{chadoutaud2026lemon}                 & 22M   & 0.726 (0.002) & 0.531 (0.004) & 0.910 (0.001) & 0.602 (0.005) & 0.546 (0.006) & \textbf{0.650 (0.017)} & 0.531 (0.004) & 0.708 (0.002) \\
CytoFormer~\citep{yao2026cytoformer}         & 681M    & 0.725 (0.001) & 0.581 (0.005) & 0.910 (0.001) & 0.638 (0.003) & 0.598 (0.005) & 0.630 (0.009) & 0.581 (0.005) & 0.713 (0.001) \\
\midrule
\textbf{C-LiTE}  & 22M & \underline{0.736 (0.001)} & 0.573 (0.003) & \underline{0.917 (0.001)} & 0.633 (0.006) & 0.590 (0.005) & 0.628 (0.010) & 0.573 (0.003) & \underline{0.722 (0.001)} \\
\textbf{C-LiTE$_V^*$}  & 22M & 0.725 (0.003) & 0.544 (0.006) & 0.909 (0.001) & 0.606 (0.004) & 0.559 (0.006) & 0.608 (0.008) & 0.544 (0.006) & 0.709 (0.003) \\
\textbf{C-LiTE$_U^*$}  & 22M & \textbf{0.742 (0.001)} & \textbf{0.596 (0.002)} & \textbf{0.920 (0.001)} & \textbf{0.662 (0.007)} & \textbf{0.613 (0.002)} & \underline{0.645 (0.006)} & \textbf{0.596 (0.002)} & \textbf{0.728 (0.002)} \\
\bottomrule
\end{tabular}%
}
\end{table}

\subsubsection{Nuclei Classification}

\paragraph{\textit{PanNuke:}}

The same pattern recurred on PanNuke, a pan-cancer nucleus classification benchmark evaluated with three-fold cross-validation, as shown in~\autoref{tab:pannuke_main}. C-LiTE (0.713$\pm$0.003) matched UNI2 and CytoFormer (both 0.715), with the lowest variance across folds of any model. Distillation again yielded a clear gain: C-LiTE$_U^*$ achieved the highest macro-F1 of 0.743$\pm$0.005, followed by C-LiTE$_V^*$ (0.723$\pm$0.005). As on NuCLS, the improvement came from minority classes rather than overall accuracy, which remained highest for UNI2; C-LiTE$_U^*$ instead led in balanced accuracy (0.734 vs.\ 0.705 for UNI2), mAP and macro precision. Per-class results are reported in~\autoref{tab:pannuke_perclass_prf1_main}.
\begin{table}[H]
\centering
\captionsetup{width=\textwidth}
\caption{Linear-probe classification on PanNuke~\citep{gamper2020pannuke}. Best value per column in \textbf{bold}, second best \underline{underlined}.}
\label{tab:pannuke_main}
\hspace*{-0.05\textwidth}\resizebox{1.1\textwidth}{!}{%
\setlength{\tabcolsep}{2pt}%
\begin{tabular}{llcccccccc}
\toprule
\textbf{Model} & \textbf{Params} & \textbf{Acc.} & \textbf{Bal. Acc.} & \textbf{AUROC} & \textbf{mAP} & \textbf{Macro F1} & \textbf{Macro P} & \textbf{Macro R} & \textbf{Weighted F1} \\
\midrule
DINOv2 (ViT-S/14)               & 22M   & 0.712 (0.003) & 0.626 (0.015) & 0.894 (0.002) & 0.699 (0.011) & 0.646 (0.011) & 0.673 (0.007) & 0.626 (0.015) & 0.708 (0.003) \\
DINOv2 (ViT-B/14)               & 86M   & 0.711 (0.005) & 0.627 (0.015) & 0.896 (0.003) & 0.704 (0.011) & 0.645 (0.012) & 0.670 (0.008) & 0.627 (0.015) & 0.707 (0.005) \\
DINOv2 (ViT-L/14)~\citep{oquab_dinov2_2023}              & 300M  & 0.717 (0.006) & 0.625 (0.017) & 0.899 (0.002) & 0.699 (0.013) & 0.645 (0.013) & 0.671 (0.010) & 0.625 (0.017) & 0.713 (0.005) \\
\midrule
Phikon~\citep{filiot_phikon-v2_2024}                & 85.8M & 0.754 (0.003) & 0.685 (0.013) & 0.922 (0.001) & 0.763 (0.015) & 0.698 (0.012) & 0.713 (0.012) & 0.685 (0.013) & 0.751 (0.003) \\
UNI2~\citep{chen_towards_2024}            & 681M & \textbf{0.778 (0.004)} & 0.705 (0.010) & \textbf{0.936 (0.002)} & 0.791 (0.010) & 0.715 (0.010) & 0.726 (0.011)  & 0.705 (0.010) & \textbf{0.776 (0.004)} \\
Virchow2~\citep{zimmermann_virchow2_2024}               & 632M  & \underline{0.775 (0.003)} & 0.698 (0.015) & \underline{0.935 (0.001)} & 0.775 (0.018) & 0.711 (0.014) & 0.727 (0.013) & 0.698 (0.015) & \underline{0.773 (0.003)} \\
\midrule
DinoBloom~\citep{linguraru_DinoBloom_2024}               & 22M   & 0.734 (0.003) & 0.640 (0.020) & 0.911 (0.001) & 0.726 (0.014) & 0.660 (0.016) & 0.686 (0.012) & 0.640 (0.020) & 0.730 (0.003) \\
LEMON~\citep{chadoutaud2026lemon}                 & 22M    & 0.730 (0.003) & 0.648 (0.014) & 0.906 (0.001) & 0.729 (0.012) & 0.670 (0.011) & 0.703 (0.009) & 0.648 (0.014) & 0.726 (0.002) \\
CytoFormer~\citep{yao2026cytoformer}                & 681M    & \underline{0.775 (0.003)} & 0.699 (0.015)  & 0.934 (0.001)  & \underline{0.805 (0.013)} & 0.715 (0.013) & 0.734 (0.011) & 0.699 (0.015) & \underline{0.773 (0.003)} \\
\midrule
\textbf{C-LiTE}  & 22M & 0.764 (0.002)  & 0.697 (0.005)  & 0.926 (0.001)  & 0.781 (0.006) & 0.713 (0.003) & 0.734 (0.004)  & 0.697 (0.005)   & 0.761 (0.002) \\
\textbf{C-LiTE$_V^*$}  & 22M & 0.758 (0.003) & \underline{0.710 (0.006)} & 0.923 (0.001) & 0.795 (0.004) & \underline{0.723 (0.005)} & \underline{0.739 (0.004)} & \underline{0.710 (0.006)} & 0.756 (0.003) \\
\textbf{C-LiTE$_U^*$}  & 22M & 0.774 (0.002) & \textbf{0.734 (0.007)} & 0.933 (0.002) & \textbf{0.819 (0.004)} & \textbf{0.743 (0.005)} & \textbf{0.754 (0.004)} & \textbf{0.734 (0.007)} & 0.772 (0.002) \\
\bottomrule
\end{tabular}%
}
\end{table}
\paragraph{\textit{CoNIC:}}

CoNIC~\citep{graham2021lizard} tests whether these gains persist under a finer taxonomy, as its six colon-specific classes resolve the single PanNuke inflammatory category into eosinophils, neutrophils, lymphocytes and plasma cells (\autoref{tab:CoNIC_main}). Here, the effect of distillation was mixed. CytoFormer achieved the highest macro-F1 ($0.769\pm0.003$) and led on most metrics, with C-LiTE$_U^*$ second ($0.727\pm0.003$), followed by C-LiTE ($0.718\pm0.002$), C-LiTE$_V^*$ ($0.683\pm0.002$) and all remaining baselines, including UNI2 ($0.682$). C-LiTE$_U^*$ was the only model to exceed CytoFormer on accuracy (0.847 vs.\ 0.833) and weighted F1 (0.843 vs.\ 0.830), and it also improved on C-LiTE in balanced accuracy (0.701 vs.\ 0.693). In contrast, C-LiTE$_V^*$ attained higher accuracy than C-LiTE (0.825 vs.\ 0.813) but a markedly lower balanced accuracy (0.652 vs.\ 0.693). The per-class results show that both distilled variants improved on the abundant classes (connective, epithelial and lymphocyte), but only C-LiTE$_U^*$ matched C-LiTE on eosinophils, and both fell below it on neutrophils and plasma cells. CytoFormer retained a clear lead on these rare inflammatory subtypes. The difference between the two distilled variants may reflect their teachers, since UNI2 performs considerably better than Virchow2 on this benchmark, but we did not test this directly. Per-class results are reported in~\autoref{tab:CoNIC_perclass_prf1_main}.

\begin{table}[H]
\centering
\captionsetup{width=\textwidth}
\caption{Linear-probe classification on CoNIC~\citep{graham2021lizard}. Best value per column in \textbf{bold}, second best \underline{underlined}.}
\label{tab:CoNIC_main}
\hspace*{-0.03\textwidth}\resizebox{1.1\textwidth}{!}{%
\setlength{\tabcolsep}{1pt}%
\begin{tabular}{llcccccccc}
\toprule
\textbf{Model} & \textbf{Params} & \textbf{Acc.} & \textbf{Bal. Acc.} & \textbf{AUROC} & \textbf{mAP} & \textbf{Macro F1} & \textbf{Macro P} & \textbf{Macro R} & \textbf{Weighted F1} \\
\midrule
DINOv2 (ViT-S/14)               & 22M   & 0.711 (0.001) & 0.461 (0.002) & 0.904 (0.001) & 0.552 (0.002) & 0.494 (0.002) & 0.588 (0.003) & 0.461 (0.002) & 0.696 (0.001) \\
DINOv2 (ViT-B/14)               & 86M   & 0.719 (0.001) & 0.480 (0.002) & 0.909 (0.001) & 0.576 (0.002) & 0.518 (0.002) & 0.609 (0.003) & 0.480 (0.002) & 0.705 (0.001) \\
DINOv2 (ViT-L/14)~\citep{oquab_dinov2_2023}             & 300M  & 0.726 (0.001) & 0.492 (0.002) & 0.916 (0.001) & 0.586 (0.003) & 0.529 (0.003) & 0.613 (0.005) & 0.492 (0.002) & 0.715 (0.001) \\
\midrule
Phikon~\citep{filiot_phikon-v2_2024}               & 85.8M & 0.767 (0.001) & 0.594 (0.003) & 0.939 (0.001) & 0.682 (0.004) & 0.625 (0.002) & 0.673 (0.002) & 0.594 (0.003) & 0.761 (0.001) \\
UNI2~\citep{chen_towards_2024}            & 681M & 0.801 (0.001) & 0.657 (0.002) & 0.957 (0.001) & 0.703 (0.003) & 0.682 (0.002) & 0.716 (0.001) & 0.657 (0.002) & 0.798 (0.001) \\
Virchow2~\citep{zimmermann_virchow2_2024}               & 632M  & 0.786 (0.001) & 0.604 (0.004) & 0.950 (0.001) & 0.692 (0.002) & 0.634 (0.004) & 0.680 (0.002) & 0.604 (0.004) & 0.781 (0.001) \\
\midrule
DinoBloom~\citep{linguraru_DinoBloom_2024}             & 22M   & 0.746 (0.001) & 0.533 (0.003) & 0.925 (0.001) & 0.620 (0.002) & 0.568 (0.003) & 0.637 (0.003) & 0.533 (0.003) & 0.737 (0.001) \\
LEMON~\citep{chadoutaud2026lemon}          & 22M & 0.796 (0.001) & 0.630 (0.002)  & 0.957 (0.001) & 0.724 (0.002) & 0.665 (0.001) & 0.728 (0.003)   & 0.630 (0.002)   & 0.790 (0.001) \\
CytoFormer~\citep{yao2026cytoformer}         & 681M & \underline{0.833 (0.001)} & \textbf{0.744 (0.004)}  & \textbf{0.969 (0.001)} & \textbf{0.814 (0.003)} & \textbf{0.769 (0.003)}  & \textbf{0.801 (0.002)} & \textbf{0.744 (0.004)} & \underline{0.830 (0.001)} \\
\midrule
\textbf{C-LiTE}  & 22M & 0.813 (0.001) & 0.693 (0.002) & 0.958 (0.001) & \underline{0.789 (0.001)} & 0.718 (0.002) & 0.751 (0.002) & 0.693 (0.002) & 0.810 (0.001) \\
\textbf{C-LiTE$_V^*$}  & 22M & 0.825 (0.001) & 0.652 (0.002) & 0.961 (0.000) & 0.746 (0.002) & 0.683 (0.002) & 0.732 (0.003) & 0.652 (0.002) & 0.820 (0.001) \\

\textbf{C-LiTE$_U^*$}  & 22M & \textbf{0.847 (0.001)} & \underline{0.701 (0.003)} & \underline{0.962 (0.000)} & 0.746 (0.002) & \underline{0.727 (0.003)} & \underline{0.764 (0.003)} & \underline{0.701 (0.003)} & \textbf{0.843 (0.001)} \\
\bottomrule
\end{tabular}%
}
\end{table}

\subsubsection{Immune Cell Classification}
\paragraph{\textit{MONKEY:}}
We next examined immune cells more directly, beginning with MONKEY, a binary lymphocyte-monocyte task, as shown in~\autoref{tab:monkey_main}. As on MIDOG, cell-level models held a clear advantage. CytoFormer (0.786$\pm$0.002), C-LiTE (0.785$\pm$0.001) and C-LiTE$_U^*$ (0.784$\pm$0.002) performed on par, with LEMON (0.777$\pm$0.002) and C-LiTE$_V^*$ (0.774$\pm$0.004) close behind, whereas UNI2 (0.744$\pm$0.004) and Virchow2 (0.723$\pm$0.003) trailed substantially. Within this leading group, C-LiTE achieved the highest AUROC (0.879) and macro precision (0.799) of all models. Per-class results are reported in~\autoref{tab:monkey_perclass_prf1_main}.
\begin{table}[H]
\centering
\captionsetup{width=\textwidth}
\caption{Linear-probe classification on MONKEY. Best value per column in \textbf{bold}, second best \underline{underlined}.}
\label{tab:monkey_main}
\hspace*{-0.05\textwidth}\resizebox{1.1\textwidth}{!}{%
\setlength{\tabcolsep}{1pt}%
\begin{tabular}{llcccccccc}
\toprule
\textbf{Model} & \textbf{Params} & \textbf{Acc.} & \textbf{Bal. Acc.} & \textbf{AUROC} & \textbf{mAP} & \textbf{Macro F1} & \textbf{Macro P} & \textbf{Macro R} & \textbf{Weighted F1} \\
\midrule
DINOv2 (ViT-S/14)               & 22M   & 0.722 (0.001) & 0.650 (0.001) & 0.754 (0.003) & 0.608 (0.002) & 0.658 (0.001) & 0.697 (0.001) & 0.650 (0.001) & 0.703 (0.001) \\
DINOv2 (ViT-B/14)               & 86M   & 0.730 (0.002) & 0.664 (0.003) & 0.770 (0.002) & 0.626 (0.003) & 0.673 (0.003) & 0.707 (0.002) & 0.664 (0.003) & 0.715 (0.002) \\
DINOv2 (ViT-L/14)~\citep{oquab_dinov2_2023}             & 300M  & 0.728 (0.002) & 0.663 (0.002) & 0.769 (0.003) & 0.629 (0.002) & 0.671 (0.002) & 0.703 (0.002) & 0.663 (0.002) & 0.713 (0.002) \\
\midrule
Phikon~\citep{filiot_phikon-v2_2024}                & 85.8M & 0.759 (0.001) & 0.709 (0.001) & 0.814 (0.001) & 0.698 (0.002) & 0.718 (0.001) & 0.738 (0.001) & 0.709 (0.001) & 0.751 (0.001) \\
UNI2~\citep{chen_towards_2024}             & 681M & 0.779 (0.004) & 0.735 (0.004) & 0.830 (0.003) & 0.727 (0.004) & 0.744 (0.004) & 0.762 (0.004) & 0.735 (0.004) & 0.773 (0.004) \\
Virchow2~\citep{zimmermann_virchow2_2024}              & 632M  & 0.763 (0.002) & 0.713 (0.003) & 0.813 (0.001) & 0.697 (0.004) & 0.723 (0.003) & 0.744 (0.003) & 0.713 (0.003) & 0.755 (0.003) \\
\midrule
DinoBloom~\citep{linguraru_DinoBloom_2024}              & 22M   & 0.748 (0.001) & 0.693 (0.002) & 0.801 (0.003) & 0.669 (0.005) & 0.702 (0.002) & 0.725 (0.002) & 0.693 (0.002) & 0.738 (0.001) \\
LEMON~\citep{chadoutaud2026lemon}          & 22M  & 0.806 (0.002) & 0.767 (0.002) & 0.861 (0.002) & 0.766 (0.003) & 0.777 (0.002) & 0.792 (0.002) & 0.767 (0.002) & 0.801 (0.002) \\
CytoFormer~\citep{yao2026cytoformer} & 681M & \textbf{0.812 (0.002)} & \textbf{0.778 (0.001)} & 0.875 (0.002) & 0.787 (0.006) & \textbf{0.786 (0.002)} & \underline{0.798 (0.002)} & \textbf{0.778 (0.001)} & \textbf{0.809 (0.002)} \\
\midrule
\textbf{C-LiTE}  & 22M & \textbf{0.812 (0.001)} & \underline{0.776 (0.001)} & \textbf{0.879 (0.002)} & \underline{0.788 (0.003)} & \underline{0.785 (0.001)} & \textbf{0.799 (0.002)} & \underline{0.776 (0.001)} & \underline{0.808 (0.001)} \\
\textbf{C-LiTE$_V^*$}  & 22M & 0.802 (0.003) & 0.767 (0.004) & 0.867 (0.002) & 0.778 (0.003) & 0.774 (0.004) & 0.786 (0.003) & 0.767 (0.004) & 0.799 (0.003)  \\
\textbf{C-LiTE$_U^*$}  & 22M & \underline{0.810 (0.002)} & \textbf{0.778 (0.002)} & \underline{0.876 (0.002)} & \textbf{0.792 (0.004)} & 0.784 (0.002) & 0.794 (0.002) & \textbf{0.778 (0.002)} & 0.807 (0.002) \\
\bottomrule
\end{tabular}%
}
\end{table}

\subsubsection{Immune Cell Phenotyping - Immunocto}
Immunocto extends this evaluation to fine-grained immune phenotyping at scale. Owing to its size (6.85M cells), it was evaluated on a single split following the original paper~\citep{simard_Immunocto_2025}, and no standard deviations are therefore reported (\autoref{sec:eval_metrics}, \autoref{tab:immune_main}). CytoFormer led on every metric (0.574), consistent with its transcriptomics-derived supervision, which directly targets molecularly defined immune populations. C-LiTE$_V^*$ (0.569), C-LiTE$_U^*$ (0.567) and C-LiTE (0.557) followed closely and outperformed all other models by a wide margin, including UNI2 (0.510) and Virchow2 (0.461). In contrast to CoNIC, distillation was beneficial here, and the gains over the teachers were the largest of any benchmark: C-LiTE$_V^*$ exceeded Virchow2 by 0.108 and C-LiTE$_U^*$ exceeded UNI2 by 0.057. Per-class results are reported in~\autoref{tab:immune_perclass_prf1_main}.

\begin{table}[H]
\centering
\captionsetup{width=\textwidth}
\caption{Linear-probe classification on Immunocto~\citep{simard_Immunocto_2025}. Best value per column in \textbf{bold}, second best \underline{underlined}.}
\label{tab:immune_main}
\resizebox{\textwidth}{!}{%
\begin{tabular}{llcccccccc}
\toprule
\textbf{Model} & \textbf{Params} & \textbf{Acc.} & \textbf{Bal. Acc.} & \textbf{AUROC} & \textbf{mAP} & \textbf{Macro F1} & \textbf{Macro P} & \textbf{Macro R} & \textbf{Weighted F1} \\
\midrule
DINOv2 (ViT-S/14)               & 22M   & 0.711 & 0.416 & 0.850 & 0.489 & 0.399 & 0.539 & 0.416 & 0.643 \\
DINOv2 (ViT-B/14)               & 86M   & 0.714 & 0.415 & 0.856 & 0.502 & 0.406 & 0.552 & 0.415 & 0.646 \\
DINOv2 (ViT-L/14)~\citep{oquab_dinov2_2023}             & 300M  & 0.717 & 0.423 & 0.861 & 0.511 & 0.414 & 0.562 & 0.423 & 0.651 \\
\midrule
Phikon~\citep{filiot_phikon-v2_2024}               & 85.8M & 0.736 & 0.458 & 0.892 & 0.548 & 0.451 & 0.579 & 0.458 & 0.682 \\
UNI2~\citep{chen_towards_2024}          & 681M & 0.754 & 0.503 & 0.912 & 0.596 & 0.510 & 0.622 & 0.503 & 0.714 \\
Virchow2~\citep{zimmermann_virchow2_2024}              & 632M  & 0.739 & 0.462 & 0.898 & 0.561 & 0.461 & 0.593 & 0.462 & 0.688  \\
\midrule
DinoBloom~\citep{linguraru_DinoBloom_2024}             & 22M   & 0.732 & 0.450 & 0.884 & 0.528 & 0.442 & 0.577 & 0.450 & 0.678 \\
LEMON~\citep{chadoutaud2026lemon}        & 22M  & 0.770 & 0.519 & 0.931 & 0.579 & 0.494 & 0.596 & 0.519 & 0.736 \\
CytoFormer~\citep{yao2026cytoformer}         & 681M & \textbf{0.799} & \textbf{0.569} & \textbf{0.952} & \textbf{0.653} & \textbf{0.574} & \textbf{0.659} & \textbf{0.569} & \textbf{0.774}  \\
\midrule
\textbf{C-LiTE}  & 22M & 0.793 & 0.558 & 0.943 & 0.631 & 0.557 & 0.644 & 0.558 & 0.764  \\
\textbf{C-LiTE$_V^*$}  & 22M & \underline{0.796} & \underline{0.562} & \underline{0.948} & 0.640 & \underline{0.569} & \underline{0.656} & \underline{0.562} & \underline{0.771}  \\
\textbf{C-LiTE$_U^*$}  & 22M & 0.794 & \underline{0.562} & 0.947 & \underline{0.643} & 0.567 & 0.651 & \underline{0.562} & 0.768 \\
\bottomrule
\end{tabular}%
}
\end{table}

\subsubsection{Multi-Organ Nuclear Phenotyping (PUMA)}

Finally, PUMA is the most fine-grained benchmark in our study, comprising ten nuclear phenotypes across multiple tissue compartments, as shown in~\autoref{tab:puma_main}. CytoFormer again achieved the highest macro-F1 (0.575), followed by C-LiTE$_U^*$ (0.557). C-LiTE (0.540) performed on par with UNI2 (0.543) and Virchow2 (0.539), trading higher macro precision (0.611) for lower balanced accuracy (0.503), and substantially outperformed LEMON (0.438) and DinoBloom (0.436). As on NuCLS, distillation from UNI2 helped whereas distillation from Virchow2 did not, with C-LiTE$_V^*$ (0.492) falling below its teacher. Per-class results are reported in~\autoref{tab:puma_perclass_prf1_main}.

\begin{table}[H]
\centering
\captionsetup{width=\textwidth}
\caption{Linear-probe classification on PUMA. Best value per column in \textbf{bold}, second best \underline{underlined}.}
\label{tab:puma_main}
\hspace*{-0.05\textwidth}\resizebox{1.1\textwidth}{!}{%
\setlength{\tabcolsep}{1pt}%
\begin{tabular}{llcccccccc}
\toprule
\textbf{Model} & \textbf{Params} & \textbf{Acc.} & \textbf{Bal. Acc.} & \textbf{AUROC} & \textbf{mAP} & \textbf{Macro F1} & \textbf{Macro P} & \textbf{Macro R} & \textbf{Weighted F1} \\
\midrule
DINOv2 (ViT-S/14)               & 22M   & 0.723 (0.004) & 0.345 (0.006) & 0.880 (0.001) & 0.424 (0.006) & 0.400 (0.009) & 0.535 (0.016) & 0.345 (0.006) & 0.694 (0.004) \\
DINOv2 (ViT-B/14)               & 86M   & 0.729 (0.002) & 0.363 (0.006) & 0.885 (0.002) & 0.434 (0.007) & 0.418 (0.007) & 0.544 (0.010) & 0.363 (0.006) & 0.703 (0.002) \\
DINOv2 (ViT-L/14)~\citep{oquab_dinov2_2023}              & 300M  & 0.728 (0.003) & 0.371 (0.004) & 0.887 (0.002) & 0.426 (0.003) & 0.413 (0.004) & 0.496 (0.009) & 0.371 (0.004) & 0.705 (0.003) \\
\midrule
Phikon~\citep{filiot_phikon-v2_2024}               & 85.8M & 0.762 (0.002) & 0.489 (0.006) & 0.919 (0.000) & 0.547 (0.009) & 0.530 (0.006) & 0.594 (0.014) & 0.489 (0.006) & 0.747 (0.002) \\
UNI2~\citep{chen_towards_2024}           & 681M & 0.788 (0.001) & \underline{0.523 (0.007)} & \textbf{0.938 (0.001)} & 0.551 (0.008) & 0.543 (0.006) & 0.569 (0.006) & \underline{0.523 (0.007)} & 0.780 (0.001) \\
Virchow2~\citep{zimmermann_virchow2_2024}               & 632M  & 0.775 (0.001) & 0.509 (0.007) & 0.933 (0.001) & 0.564 (0.002) & 0.539 (0.006) & 0.582 (0.005) & 0.509 (0.007) & 0.764 (0.002) \\
\midrule
DinoBloom~\citep{linguraru_DinoBloom_2024}              & 22M   & 0.746 (0.003) & 0.392 (0.006) & 0.904 (0.002) & 0.469 (0.003) & 0.436 (0.006) & 0.533 (0.008) & 0.392 (0.006) & 0.724 (0.003) \\
LEMON~\citep{chadoutaud2026lemon}                 & 22M   & 0.750 (0.003) & 0.384 (0.007) & 0.902 (0.001) & 0.496 (0.004) & 0.438 (0.007) & 0.583 (0.019) & 0.384 (0.007) & 0.723 (0.004) \\
CytoFormer~\citep{yao2026cytoformer}         & 681M    & \textbf{0.795 (0.002)} & \textbf{0.538 (0.006)} & \textbf{0.938 (0.001)} & \textbf{0.611 (0.009)} & \textbf{0.575 (0.004)} & \textbf{0.625 (0.008)} & \textbf{0.538 (0.006)} & \textbf{0.786 (0.002)} \\
\midrule
\textbf{C-LiTE}  & 22M & 0.784 (0.002) & 0.503 (0.005) & 0.927 (0.001) & 0.585 (0.008) & 0.540 (0.004) & 0.611 (0.009) & 0.503 (0.005) & 0.767 (0.001) \\
\textbf{C-LiTE$_V^*$}  & 22M & 0.769 (0.001) & 0.446 (0.004) & 0.916 (0.001) & 0.519 (0.004) & 0.492 (0.004) & 0.587 (0.013) & 0.446 (0.004) & 0.748 (0.001) \\
\textbf{C-LiTE$_U^*$}  & 22M & \underline{0.794 (0.002)} & 0.516 (0.008) & \underline{0.936 (0.001)} & \underline{0.590 (0.007)} & \underline{0.557 (0.008)} & \underline{0.623 (0.010)} & 0.516 (0.008) & \underline{0.781 (0.002)} \\
\bottomrule
\end{tabular}%
}
\end{table}

\paragraph{\textit{Per-Class Performance:}}

The per-class results help explain these benchmark-level differences. Across benchmarks, performance was lowest on minority and morphologically ambiguous classes, such as CD8$^{+}$ T cells (best F1 0.269) and macrophages (best F1 below 0.47) in Immunocto, endothelium in PUMA (0.40), and the ambiguous class in NuCLS (0.40). These classes also account for much of CytoFormer's advantage. It retained a clear lead on eosinophils, neutrophils and plasma cells in CoNIC, CD8$^{+}$ T cells in Immunocto, and epithelial cells and histiocytes in PUMA, which may reflect its greater capacity as well as its molecularly informed supervision. The value of cell-level pre-training is equally evident at this level: natural-image backbones reached F1 scores of only 0.31--0.41 for CoNIC eosinophils and neutrophils, compared with 0.841 and 0.753 for CytoFormer and 0.701 and 0.662 for C-LiTE.
 
Conversely, C-LiTE achieved the highest F1 of all methods for both MIDOG classes and for three of the four NuCLS classes, matched CytoFormer on MONKEY lymphocytes, and led on lymphocytes, melanophage and stroma in PUMA. On PanNuke, it led only on the inflammatory class, whereas the rare dead-cell class was best recognised by C-LiTE$_U^*$ (F1 0.699), in line with the minority-class gains from distillation observed on NuCLS and PanNuke.
\subsection{Embedding Space Visualisation}

\begin{figure}[H]
    \centering
    \includegraphics[width=1.2\textwidth]{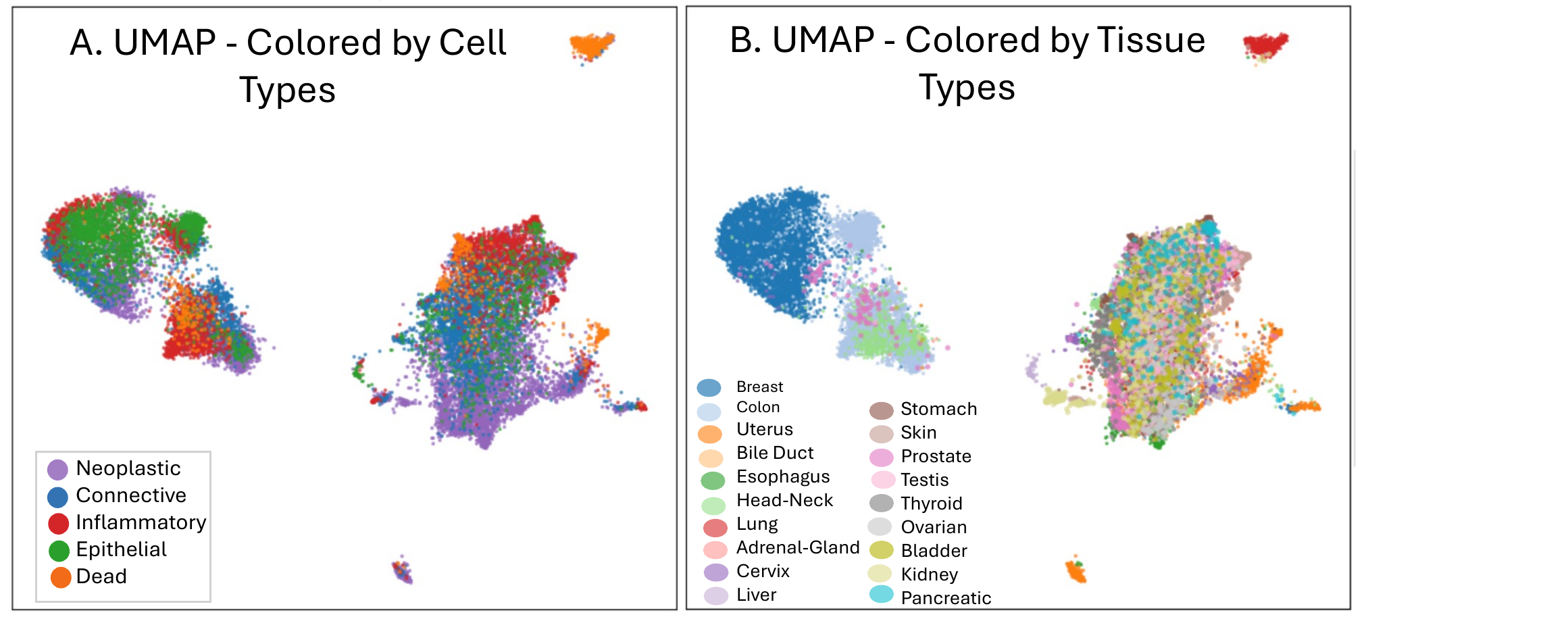}
    \captionsetup{width=\textwidth}
    \caption{UMAP projection of cell embeddings. (A) Colored by cell types extracted from C-LiTE.  (B) Colored by tissue types.
    }
    \label{fig:umap}
\end{figure}

To complement the quantitative evaluation, we qualitatively examine the geometry of the learnt embedding space using UMAP projections of C-LiTE cell embeddings. We examine whether the embedding space organises cells by morphological structures, and whetehr cells  of the same tissue type cluster together.
As shown in~\autoref{fig:umap}, the C-LiTE representation forms visibly compact and separable clusters by cell type, with no distinguishable residual grouping by tissue.

\subsection{Inference Speed and Latency}
\label{sec:inference_speed}

We benchmark inference efficiency for C-LiTE and CytoFormer under matched hardware (a single NVIDIA A100 GPU) and precision, measuring parameter count, FLOPs, throughput and peak GPU memory. For both models, the batch size is set to the largest value that fits in available GPU memory, and throughput is reported at that batch size. ~\autoref{tab:C-LiTE_vs_cytoformer} summarises the comparison. While trailing CytoFormer by only 0.012 average macro-F1, C-LiTE uses $30\times$ fewer parameters, requires $16.2\times$ fewer FLOPs per input, achieves $21.1\times$ higher throughput and consumes $6.0\times$ less GPU memory. This gap is consistent with CytoFormer's encoder being initialised from UNI2-h, a 681M-parameter ViT-H architecture, and fine-tuned end-to-end with per-organ classification heads, whereas C-LiTE is initialised from a 22M-parameter ViT-5-S and pre-trained with self-supervision on cell crops. These results indicate that C-LiTE offers a substantially more favourable accuracy-efficiency trade-off than any other method achieving comparable downstream performance, making it considerably more practical for large-scale, cell-level analysis of WSIs.

\begin{table}[H]
\centering
\scriptsize
\caption{Efficiency comparison between C-LiTE  and CytoFormer. All measurements taken on an NVIDIA A100 GPU at each model's native input resolution; batch size is the largest fitting in available GPU memory.}
\label{tab:C-LiTE_vs_cytoformer}
\begin{tabular}{lccc}
\toprule
\textbf{Metric} & \textbf{C-LiTE } & \textbf{CytoFormer} & \textbf{Comparison} \\
\midrule
Params ↓        & 22.5M      & 681.56M     & 30$\times$ smaller  \\
FLOPs ↓         & 11.13G      & 180.39G     & 16.2$\times$ less ops  \\
Throughput ↑    & 1039.24/s   & 49.18/s     & 21.1$\times$ faster \\
Peak memory ↓  & 4451.52MB   & 26541.14MB  & 6.0$\times$ less memory consumption    \\
\bottomrule
\end{tabular}
\vspace{1pt}
\begin{flushleft}
\scriptsize
\textbf{Params}: the total number of trainable and non-trainable parameters, in millions.\\
\textbf{FLOPs}: the number of floating-point operations, in billions, per input patch.\\
\textbf{Throughput}: the number of input patches processed per second.\\
\textbf{Peak GPU memory}: the maximum GPU memory allocated during inference, measured in MB.
\end{flushleft}
\end{table}
\subsection{Ablation}
\label{sec:ablation_results}
We conducted a series of ablation experiments to investigate the effects of input crop resolution, patch size, and patch-embedding initialisation on the quality of the learned representations. All configurations were evaluated using the same linear-probing protocol and performance metrics described in \autoref{sec:experiments}.
\subsubsection{Ablation Configurations}
\label{sec:ablation_setup}

Throughout this section, configurations are denoted using the format Backbone / Patch Size / Crop Size, specifying the backbone architecture, patch-embedding dimensions, and input crop dimensions, respectively. All spatial dimensions are expressed in pixels. For example, ViT-S / Patch14 / Crop256 denotes a ViT-S backbone with a $14\times14$ px patch embedding and a $256\times256$ px input crop. Figure \ref{fig:ablation} provides a visual overview of the different configurations evaluated in the ablation study.

We first established a reference configuration using a standard ViT-S backbone with its native patch size of 14 and an input resolution of $256\times256$ px. The subsequent experiments employed ViT-5-S, the pre-trained backbone used in the final C-LiTE architecture. To assess the effect of input resolution, we evaluated three crop sizes ($224\times224$, $96\times96$, and $64\times64$ px), while keeping the patch size fixed at 16. This comparison enabled us to examine the effect of reducing the input resolution while maintaining a constant patch-embedding configuration.

We next investigated the effect of reducing the patch size to 4, while retaining the ViT-5-S backbone and $64\times64$ px input resolution. This configuration corresponds to the fine-grained tokenisation strategy adopted in C-LiTE. To evaluate the contribution of Moore-Penrose pseudoinverse-based patch-embedding initialisation, we compared two configurations that differed only in the initialisation of the patch projection: random initialisation (ViT-5-S / Patch4 / Crop64, w/o PI-PatchEmbed) and pseudoinverse-based transfer of the pre-trained patch-embedding weights (ViT-5-S / Patch4 / Crop64 + PI-PatchEmbed). This controlled comparison isolates the contribution of PI-PatchEmbed to representation quality. The latter configuration corresponds to the complete C-LiTE architecture.

The ablations indicate that C-LiTE's advantage stems primarily from fine patch granularity, with PI-PatchEmbed initialisation providing a further smaller gain. Three effects, consistent across benchmarks, support this conclusion.

Cropping to cell scale alone was detrimental. With the patch size fixed at 16, reducing the crop from 224 to 64 px, which leaves a grid of only $4 \times 4$ tokens, lowered macro-F1 on every benchmark where this configuration was evaluated. Reducing the patch size to 4 at the same crop reversed this loss: the resulting $16 \times 16$ token grid improved macro-F1 on all benchmarks where the comparison was available, and this was the largest single effect on all of them except PanNuke. The improvement is not merely a consequence of sequence length. Patch16/Crop224 produces a comparable number of tokens ($14 \times 14$), yet Patch4/Crop64 clearly outperformed both on CoNIC, MIDOG, and NuCLS. The key consideration, therefore, is the spatial resolution at which individual cells are represented. Finally, initialising the $4 \times 4$ patch projection with PI-PatchEmbed instead of random initialisation improved macro-F1 on five of the seven datasets and left it unchanged on PUMA. The strength of these effects varied considerably between benchmarks, as described below.

The effects were largest on CoNIC (\autoref{tab:CoNIC_ablation1}). Resizing the crops from 224 to 64 px at patch size 16 reduced macro-F1 from 0.654 to 0.567, the largest drop observed on any benchmark. Reducing the patch size to 4 more than recovered this loss, raising macro-F1 to 0.701, and PI-PatchEmbed initialisation added a further 0.017, bringing C-LiTE to 0.718.
\begin{table}[H]
\centering
\captionsetup{width=\textwidth}
\caption{Ablation over patch size, crop size and patch-embedding initialisation on CoNIC~\citep{graham2021lizard}. Best value per column in \textbf{bold}. C-LiTE rows match ~\autoref{tab:CoNIC_main}.}
\label{tab:CoNIC_ablation1}
\resizebox{\textwidth}{!}{%
\begin{tabular}{lcccccccc}
\toprule
\textbf{Configuration} & \textbf{Patch} & \textbf{Crop} & \textbf{Acc.} & \textbf{Bal. Acc.} & \textbf{AUROC} & \textbf{mAP} & \textbf{Macro F1} & \textbf{Weighted F1} \\
\midrule
ViT-S / Patch14 / Crop256 & 14 & 256 & 0.748 & 0.621 & 0.943 & 0.712 & 0.650 & 0.779 \\
\midrule
Patch16 / Crop224 & 16 & 224 & 0.787 & 0.623 & 0.946 & 0.714 & 0.654 & 0.782 \\
Patch16 / Crop96 & 16 & 96 & 0.785 & 0.609 & 0.947 & 0.708 & 0.642 & 0.779 \\
Patch16 / Crop64 & 16 & 64 & 0.761 & 0.531 & 0.936 & 0.623 & 0.567 & 0.752 \\
\midrule
Patch4 / Crop64 (w/o PI-resize) & 4 & 64 & 0.802 & 0.673 & 0.953 & 0.768 & 0.701 & 0.799 \\
Patch4 / Crop64 + PI-PatchEmbed (C-LiTE) & 4 & 64 & 0.813 & \textbf{0.693} & 0.958 & \textbf{0.789} & \textbf{0.718} & 0.810 \\
\midrule
Patch4 / Crop64 (C-LiTE$_V^*$) & 4 & 64 & 0.825 & 0.652 & 0.961 & 0.746 & 0.683 & 0.820 \\
Patch4 / Crop64 (C-LiTE$_U^*$) & 4 & 64 & \textbf{0.827} & 0.652 & \textbf{0.962} & 0.746 & 0.685 & \textbf{0.822} \\
\bottomrule
\end{tabular}%
}
\end{table}

MONKEY dataset showed a same pattern with smaller magnitudes (\autoref{tab:monkey_ablation1}). The crop reduction lowered macro-F1 from 0.779 to 0.737, and finer patches more than recovered this loss (0.782). PI-PatchEmbed contributed only 0.003, so almost all of the improvement on this benchmark came from the finer patches.

\begin{table}[H]
\centering
\captionsetup{width=\textwidth}
\caption{Ablation over patch size, crop size and patch-embedding initialisation on MONKEY~\citep{Van2026monkey}. Best value per column in \textbf{bold}. C-LiTE rows match ~\autoref{tab:monkey_main}.}
\label{tab:monkey_ablation1}
\resizebox{\textwidth}{!}{%
\begin{tabular}{lcccccccc}
\toprule
\textbf{Configuration} & \textbf{Patch} & \textbf{Crop} & \textbf{Acc.} & \textbf{Bal. Acc.} & \textbf{AUROC} & \textbf{mAP} & \textbf{Macro F1} & \textbf{Weighted F1} \\
\midrule
ViT-S / Patch14 / Crop256 & 14 & 256 & 0.793 & 0.750 & 0.858 & 0.755 & 0.760 & 0.788 \\
\midrule
Patch16 / Crop224 & 16 & 224 & 0.808 & 0.769 & 0.873 & 0.783 & 0.779 & 0.803 \\
Patch16 / Crop96 & 16 & 96 & 0.804 & 0.763 & 0.866 & 0.767 & 0.773 & 0.799 \\
Patch16 / Crop64 & 16 & 64 & 0.775 & 0.727 & 0.835 & 0.715 & 0.737 & 0.768 \\
\midrule
Patch4 / Crop64 (w/o PI-resize) & 4 & 64 & 0.810 & 0.772 & 0.877 & 0.784 & 0.782 & 0.806 \\
Patch4 / Crop64 + PI-PatchEmbed (C-LiTE) & 4 & 64 & \textbf{0.812} & 0.776 & \textbf{0.879} & 0.788 & \textbf{0.785} & \textbf{0.808} \\
\midrule
Patch4 / Crop64 (C-LiTE$_V^*$) & 4 & 64 & 0.802 & 0.767 & 0.867 & 0.778 & 0.774 & 0.799 \\
Patch4 / Crop64 (C-LiTE$_U^*$) & 4 & 64 & 0.810 & \textbf{0.778} & 0.876 & \textbf{0.792} & 0.784 & 0.807 \\
\bottomrule
\end{tabular}%
}
\end{table}

On MIDOG (\autoref{tab:midog_ablation1}), finer patches raised macro-F1 from 0.679 to 0.781, 0.045 above the best patch-16 configuration. With PI-PatchEmbed, C-LiTE reached 0.790, the highest value of any configuration on this benchmark, including the distilled variants.

\begin{table}[H]
\centering
\captionsetup{width=\textwidth}
\caption{Ablation over patch size, crop size and patch-embedding initialisation on MIDOG~\citep{aubreville2026mitosisdetectionwildmultitumor}. Best value per column in \textbf{bold}. C-LiTE rows match ~\autoref{tab:midog_main}.}
\label{tab:midog_ablation1}
\resizebox{\textwidth}{!}{%
\begin{tabular}{lcccccccc}
\toprule
\textbf{Configuration} & \textbf{Patch} & \textbf{Crop} & \textbf{Acc.} & \textbf{Bal. Acc.} & \textbf{AUROC} & \textbf{mAP} & \textbf{Macro F1} & \textbf{Weighted F1} \\
\midrule
ViT-S / Patch14 / Crop256 & 14 & 256 & 0.847 & 0.710 & 0.857 & 0.948 & 0.737 & 0.835 \\
\midrule
Patch16 / Crop224 & 16 & 224 & 0.848 & 0.708 & 0.846 & 0.944 & 0.736 & 0.835 \\
Patch16 / Crop96 & 16 & 96 & 0.847 & 0.708 & 0.824 & 0.932 & 0.736 & 0.834 \\
Patch16 / Crop64 & 16 & 64 & 0.825 & 0.653 & 0.793 & 0.921 & 0.679 & 0.803 \\
\midrule
Patch4 / Crop64 (w/o PI-resize) & 4 & 64 & 0.865 & 0.755 & 0.894 & 0.960 & 0.781 & 0.860 \\
Patch4 / Crop64 + PI-PatchEmbed (C-LiTE) & 4 & 64 & \textbf{0.871} & \textbf{0.768} & \textbf{0.902} & \textbf{0.967} & \textbf{0.790} & \textbf{0.865} \\
\midrule
Patch4 / Crop64 (C-LiTE$_V^*$) & 4 & 64 & 0.870 & 0.762 & 0.892 & 0.963 & 0.786 & 0.863 \\
Patch4 / Crop64 (C-LiTE$_U^*$) & 4 & 64 & 0.865 & 0.748 & 0.885 & 0.961 & 0.775 & 0.857 \\
\bottomrule
\end{tabular}%
}
\end{table}

PanNuke was the exception (\autoref{tab:pannuke_ablation1}). All non-distilled configurations lay within a narrow range (0.698--0.713), and the patch-size reduction alone gained only 0.004, leaving the randomly initialised patch-4 model below Patch16/Crop224 (0.702 vs.\ 0.708). Only with PI-PatchEmbed (0.713) did it surpass all patch-16 configurations and the ViT-S reference. For PanNuke's five broad classes, which are largely separable at coarse resolution, the pre-trained patch projection thus mattered more than patch granularity.

\begin{table}[H]
\centering
\captionsetup{width=\textwidth}
\caption{Ablation over patch size, crop size and patch-embedding initialisation on PanNuke~\citep{gamper2020pannuke}. Best value per column in \textbf{bold}. C-LiTE rows match ~\autoref{tab:pannuke_main}.}
\label{tab:pannuke_ablation1}
\resizebox{\textwidth}{!}{%
\begin{tabular}{lcccccccc}
\toprule
\textbf{Configuration} & \textbf{Patch} & \textbf{Crop} & \textbf{Acc.} & \textbf{Bal. Acc.} & \textbf{AUROC} & \textbf{mAP} & \textbf{Macro F1} & \textbf{Weighted F1} \\
\midrule
ViT-S / Patch14 / Crop256 & 14 & 256 & 0.755 & 0.681 & 0.923 & 0.769 & 0.707 & 0.750 \\
\midrule
Patch16 / Crop224 & 16 & 224 & 0.760 & 0.688 & 0.925 & 0.770 & 0.708 & 0.755 \\
Patch16 / Crop96 & 16 & 96 & 0.760 & 0.686 & 0.923 & 0.769 & 0.705 & 0.754 \\
Patch16 / Crop64 & 16 & 64 & 0.754 & 0.681 & 0.921 & 0.763 & 0.698 & 0.752 \\
\midrule
Patch4 / Crop64 (w/o PI-resize) & 4 & 64 & 0.761 & 0.684 & 0.921 & 0.776 & 0.702 & 0.754 \\
Patch4 / Crop64 + PI-PatchEmbed (C-LiTE) & 4 & 64 & 0.764 & 0.697 & 0.926 & 0.781 & 0.713 & 0.761 \\
\midrule
Patch4 / Crop64 (C-LiTE$_V^*$) & 4 & 64 & 0.758 & 0.710 & 0.923 & 0.795 & 0.723 & 0.756 \\
Patch4 / Crop64 (C-LiTE$_U^*$) & 4 & 64 & \textbf{0.774} & \textbf{0.734} & \textbf{0.933} & \textbf{0.819} & \textbf{0.743} & \textbf{0.772} \\
\bottomrule
\end{tabular}%
}
\end{table}
NuCLS, by contrast, benefited strongly from both factors (\autoref{tab:nucls_ablation1}). Macro-F1 was nearly flat across the patch-16 configurations (0.437--0.456) but rose to 0.564 with patch size 4. PI-PatchEmbed then produced its largest gain of any benchmark, lifting macro-F1 to 0.590 and balanced accuracy from 0.559 to 0.573.

\begin{table}[H]
\centering
\captionsetup{width=\textwidth}
\caption{Ablation over patch size, crop size and patch-embedding initialisation on NuCLS~\citep{amgad2022nucls}. Best value per column in \textbf{bold}. C-LiTE rows match ~\autoref{tab:nucls_main}.}
\label{tab:nucls_ablation1}
\resizebox{\textwidth}{!}{%
\begin{tabular}{lcccccccc}
\toprule
\textbf{Configuration} & \textbf{Patch} & \textbf{Crop} & \textbf{Acc.} & \textbf{Bal. Acc.} & \textbf{AUROC} & \textbf{mAP} & \textbf{Macro F1} & \textbf{Weighted F1} \\
\midrule
ViT-S / Patch14 / Crop256 & 14 & 256 & 0.671 & 0.537 & 0.869 & 0.588 & 0.443 & 0.648 \\
\midrule
Patch16 / Crop224 & 16 & 224 & 0.668 & 0.543 & 0.868 & 0.589 & 0.449 & 0.650 \\
Patch16 / Crop96 & 16 & 96 & 0.664 & 0.546 & 0.867 & 0.589 & 0.456 & 0.658 \\
Patch16 / Crop64 & 16 & 64 & 0.656 & 0.531 & 0.867 & 0.585 & 0.437 & 0.641 \\
\midrule
Patch4 / Crop64 (w/o PI-resize) & 4 & 64 & 0.704 & 0.559 & 0.890 & 0.608 & 0.564 & 0.695 \\
Patch4 / Crop64 + PI-PatchEmbed (C-LiTE) & 4 & 64 & 0.736 & 0.573 & 0.917 & 0.633 & 0.590 & 0.722 \\
\midrule
Patch4 / Crop64 (C-LiTE$_V^*$) & 4 & 64 & 0.725 & 0.544 & 0.909 & 0.606 & 0.559 & 0.709 \\
Patch4 / Crop64 (C-LiTE$_U^*$) & 4 & 64 & \textbf{0.742} & \textbf{0.596} & \textbf{0.920} & \textbf{0.662} & \textbf{0.613} & \textbf{0.728} \\
\bottomrule
\end{tabular}%
}
\end{table}

On PUMA (\autoref{tab:puma_ablation1}), where Patch16/Crop64 was not evaluated, differences between configurations were small. The patch-4 configuration (0.540) outperformed Patch16/Crop96 (0.534) and Patch16/Crop224 (0.526), and PI-PatchEmbed left macro-F1 unchanged while improving mAP from 0.573 to 0.585.

\begin{table}[H]
\centering
\captionsetup{width=\textwidth}
\caption{Ablation over patch size, crop size and patch-embedding initialisation on PUMA~\citep{schuiveling_associations_2027}. Best value per column in \textbf{bold}. C-LiTE rows match ~\autoref{tab:puma_main}.}
\label{tab:puma_ablation1}
\resizebox{\textwidth}{!}{%
\begin{tabular}{lcccccccc}
\toprule
\textbf{Configuration} & \textbf{Patch} & \textbf{Crop} & \textbf{Acc.} & \textbf{Bal. Acc.} & \textbf{AUROC} & \textbf{mAP} & \textbf{Macro F1} & \textbf{Weighted F1} \\
\midrule
ViT-S / Patch14 / Crop256 & 14 & 256 & 0.778 & 0.478 & 0.928 & 0.550 & 0.513 & 0.761 \\
\midrule
Patch16 / Crop224 & 16 & 224 & 0.781 & 0.485 & 0.926 & 0.571 & 0.526 & 0.765 \\
Patch16 / Crop96 & 16 & 96 & 0.780 & 0.490 & 0.926 & 0.552 & 0.534 & 0.764 \\

\midrule
Patch4 / Crop64 (w/o PI-resize) & 4 & 64 & 0.784 & 0.501 & 0.925 & 0.573 & 0.540 & 0.767 \\
Patch4 / Crop64 + PI-PatchEmbed (C-LiTE) & 4 & 64 & 0.784 & 0.503 & 0.927 & 0.585 & 0.540 & 0.767 \\
\midrule
Patch4 / Crop64 (C-LiTE$_V^*$) & 4 & 64 & 0.769 & 0.446 & 0.916 & 0.519 & 0.492 & 0.748 \\
Patch4 / Crop64 (C-LiTE$_U^*$) & 4 & 64 & \textbf{0.794} & \textbf{0.516} & \textbf{0.936} & \textbf{0.590} & \textbf{0.557} & \textbf{0.781} \\
\bottomrule
\end{tabular}%
}
\end{table}

Across these six benchmarks, distillation did not consistently improve on C-LiTE. C-LiTE$_U^*$ achieved higher macro-F1 on PanNuke (0.743 vs.\ 0.713), NuCLS (0.613 vs.\ 0.590) and PUMA (0.557 vs.\ 0.540) but lower macro-F1 on CoNIC, MONKEY and MIDOG, whereas C-LiTE$_V^*$ improved only on PanNuke (0.723) and declined most on PUMA (0.492).

Overall, the full C-LiTE configuration was the best non-distilled configuration on five of the seven datasets and tied for best on PUMA, confirming that both design choices contribute, with their relative importance depending on the benchmark.

\section{Discussion and Conclusions}
\label{sec:discussion}

Characterising individual cells at scale is central to computational pathology. Nuclear morphology, mitotic activity and immune composition all inform diagnosis, prognosis and treatment decisions. Yet a single WSI can contain thousands of nuclei, and each must be represented before it can be analysed. Although existing foundation models provide transferable representations, most were developed for tissue-level inputs and can be computationally expensive when applied at the cellular level. This raises an important question: are large foundation models necessary to obtain high-quality cellular representations, or can a more compact architecture, specifically designed for cell-level analysis, achieve comparable performance?.

In this paper, we have proposed C-LiTE, a light-weight cell-centric foundation model. C-LiTE adapts ViT-5-S to $64\times64$ px cell crops through a $4\times4$ patch projection initialised with PI-PatchEmbed for projecting pre-trained model weights during initialisation.
It is pre-trained via SSL with DINO and iBOT objectives on approximately 60M cells from 19 cancer types. We have shown that C-LiTE performs comparably to the largest cell-level model while using $30\times$ fewer parameters. On average, it also outperforms the tissue-level foundation models evaluated. This suggests that, at the cellular scale, adapting the input representation and model architecture to the characteristics of individual cells may be more consequential than increasing model capacity.

The patch size determines how much of each cell a single token represents. At 0.5~mpp, a $16\times16$ patch spans $8~\mu$m, roughly the diameter of a nucleus. Each token then summarises a whole nucleus, leaving no room to model its internal structure. A $4\times4$ patch spans $2~\mu$m at this scale; the model can capture chromatin texture and membrane contour, the features pathologists use to separate, for example, atypical from normal mitoses. Our ablations support this view, as the finer patch outperformed configurations with similar or larger token grids. Fine spatial granularity, rather than a longer token sequence, is therefore the key design choice at cellular resolution. The same reasoning may explain why LEMON, which represents each nucleus with only 25 tokens, trailed on the more fine-grained benchmarks. It also exposes a weakness of upsampling cell crops to $224\times224$ px for tissue-level models, which raises computational cost but adds no information. PI-PatchEmbed makes native-resolution processing practical. It transfers pre-trained filters to the finer patch size, so the patch projection does not have to be learned from scratch.

Resolving a cell finely determines how well a model can see it. What the model needs to see depends on the task. For some tasks, cell identity lies in the cell itself. Mitoses, lymphocytes and monocytes fall into this group, and cell-level models were clearly favoured on them. For other tasks, classes reflect the surrounding tissue, as in the broad NuCLS categories. Patch-level models were competitive here, because their features link each cell to its micro-environment. For a third group, identity is defined by molecular markers. CD4$^{+}$ and CD8$^{+}$ T cells, for instance, are nearly indistinguishable in H\&E. CytoFormer's transcriptomics-derived supervision gives it an advantage on such cells that a morphology-only model may never fully close. Still, C-LiTE produces results close to CytoFormer on Immunocto, which suggests that much of this molecular signal leaves a visible trace in morphology. These three sources of identity, cellular, contextual and molecular, provide a useful lens for choosing a model for a given task.

The same perspective helps explain the mixed effects of distillation. By transferring a tissue-level teacher's representations to the student, distillation can provide contextual information that complements cell morphology. Such cues are particularly valuable for distinguishing minority classes whose nuclear appearances overlap with those of more prevalent classes as well as for identifying cells through their spatial relationships and surrounding tissue environment. This may explain why distillation improved minority-class recognition on benchmarks with broad nuclear categories, where tissue context provides useful additional evidence for classification. On the fine-grained inflammatory subtypes of CoNIC, which depend on subtle cellular detail \cite{laviron2022tumor, pmlr-v254-pochet24a}, it reduced performance.
The teachers saw upsampled crops, so the students may inherit tissue-scale biases that pre-training at native resolution does not fully remove. The students nonetheless often outperformed their much larger teachers. Cell-level pre-training thus appears to adapt the teacher's representation to the cellular scale rather than merely compressing it. In practice, C-LiTE is the more robust default, while C-LiTE$_U^*$ may suit tasks with broad categories and severe class imbalance.

% Accuracy is only useful if a model can also be deployed widely. 

C-LiTE embeds one million cells in approximately 16~minutes on a single A100 GPU (\autoref{sec:inference_speed}), compared with about 5.6~hours for CytoFormer. It must be noted that absolute timings depend on hardware, batch size and implementation. The relative difference reflects the gap in computational cost and should hold across settings. C-LiTE can represent each detected cell with a general-purpose embedding, enabling label-free phenotyping, morphological profiling and spatial analysis. Reducing inference cost in this way also aligns with the Green AI view that efficiency should be an explicit objective alongside accuracy \citep{schwartz2020green}. Wide deployment also depends on generalisation beyond the pre-training data. Here, the pre-training corpus is a further contribution of this work. To our knowledge, it is one of the largest and most diverse collections of cell-centred crops assembled to date.

We release the corpus and models so that this can be tested directly. These findings should be read in light of several limitations, each of which points to future work. First, all models were evaluated by linear probing. Performance under fine-tuning, or within detection and segmentation pipelines, remains to be tested. Second, pre-training used HoVer-Net detections at a single resolution of 0.5~mpp; therefore, robustness to detection noise and higher-resolution data was not assessed. Multi-resolution pre-training could address the latter. Third, the corpus was limited to TCGA H\&E slides, and behaviour on other stains and on immunohistochemistry remains untested. Fourth, the pseudoinverse provides the minimum-norm least-squares solution, yielding a smaller spatial filter whose projection into the original spatial domain minimises the squared reconstruction error relative to the pretrained filter. This provides an approximate transfer of the pretrained spatial structure rather than guaranteeing an exact preservation of the original patch embeddings. Finally, the ablations did not vary the pre-training data. Beyond these limitations, adding molecular supervision to a lightweight backbone could combine the strengths of C-LiTE and CytoFormer at practical cost.

Overall, this study shows that strong cellular representations do not require large models. Fine spatial granularity and diverse pre-training data were shown to be more important than parameter count. C-LiTE offers a practical and efficient foundation for cell-level analysis across whole slides and cohorts.

\section*{Code and Data availability}
The source code for C-LiTE is publicly available at \url{https://github.com/eshasadia/C-LiTE}. An example notebook demonstrating its application has also been incorporated into TIA Toolbox (\url{https://github.com/tissueimageanalytics/tiatoolbox}). The pre-trained model weights are available through Hugging Face at \url{https://huggingface.co/eshasadia/C-LiTE}.
The pre-training corpus was derived from The Cancer Genome Atlas (TCGA), with WSIs retrieved from the NCI Imaging Data Commons (\url{https://portal.imaging.datacommons.cancer.gov}). The generated cell-level corpus is publicly available through the EMBL-EBI BioStudies database under accession number S-BSST3558 (\url{https://www.ebi.ac.uk/biostudies/studies/S-BSST3558}).
\section*{Declaration of Generative AI }

During the preparation of this work the author(s) used ChatGPT in order to improve readability and language.

\section*{Declaration of competing interest}
The authors declare that they have no known competing financial interests or personal relationships that could have appeared to influence the work reported in this paper.

\bibliographystyle{elsarticle-harv}  
\bibliography{references.bib}
\clearpage
\appendix
\setcounter{table}{0}
\setcounter{figure}{0}
\section{Supplementary Information}

\label{sec:supplementary}

\renewcommand{\thetable}{A.\arabic{table}}

\subsection{Benchmark composition}
\label{sec:supp_datasets}
The seven benchmark datasets used for evaluation are summarised in \autoref{tab:benchmark_summary}. They cover mitosis classification, nucleus classification and immune-cell phenotyping, and include one non-H\&E benchmark (MONKEY, PAS stain). Their class distributions are given in~\autoref{tab:midog_data}--\autoref{tab:puma_data}. All benchmarks are imbalanced, and several contain classes that make up less than 2\% of cells, including dead cells in PanNuke, eosinophils and neutrophils in CoNIC, and neutrophils, melanophage and plasma cells in PUMA. This imbalance motivated the use of macro-F1 and balanced accuracy as the primary metrics (\autoref{sec:eval_metrics}). The mitosis benchmark was compiled from five public datasets to increase variation in tissue type, scanner and institution (\autoref{tab:midog_data}). For NuCLS, we used the super-level taxonomy described in~\autoref{sec:benchmark_datasets}.

\begin{table}[H]
\centering
\small
\caption{Downstream benchmarks used for linear-probe evaluation. Cell counts are the number of labelled cells used for evaluation. CV stands for cross-validation.}
\label{tab:benchmark_summary}
\resizebox{\textwidth}{!}{%
\begin{tabular}{llcrrl}
\toprule
\textbf{Benchmark} & \textbf{Task} & \textbf{Stain} & \textbf{Classes} & \textbf{Cells} & \textbf{Evaluation} \\
\midrule
MIDOG~\citep{aubreville2026mitosisdetectionwildmultitumor} & Atypical vs.\ normal mitosis & H\&E & 2 & 22,746 & 5-fold CV \\
NuCLS~\citep{amgad2022nucls} & Nuclear classification (super level) & H\&E & 5 & 59,485 & 5-fold CV \\
PanNuke~\citep{gamper2020pannuke} & Pan-cancer nucleus classification & H\&E & 5 & 187,259 & 3 official folds \\
CoNIC~\citep{graham2021lizard} & Colon nucleus classification & H\&E & 6 & 569,861 & 5-fold CV \\
MONKEY~\citep{Van2026monkey} & Lymphocyte vs.\ monocyte & PAS & 2 & 90,367 & 5-fold CV \\
Immunocto~\citep{simard_Immunocto_2025} & Immune-cell phenotyping & H\&E & 5 & 6,848,454 & Official train/test split \\
PUMA~\citep{schuiveling_associations_2027} & Melanoma nuclear phenotyping & H\&E & 10 & 97,378 & 5-fold CV \\
\bottomrule
\end{tabular}%
}
\end{table}

\begin{table}[H]
\centering
\footnotesize
\caption{Composition of the mitosis classification benchmark, compiled from five public datasets.}
\label{tab:midog_data}
\begin{tabular}{p{1.8cm} p{3cm} r r r p{4cm}}
\toprule
\textbf{No.} & \textbf{Dataset} & \textbf{Normal} & \textbf{Atypical} & \textbf{Total} & \textbf{Domains} \\
\midrule
1 & MIDOG & 10,189 & 1,749 & 11,938 & 9 \\
2 & TCGA Subtyping & 3,603 & 980 & 4,583 & - \\
3 & Ami-Br & 2,889 & 833 & 3,722 & Human Breast tissue \\
4 & OMG-Octo & 1,376 & 379 & 1,755 & - \\
5 & AtNorm-Br & 619 & 129 & 748 & TCGA Breast tissue \\
\midrule
\textbf{Overall Total} && 18,676 & 4,070 & 22,746 & - \\
\bottomrule
\end{tabular}
\end{table}

\begin{table}[H]
\centering
\small
\caption{Class distribution of the NuCLS (super level) benchmark, sorted by frequency.}
\label{tab:nucls_data}
\begin{tabular}{lrr}
\toprule
\textbf{Class} & \textbf{Count} & \textbf{Proportion (\%)} \\
\midrule
Tumour & 21,296 & 35.80 \\
sTIL & 19,192 & 32.26 \\
Non-TIL stromal & 10,486 & 17.63 \\
Ambiguous & 7,911 & 13.30 \\
Other nucleus & 600 & 1.01 \\
\midrule
\textbf{Total} & \textbf{59,485} & 100.00 \\
\bottomrule
\end{tabular}
\end{table}

\begin{table}[H]
\centering
\small
\caption{Class distribution of the PanNuke benchmark, sorted by frequency.}
\label{tab:pannuke_data}
\begin{tabular}{lrr}
\toprule
\textbf{Class} & \textbf{Count} & \textbf{Proportion (\%)} \\
\midrule
Neoplastic & 76,448 & 40.82 \\
Connective & 49,892 & 26.64 \\
Inflammatory & 31,846 & 17.01 \\
Epithelial & 26,230 & 14.01 \\
Dead & 2,843 & 1.52 \\
\midrule
\textbf{Total} & \textbf{187,259} & 100.00 \\
\bottomrule
\end{tabular}
\end{table}

\begin{table}[H]
\centering
\small
\caption{Class distribution of the CoNIC benchmark, sorted by frequency.}
\label{tab:CoNIC_data}
\begin{tabular}{lrr}
\toprule
\textbf{Class} & \textbf{Count} & \textbf{Proportion (\%)} \\
\midrule
Epithelial   & 282,082 & 49.50 \\
Connective   & 125,950 & 22.10 \\
Lymphocyte   & 120,933 & 21.22 \\
Plasma cell  & 31,965  & 5.61 \\
Neutrophil   & 5,082   & 0.89 \\
Eosinophil   & 3,849   & 0.68 \\

\midrule
\textbf{Total} & \textbf{569,861} & \textbf{100.00} \\
\bottomrule
\end{tabular}
\end{table}

\begin{table}[H]
\centering
\small
\caption{Class distribution of the MONKEY benchmark, sorted by frequency.}
\label{tab:monkey_data}
\begin{tabular}{lrr}
\toprule
\textbf{Class} & \textbf{Count} & \textbf{Proportion (\%)} \\
\midrule
Lymphocyte & 59,101 & 65.40 \\
Monocyte & 31,266 & 34.60 \\
\midrule
\textbf{Total} & \textbf{90,367} & 100.00 \\
\bottomrule
\end{tabular}
\end{table}

\begin{table}[H]
\centering
\small
\caption{Class distribution of the Immunocto benchmark, sorted by frequency.}
\label{tab:Immunocto_data}
\begin{tabular}{lrr}
\toprule
\textbf{Class} & \textbf{Count} & \textbf{Proportion (\%)} \\
\midrule
Other & 4,565,636 & 66.67 \\
CD4$^{+}$ T cell & 952,929 & 13.91 \\
B cell & 480,900 & 7.02 \\
CD8$^{+}$ T cell & 467,395 & 6.82 \\
Macrophage & 381,594 & 5.57 \\
\midrule
\textbf{Total} & \textbf{6,848,454} & 100.00 \\
\bottomrule
\end{tabular}
\end{table}

\begin{table}[H]
\centering
\small
\caption{Class distribution of the PUMA benchmark, sorted by frequency.}
\label{tab:puma_data}
\begin{tabular}{lrr}
\toprule
\textbf{Class} & \textbf{Count} & \textbf{Proportion (\%)} \\
\midrule
Tumour & 57,351 & 58.48 \\
Lymphocyte & 21,668 & 22.09 \\
Histiocyte & 7,866 & 8.02 \\
Stroma & 3,870 & 3.95 \\
Epithelium & 2,216 & 2.26 \\
Apoptosis & 1,814 & 1.85 \\
Endothelium & 1,698 & 1.73 \\
Plasma cell & 520 & 0.53 \\
Neutrophil & 375 & 0.38 \\
Melanophage & 692 & 0.71 \\
\midrule
\textbf{Total} & \textbf{98,070} & \textbf{100.00} \\
\bottomrule
\end{tabular}
\end{table}

\subsection{Pre-training configuration}
\label{sec:supp_config}
 The pre-training configurations for all three C-LiTE variants are summarised in \autoref{tab:training_config}. All models were trained on $64\times64$ px cell crops at 0.5~mpp. With a patch size of $4\times4$, each input was therefore represented by a $16\times16$ token grid. We used two global views per cell and omitted local crops, which are commonly employed in DINOv2. This choice reflects the nature of the input data: each crop already contains a single complete cell, such that further cropping would risk excluding part of the object of interest.
 \begin{figure}[H]
    \centering
    \includegraphics[width=1\textwidth]{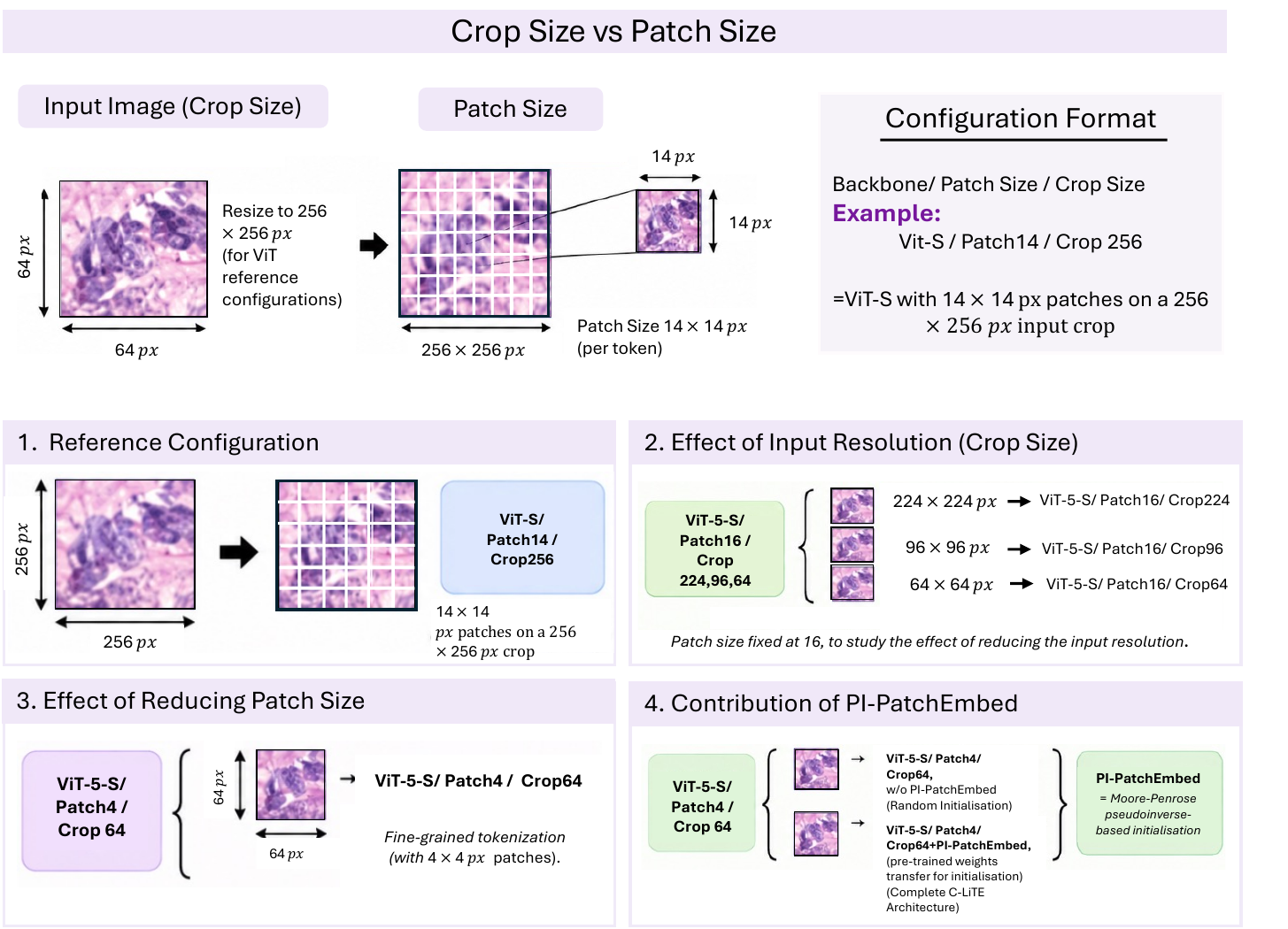}
    \captionsetup{width=\textwidth}
    \caption{Schematic illustration of crop size and patch size across the evaluated configurations. The original 64 $\times$ 64 px histopathology crop was resized to 256 $\times$ 256 px for the ViT-S reference configuration. Subsequent ViT-5-S experiments examined different crop resolutions, finer 4 $\times$ 4 px patch tokenisation, and the contribution of pseudoinverse-based patch-embedding initialisation (PI-PatchEmbed) to the final C-LiTE architecture.
    }
    \label{fig:ablation}
\end{figure}
\begin{table}[H]
\centering
\small
\caption{Self-supervised pre-training configuration of C-LiTE.}
\label{tab:training_config}
\begin{tabular}{ll}
\toprule
\textbf{Setting} & \textbf{Value} \\
\midrule
\multicolumn{2}{l}{\textit{Architecture and input}} \\
Backbone & ViT-5-S (22M parameters) \\
Input resolution & $64\times64$ px at 0.5~mpp \\
Patch size & $4\times4$ (initialised with PI-PatchEmbed) \\
Views & 2 global views; no local views \\
\midrule
\multicolumn{2}{l}{\textit{Data}} \\
Pre-training corpus & $\sim$60M cell crops, 19 cancer types \\
Sampling & Balanced across cancer types \\
\midrule
\multicolumn{2}{l}{\textit{Optimisation}} \\
Objectives & DINO, iBOT, KoLeo regulariser \\
Optimiser & AdamW \\
Base learning rate & $2.0\times10^{-4}$ \\
Weight decay & $0.04 \rightarrow 0.2$ \\
Epochs & 200 (20 warm-up) \\
Batch size & 256 per GPU (512 effective) \\
\midrule
\multicolumn{2}{l}{\textit{Self-distillation}} \\
Student temperature & 0.1 \\
Teacher temperature & 0.07 \\
Teacher EMA momentum & 0.994 \\
\midrule
\multicolumn{2}{l}{\textit{Compute}} \\
Hardware & 2 $\times$ NVIDIA A100 (80~GB) \\
Precision & FP16 mixed precision \\
\bottomrule
\end{tabular}
\end{table}
\subsection{Distillation configuration}
\label{sec:supp_distill}
 
We initialised C-LiTE$_U^*$ and C-LiTE$_V^*$ using ViT-5-S student models distilled from UNI2 \cite{chen2024towards} and Virchow2 \cite{zimmermann_virchow2_2024}, respectively, prior to self-supervised pre-training (\autoref{tab:distill_config}). During distillation, the frozen teacher and student received the same augmented view of each cell, and the student was trained to reproduce the teacher's embedding without supervision from class labels. As both teacher models require inputs of size $224\times224$, cell crops were resized to this resolution exclusively during distillation. The linear projection used to align the teacher and student embedding dimensions was subsequently discarded.
\begin{table}[H]
\centering
\small
\caption{Knowledge distillation configuration used to initialise C-LiTE$_U^*$ and C-LiTE$_V^*$.}
\label{tab:distill_config}
\begin{tabular}{ll}
\toprule
\textbf{Setting} & \textbf{Value} \\
\midrule
Teachers (frozen) & UNI2 ($d_t=1536$); Virchow2 ($d_t=1280$) \\
Student & ViT-5-S/16 ($d_s=384$) \\
Training data & 19M cell crops (1M per cancer type) \\
Input & $64\times64$ crops resized to $224\times224$ \\
Loss & Mean squared error + cosine distance \\
\bottomrule
\end{tabular}

\end{table}
\subsection{Per-class evaluation}
\label{sec:supp_perclass}
 
Per-class precision, recall, and F1 scores were derived from the same linear probes described in \autoref{sec:results}, with each class treated in turn as the positive class. The results are presented in \autoref{tab:midog_perclass_prf1_main}--\autoref{tab:puma_perclass_prf1_main}, following the order in which the benchmark datasets are introduced in the main text.

Two consistent patterns emerge across the benchmarks. First, performance is generally strong and relatively consistent across models for the most prevalent classes, whereas inter-model differences are more pronounced for the less frequent classes identified in \autoref{sec:supp_datasets}. For instance, on MIDOG, the F1 score for normal mitoses varies by only 0.02 across models, compared with a variation of 0.16 for atypical mitoses. Second, classification errors for the rarest immune cell populations are primarily characterised by low recall. In Immunocto, recall for CD8$^{+}$ T cells and macrophages remains below 0.18 and 0.35, respectively, while precision is substantially higher. This disparity between precision and recall suggests that these populations are predominantly under-detected, with relatively few false-positive predictions, rather than frequently misclassified as other cell types.

\begin{table}[H]
\centering
\captionsetup{width=\textwidth}
\caption{Per-class precision (P), recall (R) and F1 on MIDOG (atypical versus normal mitosis classification). Values are mean (standard deviation) over five-fold cross-validation. The best value in each column is shown in \textbf{bold} and the second best is \underline{underlined}.}

\label{tab:midog_perclass_prf1_main}
\resizebox{\textwidth}{!}{%
\begin{tabular}{l ccc ccc}
\toprule
& \multicolumn{3}{c}{\textbf{Atypical}} & \multicolumn{3}{c}{\textbf{Normal}} \\
\cmidrule(lr){2-4}\cmidrule(lr){5-7}
\textbf{Model} & P & R & F1 & P & R & F1 \\
\midrule
DINOv2 (ViT-S/14) & 0.697 (0.014) & 0.389 (0.022) & 0.499 (0.021) & 0.853 (0.004) & 0.954 (0.002) & 0.901 (0.003) \\
DINOv2 (ViT-B/14) & 0.697 (0.013) & 0.422 (0.009) & 0.525 (0.010) & 0.859 (0.002) & 0.951 (0.002) & 0.902 (0.002) \\
DINOv2 (ViT-L/14)  \citep{oquab_dinov2_2023} & 0.678 (0.011) & 0.442 (0.009) & 0.536 (0.009) & 0.862 (0.002) & 0.943 (0.002) & 0.901 (0.002) \\
\midrule
Phikon \citep{filiot_phikon-v2_2024}   & 0.680 (0.019) & 0.445 (0.018) & 0.538 (0.018) & 0.863 (0.004) & 0.944 (0.003) & 0.902 (0.003) \\
Virchow2 \citep{zimmermann_virchow2_2024} & 0.667 (0.026) & 0.478 (0.011) & 0.557 (0.016) & 0.869 (0.003) & 0.935 (0.006) & 0.901 (0.004) \\
UNI2  \citep{chen_towards_2024}      & 0.647 (0.014) & 0.485 (0.014) & 0.554 (0.014) & 0.870 (0.003) & 0.929 (0.003) & 0.898 (0.003) \\
\midrule
DinoBloom  \citep{linguraru_DinoBloom_2024} & 0.727 (0.022) & 0.514 (0.011) & 0.602 (0.013) & 0.878 (0.003) & 0.948 (0.005) & 0.912 (0.004) \\
LEMON  \citep{chadoutaud2026lemon}    & 0.749 (0.017) & 0.495 (0.010) & 0.596 (0.009) & 0.875 (0.002) & \textbf{0.955 (0.004)} & 0.913 (0.002) \\
CytoFormer \citep{yao2026cytoformer} & 0.691 (0.011) & 0.539 (0.011) & 0.606 (0.009) & 0.883 (0.002) & 0.935 (0.003) & 0.908 (0.002) \\
\midrule
\textbf{C-LiTE } & \underline{0.752 (0.013)} & \textbf{0.589 (0.011)} & \textbf{0.660 (0.010)} & \textbf{0.895 (0.003)} & 0.948 (0.004) & \textbf{0.921 (0.002)} \\
\textbf{C-LiTE$_V^*$} & \textbf{0.756 (0.011)} & \underline{0.575 (0.016)} & \underline{0.653 (0.013)} & \underline{0.892 (0.004)} & 0.950 (0.003) & \underline{0.920 (0.002)} \\
\textbf{C-LiTE$_U^*$} & \underline{0.752 (0.019)} & 0.545 (0.014) & 0.632 (0.015) & 0.886 (0.003) & \underline{0.951 (0.004)} & 0.917 (0.003) \\
\bottomrule
\end{tabular}%
}
\end{table}

\begin{table}[H]
\centering
\captionsetup{width=\textwidth}
\caption{Per-class precision (P), recall (R) and F1 on NuCLS (super level) (nuclear classification). Values are mean (standard deviation) over five-fold cross-validation. The best value in each column is shown in \textbf{bold} and the second best is \underline{underlined}.}

\label{tab:nucls_perclass_prf1_main}

\hspace*{-0.03\textwidth}\resizebox{1.06\textwidth}{!}{%
\begin{tabular}{l ccc ccc}
\toprule
& \multicolumn{3}{c}{\textbf{Ambiguous}} & \multicolumn{3}{c}{\textbf{non-TIL Stromal}} \\
\cmidrule(lr){2-4}\cmidrule(lr){5-7}
\textbf{Model} & P & R & F1 & P & R & F1 \\
\midrule
DINOv2 (ViT-S/14) & 0.397 (0.017) & 0.158 (0.012) & 0.226 (0.015) & 0.556 (0.007) & 0.457 (0.005) & 0.502 (0.006) \\
DINOv2 (ViT-B/14) & 0.375 (0.015) & 0.172 (0.011) & 0.236 (0.014) & 0.567 (0.007) & 0.470 (0.010) & 0.514 (0.008) \\
DINOv2 (ViT-L/14)  \citep{oquab_dinov2_2023} & 0.373 (0.013) & 0.186 (0.010) & 0.248 (0.012) & 0.565 (0.003) & 0.475 (0.002) & 0.516 (0.002) \\
\midrule
Phikon  \citep{filiot_phikon-v2_2024}  & 0.427 (0.012) & 0.240 (0.009) & 0.307 (0.009) & 0.587 (0.008) & 0.503 (0.009) & 0.542 (0.006) \\
Virchow2 \citep{zimmermann_virchow2_2024} & 0.444 (0.010) & 0.277 (0.008) & 0.341 (0.008) & 0.604 (0.004) & 0.527 (0.008) & 0.563 (0.005) \\
UNI2  \citep{chen_towards_2024}      & 0.430 (0.011) & 0.281 (0.011) & 0.339 (0.010) & 0.597 (0.010) & 0.534 (0.011) & 0.563 (0.008) \\
\midrule
DinoBloom  \citep{linguraru_DinoBloom_2024} & 0.420 (0.008) & 0.206 (0.006) & 0.276 (0.007) & 0.576 (0.010) & 0.486 (0.008) & 0.528 (0.009) \\
LEMON \citep{chadoutaud2026lemon}     & 0.495 (0.017) & 0.309 (0.009) & 0.381 (0.010) & 0.615 (0.010) & 0.520 (0.010) & 0.564 (0.010) \\
CytoFormer \citep{yao2026cytoformer} & 0.467 (0.005) & \underline{0.328 (0.006)} & 0.385 (0.006) & 0.598 (0.004) & 0.535 (0.004) & 0.565 (0.004) \\
\midrule
\textbf{C-LiTE} & \underline{0.498 (0.012)} & 0.326 (0.003) & \underline{0.394 (0.002)} & \underline{0.630 (0.008)} & \underline{0.548 (0.004)} & \underline{0.586 (0.004)} \\
\textbf{C-LiTE$_V^*$} & 0.481 (0.010) & 0.301 (0.003) & 0.370 (0.005) & 0.613 (0.005) & 0.532 (0.005) & 0.569 (0.001) \\
\textbf{C-LiTE$_U^*$} & \textbf{0.507 (0.015)} & \textbf{0.330 (0.004)} & \textbf{0.400 (0.006)} & \textbf{0.634 (0.009)} & \textbf{0.551 (0.009)} & \textbf{0.589 (0.009)} \\
\bottomrule
\end{tabular}%
}

\vspace{0.5em}

\hspace*{-0.03\textwidth}\resizebox{1.06\textwidth}{!}{%
\begin{tabular}{l ccc ccc}
\toprule
& \multicolumn{3}{c}{\textbf{sTIL}} & \multicolumn{3}{c}{\textbf{Tumour}} \\
\cmidrule(lr){2-4}\cmidrule(lr){5-7}
\textbf{Model} & P & R & F1 & P & R & F1 \\
\midrule
DINOv2 (ViT-S/14) & 0.709 (0.006) & 0.847 (0.004) & 0.772 (0.004) & 0.747 (0.002) & 0.862 (0.005) & 0.800 (0.003) \\
DINOv2 (ViT-B/14) & 0.715 (0.004) & 0.841 (0.004) & 0.773 (0.004) & 0.759 (0.005) & 0.865 (0.007) & 0.808 (0.005) \\
DINOv2 (ViT-L/14)  \citep{oquab_dinov2_2023} & 0.720 (0.005) & 0.845 (0.003) & 0.777 (0.003) & 0.768 (0.003) & 0.858 (0.003) & 0.811 (0.001) \\
\midrule
Phikon  \citep{filiot_phikon-v2_2024}  & 0.732 (0.005) & 0.843 (0.006) & 0.783 (0.005) & 0.788 (0.004) & 0.871 (0.003) & 0.827 (0.003) \\
Virchow2 \citep{zimmermann_virchow2_2024} & 0.746 (0.006) & 0.846 (0.003) & 0.793 (0.004) & 0.803 (0.004) & \underline{0.876 (0.005)} & \underline{0.838 (0.003)} \\
UNI2  \citep{chen_towards_2024}      & 0.755 (0.005) & 0.844 (0.006) & 0.797 (0.004) & \textbf{0.807 (0.004)} & 0.871 (0.004) & \underline{0.838 (0.003)} \\
\midrule
DinoBloom  \citep{linguraru_DinoBloom_2024} & 0.731 (0.004) & 0.853 (0.006) & 0.788 (0.003) & 0.770 (0.003) & 0.871 (0.003) & 0.818 (0.003) \\
LEMON \citep{chadoutaud2026lemon}     & 0.760 (0.006) & \underline{0.870 (0.005)} & 0.811 (0.004) & 0.783 (0.004) & 0.868 (0.004) & 0.824 (0.003) \\
CytoFormer \citep{yao2026cytoformer} & 0.772 (0.006) & 0.846 (0.004) & 0.808 (0.004) & 0.798 (0.005) & 0.865 (0.003) & 0.830 (0.003) \\
\midrule
\textbf{C-LiTE} & \underline{0.773 (0.008)} & 0.867 (0.009) & \underline{0.818 (0.003)} & 0.800 (0.005) & 0.874 (0.005) & 0.835 (0.002) \\
\textbf{C-LiTE$_V^*$} & 0.765 (0.005) & 0.869 (0.009) & 0.814 (0.002) & 0.791 (0.002) & 0.866 (0.003) & 0.827 (0.001) \\
\textbf{C-LiTE$_U^*$} & \textbf{0.778 (0.007)} & \textbf{0.873 (0.004)} & \textbf{0.822 (0.005)} & \underline{0.806 (0.005)} & \textbf{0.880 (0.002)} & \textbf{0.842 (0.004)} \\
\bottomrule
\end{tabular}%
}
\end{table}

\begin{table}[H]
\centering
\captionsetup{width=\textwidth}
\caption{Per-class precision (P), recall (R) and F1 on PanNuke (nucleus classification). Values are mean (standard deviation) over the three official folds. The best value in each column is shown in \textbf{bold} and the second best is \underline{underlined}.}
\label{tab:pannuke_perclass_prf1_main}

\hspace*{-0.03\textwidth}\resizebox{1.06\textwidth}{!}{%
\begin{tabular}{l ccc ccc ccc}
\toprule
& \multicolumn{3}{c}{\textbf{Connective}} & \multicolumn{3}{c}{\textbf{Dead}} & \multicolumn{3}{c}{\textbf{Epithelial}} \\
\cmidrule(lr){2-4}\cmidrule(lr){5-7}\cmidrule(lr){8-10}
\textbf{Model} & P & R & F1 & P & R & F1 & P & R & F1 \\
\midrule
DINOv2 (ViT-S/14) & 0.683 (0.006) & 0.640 (0.006) & 0.661 (0.003) & 0.615 (0.039) & 0.548 (0.065) & 0.578 (0.049) & 0.688 (0.014) & 0.559 (0.008) & 0.617 (0.007) \\
DINOv2 (ViT-B/14) & 0.676 (0.004) & 0.641 (0.005) & 0.658 (0.002) & 0.609 (0.042) & 0.536 (0.055) & 0.570 (0.048) & 0.690 (0.011) & 0.572 (0.012) & 0.626 (0.009) \\
DINOv2 (ViT-L/14)  \citep{oquab_dinov2_2023} & 0.681 (0.006) & 0.648 (0.005) & 0.664 (0.000) & 0.589 (0.052) & 0.524 (0.070) & 0.553 (0.058) & 0.691 (0.008) & 0.575 (0.012) & 0.627 (0.006) \\
\midrule
Phikon \citep{filiot_phikon-v2_2024}   & 0.710 (0.005) & 0.682 (0.005) & 0.696 (0.004) & 0.669 (0.055) & 0.598 (0.050) & 0.631 (0.049) & 0.735 (0.006) & 0.705 (0.015) & 0.720 (0.010) \\
Virchow2 \citep{zimmermann_virchow2_2024} & 0.730 (0.004) & 0.703 (0.005) & 0.716 (0.003) & 0.669 (0.063) & 0.590 (0.070) & 0.627 (0.067) & \textbf{0.766} (0.001) & \textbf{0.744} (0.011) & \underline{0.755} (0.006) \\
UNI2  \citep{chen_towards_2024}      & 0.731 (0.005) & 0.706 (0.001) & 0.718 (0.002) & 0.616 (0.054) & 0.575 (0.045) & 0.594 (0.045) & \underline{0.782} (0.002) & \textbf{0.744} (0.009) & \textbf{0.763} (0.005) \\
\midrule
DinoBloom  \citep{linguraru_DinoBloom_2024} & 0.696 (0.007) & 0.671 (0.003) & 0.683 (0.003) & 0.622 (0.062) & 0.536 (0.088) & 0.574 (0.074) & 0.712 (0.007) & 0.615 (0.009) & 0.660 (0.006) \\
LEMON  \citep{chadoutaud2026lemon}    & 0.706 (0.005) & 0.690 (0.002) & 0.698 (0.001) & 0.653 (0.055) & 0.461 (0.063) & 0.538 (0.053) & 0.686 (0.018) & 0.547 (0.003) & 0.608 (0.007) \\
CytoFormer \citep{yao2026cytoformer} & \underline{0.736} (0.002) & \underline{0.716} (0.003) & \underline{0.726} (0.002) & \underline{0.678} (0.054) & 0.590 (0.064) & 0.630 (0.058) & 0.756 (0.002) & 0.703 (0.010) & 0.729 (0.005) \\
\midrule
C-LiTE & 0.733 (0.004) & 0.710 (0.007) & 0.721 (0.006) & 0.668 (0.039) & 0.592 (0.028) & 0.627 (0.030) & 0.733 (0.015) & 0.632 (0.011) & 0.679 (0.012) \\
\textbf{C-LiTE$_V^*$} & 0.721 (0.005) & 0.706 (0.002) & 0.713 (0.003) & \underline{0.699} (0.023) & \underline{0.641} (0.023) & \underline{0.668} (0.020) & 0.750 (0.005) & 0.620 (0.005) & 0.679 (0.004) \\
\textbf{C-LiTE$_U^*$} & \textbf{0.740} (0.006) & \textbf{0.717} (0.004) & \textbf{0.728} (0.003) & \textbf{0.711} (0.022) & \textbf{0.687} (0.029) & \textbf{0.699} (0.023) & \textbf{0.766} (0.004) & 0.656 (0.006) & 0.707 (0.004) \\
\bottomrule
\end{tabular}%
}

\vspace{0.5em}

% ---- Panel 2 ----
\hspace*{-0.03\textwidth}\resizebox{1.06\textwidth}{!}{%
\begin{tabular}{l ccc ccc}
\toprule
& \multicolumn{3}{c}{\textbf{Inflammatory}} & \multicolumn{3}{c}{\textbf{Neoplastic}} \\
\cmidrule(lr){2-4}\cmidrule(lr){5-7}
\textbf{Model} & P & R & F1 & P & R & F1 \\
\midrule

DINOv2 (ViT-S/14) & 0.669 (0.004) & 0.657 (0.008) & 0.663 (0.006) & 0.746 (0.012) & 0.834 (0.009) & 0.787 (0.008) \\
DINOv2 (ViT-B/14) & 0.662 (0.004) & 0.654 (0.012) & 0.658 (0.007) & 0.751 (0.014) & 0.827 (0.012) & 0.787 (0.011) \\
DINOv2 (ViT-L/14)  \citep{oquab_dinov2_2023} & 0.661 (0.002) & 0.655 (0.012) & 0.658 (0.007) & 0.761 (0.013) & 0.834 (0.011) & 0.796 (0.009) \\
\midrule
Phikon \citep{filiot_phikon-v2_2024}   & 0.700 (0.007) & 0.674 (0.002) & 0.687 (0.004) & 0.808 (0.008) & 0.855 (0.004) & 0.831 (0.004) \\
Virchow2 \citep{zimmermann_virchow2_2024} & 0.710 (0.008) & 0.692 (0.004) & 0.701 (0.005) & \textbf{0.830} (0.006) & \textbf{0.870} (0.006) & \textbf{0.849} (0.006) \\
UNI2  \citep{chen_towards_2024}      & 0.712 (0.002) & 0.697 (0.002) & 0.705 (0.002) & \underline{0.828} (0.006) & \underline{0.869} (0.007) & \underline{0.848} (0.006) \\
\midrule
DinoBloom  \citep{linguraru_DinoBloom_2024} & 0.688 (0.004) & 0.676 (0.013) & 0.682 (0.009) & 0.778 (0.010) & 0.842 (0.008) & 0.809 (0.007) \\
LEMON  \citep{chadoutaud2026lemon}    & 0.701 (0.005) & 0.704 (0.011) & 0.702 (0.007) & 0.767 (0.010) & 0.840 (0.005) & 0.802 (0.007) \\
CytoFormer \citep{yao2026cytoformer} & \underline{0.735} (0.007) & 0.726 (0.002) & 0.730 (0.004) & 0.819 (0.009) & 0.861 (0.006) & 0.840 (0.006) \\
\midrule
C-LiTE & 0.728 (0.006) & \underline{0.733} (0.008) & \underline{0.731} (0.007) & 0.801 (0.006) & 0.857 (0.006) & 0.828 (0.006) \\
\textbf{C-LiTE$_V^*$} & 0.723 (0.004) & 0.706 (0.002) & 0.715 (0.003) & 0.798 (0.003) & 0.856 (0.004) & 0.826 (0.003) \\
\textbf{C-LiTE$_U^*$} & \textbf{0.741} (0.004) & \textbf{0.743} (0.008) & \textbf{0.742} (0.004) & 0.811 (0.002) & 0.866 (0.002) & 0.838 (0.002) \\
\bottomrule
\end{tabular}%
}
\end{table}

\begin{table}[H]
\centering
\captionsetup{width=\textwidth}
\caption{Per-class precision (P), recall (R) and F1 on CoNIC (nucleus classification). Values are mean (standard deviation) over five-fold cross-validation. The best value in each column is shown in \textbf{bold} and the second best is \underline{underlined}.}
\label{tab:CoNIC_perclass_prf1_main}

% ---- Panel 1 ----
\hspace*{-0.03\textwidth}\resizebox{1.06\textwidth}{!}{%
\begin{tabular}{l ccc ccc ccc}
\toprule
& \multicolumn{3}{c}{\textbf{Connective}} & \multicolumn{3}{c}{\textbf{Eosinophil}} & \multicolumn{3}{c}{\textbf{Epithelial}} \\
\cmidrule(lr){2-4}\cmidrule(lr){5-7}\cmidrule(lr){8-10}
\textbf{Model} & P & R & F1 & P & R & F1 & P & R & F1 \\
\midrule
DINOv2 (ViT-S/14) & 0.582 (0.002) & 0.522 (0.002) & 0.548 (0.002) & 0.441 (0.001) & 0.252 (0.001) & 0.316 (0.001) & 0.821 (0.001) & 0.904 (0.001) & 0.860 (0.001) \\

DINOv2 (ViT-B/14) & 0.584 (0.001) & 0.531 (0.003) & 0.555 (0.002) & 0.470 (0.001) & 0.254 (0.001) & 0.329 (0.001) & 0.818 (0.001) & 0.915 (0.001) & 0.864 (0.001) \\

DINOv2 (ViT-L/14) \citep{oquab_dinov2_2023} & 0.591 (0.001) & 0.576 (0.001) & 0.578 (0.001) & 0.501 (0.002) & 0.249 (0.001) & 0.327 (0.001) & 0.855 (0.000) & 0.896 (0.001) & 0.875 (0.001) \\
\midrule
Phikon \citep{filiot_phikon-v2_2024} & 0.644 (0.001) & 0.648 (0.002) & 0.639 (0.001) & 0.647 (0.010) & 0.513 (0.001) & 0.566 (0.003) & 0.894 (0.001) & 0.923 (0.001) & 0.904 (0.001) \\
Virchow2 \citep{zimmermann_virchow2_2024} & 0.673 (0.002) & 0.682 (0.001) & 0.679 (0.001) & 0.622 (0.011) & 0.455 (0.004) & 0.523 (0.006) & 0.911 (0.001) & 0.935 (0.001) & 0.920 (0.001) \\
UNI2 \citep{chen_towards_2024} & 0.687 (0.002) & 0.698 (0.002) & 0.692 (0.002) & 0.699 (0.013) & 0.600 (0.011) & 0.646 (0.011) & \underline{0.914} (0.001) & \textbf{0.944} (0.000) & \underline{0.929} (0.001) \\
\midrule
DinoBloom \citep{linguraru_DinoBloom_2024} & 0.625 (0.002) & 0.611 (0.002) & 0.618 (0.002) & 0.535 (0.010) & 0.344 (0.010) & 0.419 (0.007) & 0.862 (0.001) & 0.907 (0.001) & 0.884 (0.001) \\
LEMON \citep{chadoutaud2026lemon} & 0.692 (0.001) & 0.717 (0.003) & 0.704 (0.002) & 0.754 (0.010) & 0.627 (0.017) & 0.685 (0.010) & 0.878 (0.001) & 0.911 (0.001) & 0.894 (0.001) \\
CytoFormer \citep{yao2026cytoformer} & 0.738 (0.003) & \underline{0.761} (0.003) & 0.749 (0.003) & \textbf{0.879} (0.008) & \textbf{0.807} (0.014) & \textbf{0.841} (0.008) & 0.912 (0.000) & 0.932 (0.001) & 0.922 (0.001) \\
\midrule
\textbf{C-LiTE} & 0.720 (0.002) & 0.737 (0.002) & 0.728 (0.001) & 0.755 (0.007) & 0.654 (0.010) & 0.701 (0.006) & 0.897 (0.001) & 0.917 (0.001) & 0.907 (0.001) \\
\textbf{C-LiTE$_V^*$} & \underline{0.765} (0.002) & 0.739 (0.001) & \underline{0.752} (0.001) & 0.714 (0.016) & 0.572 (0.013) & 0.635 (0.013) & 0.903 (0.001) & 0.926 (0.001) & 0.915 (0.001) \\

\textbf{C-LiTE$_U^*$} & \textbf{0.793} (0.001) & \textbf{0.773} (0.003) & \textbf{0.783} (0.002) & \underline{0.757} (0.009) & \underline{0.667} (0.008) & \underline{0.709} (0.008) & \textbf{0.922} (0.000) & \underline{0.939} (0.000) & \textbf{0.930} (0.000) \\
\bottomrule
\end{tabular}%
}

\vspace{0.5em}

% ---- Panel 2 ----
\hspace*{-0.03\textwidth}\resizebox{1.06\textwidth}{!}{%
\begin{tabular}{l ccc ccc ccc}
\toprule
& \multicolumn{3}{c}{\textbf{Lymphocyte}} & \multicolumn{3}{c}{\textbf{Neutrophil}} & \multicolumn{3}{c}{\textbf{Plasma}} \\
\cmidrule(lr){2-4}\cmidrule(lr){5-7}\cmidrule(lr){8-10}
\textbf{Model} & P & R & F1 & P & R & F1 & P & R & F1 \\
\midrule
DINOv2 (ViT-S/14) & 0.585 (0.001) & 0.672 (0.001) & 0.619 (0.001) & 0.561 (0.001) & 0.225 (0.001) & 0.313 (0.003) & 0.540 (0.001) & 0.223 (0.001) & 0.308 (0.002) \\
DINOv2 (ViT-B/14) & 0.601 (0.001) & 0.652 (0.001) & 0.623 (0.001) & 0.611 (0.001) & 0.293 (0.001) & 0.390 (0.002) & 0.578 (0.001) & 0.251 (0.001) & 0.345 (0.001) \\
DINOv2 (ViT-L/14) \citep{oquab_dinov2_2023} & 0.598 (0.001) & 0.650 (0.001) & 0.621 (0.001) & 0.603 (0.001) & 0.325 (0.001) & 0.414 (0.007) & 0.564 (0.001) & 0.261 (0.001) & 0.356 (0.004) \\
\midrule
Phikon \citep{filiot_phikon-v2_2024} & 0.644 (0.001) & 0.671 (0.001) & 0.658 (0.001) & 0.652 (0.001) & 0.454 (0.001) & 0.527 (0.001) & 0.593 (0.001) & 0.374 (0.001) & 0.457 (0.001) \\
Virchow2 \citep{zimmermann_virchow2_2024} & 0.651 (0.003) & 0.702 (0.001) & 0.675 (0.001) & 0.642 (0.011) & 0.463 (0.005) & 0.532 (0.005) & 0.594 (0.003) & 0.405 (0.006) & 0.477 (0.005) \\
UNI2 \citep{chen_towards_2024} & 0.683 (0.001) & 0.701 (0.002) & 0.692 (0.002) & 0.695 (0.007) & 0.564 (0.012) & 0.622 (0.009) & 0.614 (0.002) & 0.438 (0.004) & 0.511 (0.004) \\
\midrule
DinoBloom \citep{linguraru_DinoBloom_2024} & 0.620 (0.001) & 0.683 (0.001) & 0.650 (0.001) & 0.610 (0.001) & 0.358 (0.001) & 0.451 (0.001) & 0.580 (0.001) & 0.288 (0.001) & 0.385 (0.001) \\
LEMON \citep{chadoutaud2026lemon} & 0.728 (0.002) & 0.762 (0.002) & 0.745 (0.002) & 0.655 (0.011) & 0.341 (0.012) & 0.448 (0.011) & \underline{0.663} (0.008) & 0.421 (0.008) & 0.515 (0.006) \\
CytoFormer \citep{yao2026cytoformer} & \textbf{0.764} (0.003) & 0.779 (0.001) & 0.772 (0.002) & \textbf{0.838} (0.014) & \textbf{0.683} (0.015) & \textbf{0.753} (0.014) & \textbf{0.672} (0.005) & \textbf{0.502} (0.005) & \textbf{0.575} (0.005) \\
\midrule
\textbf{C-LiTE} & 0.740 (0.001) & 0.768 (0.001) & 0.754 (0.001) & \underline{0.731} (0.007) & \underline{0.604} (0.010) & \underline{0.662} (0.007) & 0.662 (0.004) & \underline{0.476} (0.004) & \underline{0.554} (0.004) \\
\textbf{C-LiTE$_V^*$} & 0.749 (0.002) & \underline{0.825} (0.001) & \underline{0.785} (0.002) & 0.649 (0.008) & 0.454 (0.009) & 0.534 (0.007) & 0.611 (0.005) & 0.393 (0.006) & 0.478 (0.006) \\
\textbf{C-LiTE$_U^*$} & \underline{0.763} (0.002) & \textbf{0.830} (0.002) & \textbf{0.795} (0.002) & 0.690 (0.004) & 0.536 (0.007) & 0.603 (0.005) & 0.656 (0.004) & 0.461 (0.004) & 0.542 (0.002) \\
\bottomrule
\end{tabular}%
}
\end{table}

\begin{table}[H]
\centering
\captionsetup{width=\textwidth}
\caption{Per-class precision (P), recall (R) and F1 on MONKEY (lymphocyte versus monocyte classification). Values are mean (standard deviation) over five-fold cross-validation. The best value in each column is shown in \textbf{bold} and the second best is \underline{underlined}.}
\label{tab:monkey_perclass_prf1_main}
\resizebox{\textwidth}{!}{%
\begin{tabular}{l ccc ccc}
\toprule
& \multicolumn{3}{c}{\textbf{Lymphocyte}} & \multicolumn{3}{c}{\textbf{Monocyte}} \\
\cmidrule(lr){2-4}\cmidrule(lr){5-7}
\textbf{Model} & P & R & F1 & P & R & F1 \\
\midrule

DINOv2 (ViT-S/14) & 0.740 (0.001) & 0.880 (0.002) & 0.810 (0.001) & 0.650 (0.002) & 0.420 (0.003) & 0.510 (0.002) \\
DINOv2 (ViT-B/14) & 0.750 (0.002) & 0.880 (0.001) & 0.810 (0.001) & 0.660 (0.002) & 0.450 (0.006) & 0.540 (0.005) \\
DINOv2 (ViT-L/14)  \citep{oquab_dinov2_2023} & 0.750 (0.002) & 0.870 (0.002) & 0.810 (0.001) & 0.650 (0.003) & 0.450 (0.006) & 0.540 (0.004) \\
\midrule
Phikon  \citep{filiot_phikon-v2_2024}  & 0.780 (0.001) & 0.870 (0.002) & 0.830 (0.001) & 0.690 (0.002) & 0.550 (0.005) & 0.610 (0.002) \\
Virchow2 \citep{zimmermann_virchow2_2024} & 0.790 (0.002) & 0.880 (0.003) & 0.830 (0.002) & 0.700 (0.004) & 0.550 (0.006) & 0.620 (0.005) \\
UNI2  \citep{chen_towards_2024}      & 0.802 (0.003) & 0.880 (0.003) & 0.839 (0.003) & 0.723 (0.006) & 0.589 (0.007) & 0.649 (0.006) \\
\midrule
DinoBloom  \citep{linguraru_DinoBloom_2024} & 0.770 (0.001) & 0.870 (0.003) & 0.820 (0.001) & 0.680 (0.003) & 0.520 (0.005) & 0.590 (0.003) \\
LEMON  \citep{chadoutaud2026lemon}    & 0.825 (0.002) & \underline{0.892 (0.003)} & 0.857 (0.001) & 0.759 (0.004) & 0.643 (0.005) & 0.696 (0.003) \\
CytoFormer \citep{yao2026cytoformer} & \underline{0.834 (0.001)} & 0.890 (0.003) & \textbf{0.861 (0.001)} & \underline{0.762 (0.004)} & \underline{0.665 (0.004)} & \textbf{0.710 (0.002)} \\
\midrule
\textbf{C-LiTE } & 0.832 (0.001) & \textbf{0.893 (0.004)} & \textbf{0.861 (0.001)} & \textbf{0.765 (0.005)} & 0.659 (0.004) & \underline{0.708 (0.002)} \\
\textbf{C-LiTE$_V^*$} & 0.827 (0.003) & 0.882 (0.002) & 0.854 (0.002) & 0.745 (0.004) & 0.651 (0.007) & 0.695 (0.005) \\
\textbf{C-LiTE$_U^*$} & \textbf{0.836 (0.001)} & 0.882 (0.003) & \underline{0.858 (0.001)} & 0.751 (0.004) & \textbf{0.674 (0.003)} & \textbf{0.710 (0.002)} \\
\bottomrule
\end{tabular}%
}
\end{table}

\begin{table}[H]
\centering
\captionsetup{width=\textwidth}
\caption{Per-class precision (P), recall (R) and F1 on Immunocto (immune cell phenotyping). Results are from the single official train/test split, so no standard deviations are reported. The best value in each column is shown in \textbf{bold} and the second best is \underline{underlined}.}
\label{tab:immune_perclass_prf1_main}
\resizebox{\textwidth}{!}{%
\begin{tabular}{ll ccc ccc ccc ccc ccc}
\toprule
& & \multicolumn{3}{c}{\textbf{B Cell}} & \multicolumn{3}{c}{\textbf{CD4}} & \multicolumn{3}{c}{\textbf{CD8}} & \multicolumn{3}{c}{\textbf{Macrophage}} & \multicolumn{3}{c}{\textbf{Other}} \\
\cmidrule(lr){3-5}\cmidrule(lr){6-8}\cmidrule(lr){9-11}\cmidrule(lr){12-14}\cmidrule(lr){15-17}
\textbf{Model} & \textbf{Params} & P & R & F1 & P & R & F1 & P & R & F1 & P & R & F1 & P & R & F1 \\
\midrule
DINOv2 (ViT-S/14) & 22M  & 0.562 & \textbf{0.655} & 0.601 & 0.403 & 0.392 & 0.395 & 0.381 & 0.02 & 0.031 & 0.602 & 0.061 & 0.115 & 0.763 & 0.962 & 0.851 \\
DINOv2 (ViT-B/14) & 86M  & 0.603 & 0.632 & 0.610 & 0.414 & 0.402 & 0.400 & 0.372 & 0.013 & 0.020 & 0.622 & 0.083 & 0.140 & 0.762 & 0.963 & 0.850 \\
DINOv2 (ViT-L/14)  \citep{oquab_dinov2_2023}& 300M & 0.591 & \textbf{0.652} & 0.620 & 0.422 & 0.403 & 0.410 & 0.412 & 0.022 & 0.040 & 0.631 & 0.092 & 0.150 & 0.771 & 0.962 & 0.850 \\
\midrule
Phikon \citep{filiot_phikon-v2_2024}           & 85.8M & 0.572 & 0.641 & 0.604 & 0.452 & 0.501 & 0.471 & 0.452 & 0.051 & 0.083 & 0.633 & 0.142 & 0.226 & 0.803 & 0.963 & 0.873 \\
Virchow2 \citep{zimmermann_virchow2_2024}         & 632M  & 0.611 & 0.631 & 0.620 & 0.453 & 0.531 & 0.486 & 0.464 & 0.062 & 0.099 & 0.641 & 0.133 & 0.223 & 0.802 & 0.961 & 0.876 \\
UNI2  \citep{chen_towards_2024}              & 681M    & 0.604 & \underline{0.645} & \textbf{0.624} & 0.464 & 0.539 & 0.499 & 0.544 & 0.077 & 0.135 & 0.678 & 0.293 & 0.409 & 0.821 & 0.960 & 0.885 \\
\midrule
DinoBloom  \citep{linguraru_DinoBloom_2024}         & 22M   & 0.562 & 0.613 & 0.587 & 0.434 & 0.522 & 0.470 & 0.421 & 0.033 & 0.0621 & 0.682 & 0.131 & 0.217 & 0.802 & 0.961 & 0.871 \\
LEMON  \citep{chadoutaud2026lemon}            & 22M   & 0.456 & 0.581 & 0.511 & 0.468 & \textbf{0.772} & 0.583 & 0.513 & 0.093 & 0.157 & 0.662 & 0.200 & 0.307 & 0.881 & 0.950 & 0.914 \\
CytoFormer  \citep{yao2026cytoformer}       & 681M    & 0.607 & 0.628 & 0.617 & \textbf{0.515} & 0.749 & \textbf{0.611} & \textbf{0.568} & \textbf{0.176} & \textbf{0.269} & \textbf{0.712} & \underline{0.321} & \underline{0.443} & \underline{0.892} & \textbf{0.969} & \textbf{0.929} \\
\midrule
\textbf{C-LiTE } & 22M   & 0.588 & 0.631 & 0.609 & 0.507 & 0.751 & 0.605 & 0.551 & 0.146 & 0.231 & 0.688 & 0.294 & 0.412 & 0.888 & \underline{0.967} & 0.926 \\
\textbf{C-LiTE$_V^*$} & 22M & \textbf{0.620} & 0.582 & 0.600 & 0.506 & \underline{0.755} & \underline{0.606} & \underline{0.567} & 0.153 & 0.241 & \underline{0.697} & \textbf{0.354} & \textbf{0.469} & \textbf{0.893} & 0.965 & \underline{0.928} \\
\textbf{C-LiTE$_U^*$} & 22M & \underline{0.613} & 0.633 & \underline{0.623} & \underline{0.512} & 0.731 & 0.602 & 0.546 & \underline{0.167} & \underline{0.255} & 0.695 & 0.311 & 0.430 & 0.886 & \underline{0.967} & 0.925 \\
\bottomrule
\end{tabular}%
}
\end{table}

\begin{table}[H]
\centering
\captionsetup{width=\textwidth}
\caption{Per-class precision (P), recall (R) and F1 on PUMA (nuclear phenotyping). Values are mean (standard deviation) over five-fold cross-validation. The best value in each column is shown in \textbf{bold} and the second best is \underline{underlined}.}
\label{tab:puma_perclass_prf1_main}

\hspace*{-0.03\textwidth}\resizebox{1.06\textwidth}{!}{%
\begin{tabular}{l ccc ccc ccc ccc}
\toprule
& \multicolumn{3}{c}{\textbf{Apoptosis}} & \multicolumn{3}{c}{\textbf{Endothelium}} & \multicolumn{3}{c}{\textbf{Epithelium}} & \multicolumn{3}{c}{\textbf{Histiocyte}} \\
\cmidrule(lr){2-4}\cmidrule(lr){5-7}\cmidrule(lr){8-10}\cmidrule(lr){11-13}
\textbf{Model} & P & R & F1 & P & R & F1 & P & R & F1 & P & R & F1 \\
\midrule
DINOv2 (ViT-S/14) & 0.50 (0.028) & 0.21 (0.020) & 0.30 (0.024) & 0.43 (0.038) & 0.15 (0.024) & 0.23 (0.029) & 0.63 (0.018) & 0.31 (0.017) & 0.42 (0.018) & 0.43 (0.030) & 0.15 (0.008) & 0.23 (0.012) \\
DINOv2 (ViT-B/14) & 0.51 (0.026) & 0.21 (0.011) & 0.30 (0.012) & 0.39 (0.028) & 0.18 (0.023) & 0.24 (0.026) & 0.60 (0.020) & 0.39 (0.025) & 0.47 (0.023) & 0.41 (0.020) & 0.17 (0.007) & 0.24 (0.010) \\
DINOv2 (ViT-L/14)  \citep{oquab_dinov2_2023} & 0.38 (0.058) & 0.22 (0.025) & 0.28 (0.034) & 0.36 (0.031) & 0.20 (0.016) & 0.26 (0.020) & 0.61 (0.021) & 0.41 (0.015) & 0.49 (0.008) & 0.40 (0.011) & 0.17 (0.005) & 0.24 (0.007) \\
\midrule
Phikon \citep{filiot_phikon-v2_2024} & \underline{0.55 (0.024)} & 0.43 (0.024) & \underline{0.48 (0.022)} & 0.43 (0.032) & 0.28 (0.013) & 0.34 (0.017) & 0.71 (0.017) & 0.55 (0.018) & 0.62 (0.016) & 0.44 (0.013) & 0.26 (0.012) & 0.32 (0.012) \\
Virchow2 \citep{zimmermann_virchow2_2024} & 0.50 (0.012) & 0.43 (0.022) & 0.46 (0.013) & \textbf{0.52 (0.028)} & \underline{0.33 (0.012)} & \textbf{0.40 (0.015)} & \textbf{0.78 (0.013)} & \textbf{0.65 (0.017)} & \textbf{0.71 (0.012)} & 0.44 (0.011) & 0.29 (0.012) & 0.35 (0.012) \\
UNI2  \citep{chen_towards_2024} & 0.47 (0.015) & \underline{0.44 (0.030)} & 0.45 (0.014) & 0.43 (0.021) & \textbf{0.36 (0.021)} & \underline{0.39 (0.021)} & \underline{0.73 (0.031)} & \textbf{0.65 (0.008)} & \underline{0.69 (0.017)} & 0.52 (0.008) & \underline{0.37 (0.010)} & \underline{0.44 (0.009)} \\
\midrule
DinoBloom  \citep{linguraru_DinoBloom_2024} & 0.49 (0.017) & 0.24 (0.024) & 0.32 (0.022) & 0.41 (0.030) & 0.16 (0.018) & 0.23 (0.022) & 0.65 (0.016) & 0.39 (0.021) & 0.49 (0.018) & 0.49 (0.021) & 0.24 (0.011) & 0.32 (0.014) \\
LEMON  \citep{chadoutaud2026lemon} & 0.42 (0.022) & 0.20 (0.023) & 0.27 (0.021) & 0.40 (0.084) & 0.10 (0.013) & 0.16 (0.022) & 0.60 (0.023) & 0.19 (0.027) & 0.29 (0.032) & 0.49 (0.022) & 0.22 (0.016) & 0.30 (0.018) \\
CytoFormer \citep{yao2026cytoformer} & 0.51 (0.015) & \textbf{0.46 (0.026)} & \textbf{0.49 (0.015)} & 0.47 (0.028) & \underline{0.33 (0.019)} & \underline{0.39 (0.021)} & \underline{0.73 (0.012)} & \underline{0.58 (0.024)} & 0.64 (0.016) & \underline{0.55 (0.011)} & \textbf{0.40 (0.014)} & \textbf{0.46 (0.012)} \\
\midrule
\textbf{C-LiTE} & 0.54 (0.025) & 0.40 (0.057) & 0.46 (0.040) & \underline{0.51 (0.054)} & 0.24 (0.023) & 0.33 (0.031) & 0.60 (0.011) & 0.34 (0.021) & 0.43 (0.018) & \textbf{0.56 (0.012)} & 0.29 (0.007) & 0.38 (0.008) \\
\textbf{C-LiTE$_V^*$} & 0.52 (0.017) & 0.31 (0.052) & 0.39 (0.046) & 0.45 (0.041) & 0.18 (0.021) & 0.26 (0.028) & 0.64 (0.032) & 0.34 (0.015) & 0.44 (0.018) & 0.50 (0.011) & 0.24 (0.009) & 0.33 (0.011) \\
\textbf{C-LiTE$_U^*$} & \textbf{0.56 (0.017)} & 0.40 (0.040) & 0.47 (0.033) & \textbf{0.52 (0.005)} & 0.26 (0.009) & 0.35 (0.009) & 0.71 (0.017) & 0.50 (0.024) & 0.59 (0.018) & 0.53 (0.005) & 0.34 (0.007) & 0.41 (0.006) \\
\bottomrule
\end{tabular}%
}

\vspace{0.5em}

\hspace*{-0.03\textwidth}\resizebox{1.06\textwidth}{!}{%
\begin{tabular}{l ccc ccc ccc}
\toprule
& \multicolumn{3}{c}{\textbf{Lymphocyte}} & \multicolumn{3}{c}{\textbf{Melanophage}} & \multicolumn{3}{c}{\textbf{Neutrophil}} \\
\cmidrule(lr){2-4}\cmidrule(lr){5-7}\cmidrule(lr){8-10}
\textbf{Model} & P & R & F1 & P & R & F1 & P & R & F1 \\
\midrule
DINOv2 (ViT-S/14) & 0.64 (0.009) & 0.69 (0.008) & 0.66 (0.007) & 0.55 (0.042) & 0.34 (0.020) & 0.42 (0.022) & 0.57 (0.071) & 0.32 (0.035) & 0.41 (0.039) \\
DINOv2 (ViT-B/14) & 0.66 (0.006) & 0.69 (0.006) & 0.68 (0.005) & 0.45 (0.022) & 0.36 (0.034) & 0.40 (0.030) & \textbf{0.67 (0.046)} & 0.32 (0.038) & 0.43 (0.032) \\
DINOv2 (ViT-L/14)  \citep{oquab_dinov2_2023} & 0.64 (0.005) & 0.71 (0.007) & 0.68 (0.006) & 0.47 (0.047) & 0.37 (0.043) & 0.42 (0.044) & 0.48 (0.095) & 0.34 (0.051) & 0.40 (0.061) \\
\midrule
Phikon \citep{filiot_phikon-v2_2024} & 0.68 (0.004) & 0.73 (0.007) & 0.70 (0.003) & 0.58 (0.041) & \textbf{0.56 (0.030)} & 0.57 (0.031) & 0.48 (0.042) & 0.42 (0.024) & 0.45 (0.029) \\
Virchow2 \citep{zimmermann_virchow2_2024} & 0.70 (0.005) & 0.74 (0.005) & 0.72 (0.004) & 0.52 (0.054) & \underline{0.54 (0.035)} & 0.53 (0.038) & 0.45 (0.065) & 0.37 (0.087) & 0.41 (0.076) \\
UNI2  \citep{chen_towards_2024} & 0.73 (0.004) & 0.76 (0.007) & 0.74 (0.003) & 0.51 (0.032) & 0.50 (0.047) & 0.51 (0.033) & 0.42 (0.036) & 0.42 (0.061) & 0.42 (0.044) \\
\midrule
DinoBloom  \citep{linguraru_DinoBloom_2024} & 0.68 (0.006) & 0.74 (0.008) & 0.71 (0.006) & 0.52 (0.028) & 0.53 (0.018) & 0.52 (0.022) & 0.48 (0.063) & 0.32 (0.075) & 0.38 (0.066) \\
LEMON  \citep{chadoutaud2026lemon} & 0.72 (0.005) & 0.77 (0.010) & 0.75 (0.005) & \textbf{0.75 (0.032)} & 0.47 (0.040) & \underline{0.58 (0.032)} & 0.56 (0.038) & 0.47 (0.023) & 0.51 (0.027) \\
CytoFormer \citep{yao2026cytoformer} & \underline{0.77 (0.005)} & \underline{0.80 (0.009)} & 0.78 (0.006) & 0.60 (0.051) & 0.50 (0.022) & 0.55 (0.033) & \underline{0.61 (0.050)} & \underline{0.58 (0.057)} & \textbf{0.59 (0.047)} \\
\midrule
\textbf{C-LiTE} & 0.76 (0.003) & \textbf{0.82 (0.007)} & \underline{0.79 (0.003)} & \underline{0.70 (0.028)} & \underline{0.54 (0.031)} & \textbf{0.61 (0.027)} & 0.52 (0.039) & \textbf{0.63 (0.059)} & 0.57 (0.044) \\
\textbf{C-LiTE$_V^*$} & 0.76 (0.004) & \underline{0.80 (0.004)} & 0.78 (0.002) & 0.56 (0.037) & 0.44 (0.046) & 0.49 (0.032) & 0.53 (0.059) & 0.55 (0.055) & 0.53 (0.053) \\
\textbf{C-LiTE$_U^*$} & \textbf{0.78 (0.004)} & \textbf{0.82 (0.007)} & \textbf{0.80 (0.003)} & 0.56 (0.037) & 0.50 (0.044) & 0.53 (0.034) & 0.58 (0.044) & \underline{0.58 (0.041)} & \underline{0.58 (0.039)} \\
\bottomrule
\end{tabular}%
}

\vspace{0.5em}

\hspace*{-0.03\textwidth}\resizebox{1.06\textwidth}{!}{%
\begin{tabular}{l ccc ccc ccc}
\toprule
& \multicolumn{3}{c}{\textbf{Plasma Cell}} & \multicolumn{3}{c}{\textbf{Stroma}} & \multicolumn{3}{c}{\textbf{Tumour}} \\
\cmidrule(lr){2-4}\cmidrule(lr){5-7}\cmidrule(lr){8-10}
\textbf{Model} & P & R & F1 & P & R & F1 & P & R & F1 \\
\midrule
DINOv2 (ViT-S/14) & 0.38 (0.109) & 0.15 (0.047) & 0.22 (0.065) & 0.44 (0.013) & 0.21 (0.017) & 0.29 (0.017) & 0.78 (0.003) & 0.90 (0.002) & 0.84 (0.002) \\
DINOv2 (ViT-B/14) & 0.50 (0.064) & 0.17 (0.032) & 0.26 (0.040) & 0.47 (0.008) & 0.25 (0.011) & 0.32 (0.008) & 0.79 (0.003) & \underline{0.91 (0.002)} & 0.84 (0.002) \\
DINOv2 (ViT-L/14)  \citep{oquab_dinov2_2023} & 0.33 (0.095) & 0.13 (0.028) & 0.19 (0.039) & 0.46 (0.027) & 0.25 (0.025) & 0.33 (0.026) & 0.80 (0.003) & 0.89 (0.003) & 0.85 (0.002) \\
\midrule
Phikon \citep{filiot_phikon-v2_2024} & \textbf{0.70 (0.049)} & 0.44 (0.061) & \textbf{0.54 (0.040)} & \underline{0.53 (0.006)} & 0.31 (0.023) & 0.39 (0.018) & 0.84 (0.003) & \underline{0.91 (0.002)} & 0.87 (0.001) \\
Virchow2 \citep{zimmermann_virchow2_2024} & 0.57 (0.050) & \textbf{0.46 (0.047)} & 0.51 (0.037) & 0.47 (0.011) & 0.36 (0.012) & 0.41 (0.010) & \underline{0.86 (0.001)} & \textbf{0.92 (0.001)} & \textbf{0.89 (0.001)} \\
UNI2   \citep{chen_towards_2024} & 0.47 (0.023) & 0.42 (0.044) & 0.44 (0.026) & \underline{0.53 (0.023)} & \underline{0.39 (0.014)} & 0.45 (0.017) & \textbf{0.87 (0.002)} & \textbf{0.92 (0.002)} & \textbf{0.89 (0.001)} \\
\midrule
DinoBloom  \citep{linguraru_DinoBloom_2024} & 0.32 (0.039) & 0.12 (0.024) & 0.17 (0.024) & 0.48 (0.014) & 0.29 (0.027) & 0.36 (0.024) & 0.81 (0.004) & 0.90 (0.001) & 0.85 (0.002) \\
LEMON \citep{chadoutaud2026lemon} & 0.60 (0.187) & 0.17 (0.033) & 0.27 (0.055) & 0.51 (0.023) & 0.34 (0.015) & 0.41 (0.016) & 0.79 (0.003) & 0.90 (0.003) & 0.84 (0.002) \\
CytoFormer \citep{yao2026cytoformer} & \underline{0.64 (0.030)} & 0.44 (0.037) & \underline{0.52 (0.027)} & 0.51 (0.016) & \underline{0.39 (0.014)} & 0.44 (0.014) & \underline{0.86 (0.002)} & \underline{0.91 (0.002)} & \underline{0.88 (0.001)} \\
\midrule
\textbf{C-LiTE} & 0.52 (0.050) & \textbf{0.46 (0.036)} & 0.49 (0.036) & \textbf{0.56 (0.015)} & \underline{0.39 (0.016)} & \underline{0.46 (0.016)} & 0.83 (0.001) & \underline{0.91 (0.002)} & 0.87 (0.001) \\
\textbf{C-LiTE$_V^*$} & 0.57 (0.107) & 0.33 (0.013) & 0.41 (0.028) & \underline{0.53 (0.015)} & 0.35 (0.016) & 0.42 (0.015) & 0.81 (0.002) & \underline{0.91 (0.002)} & 0.86 (0.001) \\
\textbf{C-LiTE$_U^*$} & 0.57 (0.018) & \underline{0.45 (0.033)} & 0.50 (0.024) & \textbf{0.56 (0.022)} & \textbf{0.41 (0.015)} & \textbf{0.47 (0.017)} & 0.84 (0.003) & \textbf{0.92 (0.002)} & \underline{0.88 (0.002)} \\
\bottomrule
\end{tabular}%
}
\end{table}
\end{document}